\PassOptionsToPackage{table}{xcolor}
\documentclass[letterpaper]{article}

\usepackage{microtype}
\usepackage{graphicx}
\usepackage{booktabs} 

\usepackage{hyperref}

\usepackage[accepted]{icml2026}

\usepackage{amsmath}
\usepackage{amssymb}
\usepackage{mathtools}
\usepackage{amsthm}
\usepackage{multirow}
\usepackage{tabularx}

\usepackage{enumitem}
\usepackage{url}
\usepackage{booktabs}    
\usepackage{subcaption}
\usepackage{ulem}
\usepackage{float}
\usepackage{placeins}
\usepackage{wrapfig}
\usepackage{caption}    
\usepackage{array}      
\usepackage{amsmath}
\usepackage{arydshln}
\usepackage{makecell}
\usepackage{array} 
\usepackage{booktabs}
\usepackage{tabularx}
\usepackage[table]{xcolor}

\usepackage[capitalize,noabbrev]{cleveref}

\theoremstyle{plain}

\theoremstyle{definition}

\theoremstyle{remark}

\usepackage[textsize=tiny]{todonotes}

\icmltitlerunning{Learning More from Less: Unlocking Internal Representations for Benchmark Compression}

\begin{document}

\twocolumn[
  \icmltitle{Learning More from Less: Unlocking Internal Representations \texorpdfstring{\\}{ } for Benchmark Compression}



  \icmlsetsymbol{equal}{*}

  \begin{icmlauthorlist}
    \icmlauthor{Yueqi Zhang}{equal,sch}
    \icmlauthor{Jin Hu}{equal,sch}
    \icmlauthor{Shaoxiong Feng}{comp}
    \icmlauthor{Peiwen Yuan}{sch}
    \icmlauthor{Xinglin Wang}{sch}
    \icmlauthor{Yiwei Li}{sch}
    \icmlauthor{Jiayi Shi}{sch}
    \icmlauthor{Chuyi Tan}{sch}
    \icmlauthor{Ji Zhang}{sch}
    \icmlauthor{Boyuan Pan}{comp}
    \icmlauthor{Yao Hu}{comp}
    \icmlauthor{Kan Li}{sch}
  \end{icmlauthorlist}

  \icmlaffiliation{comp}{Xiaohongshu Inc}
  \icmlaffiliation{sch}{School of Computer Science, Beijing Institute of Technology}

  \icmlcorrespondingauthor{Kan Li}{likan@bit.edu.cn}
  \icmlcorrespondingauthor{Boyuan Pan}{panboyuan@xiaohongshu.com}
  
  \icmlkeywords{Machine Learning, ICML}

  \vskip 0.3in
]



\printAffiliationsAndNotice{\icmlEqualContribution}

\begin{abstract}
    The prohibitive cost of evaluating Large Language Models (LLMs) necessitates efficient alternatives to full-scale benchmarking. Prevalent approaches address this by identifying a small coreset of items to approximate full-benchmark performance. However, existing methods must estimate a reliable item profile from response patterns across many source models, which becomes statistically unstable when the source pool is small. This dependency is particularly limiting for newly released benchmarks with minimal historical evaluation data. We argue that discrete correctness labels are a lossy view of the model's decision process and fail to capture information encoded in hidden states. To address this, we introduce \textsc{RepCore}, which aligns heterogeneous hidden states into a unified latent space to construct representative coresets. Using these subsets for performance extrapolation, \textsc{RepCore} achieves precise estimation accuracy with as few as ten source models. Experiments on five benchmarks and over 200 models show consistent gains over output-based baselines in ranking correlation and estimation accuracy. Spectral analysis further indicates that the aligned representations contain separable components reflecting broad response tendencies and task-specific reasoning patterns.
\end{abstract}
\section{Introduction}

\begin{figure*}[!t]
  \centering
  \includegraphics[width=\textwidth]{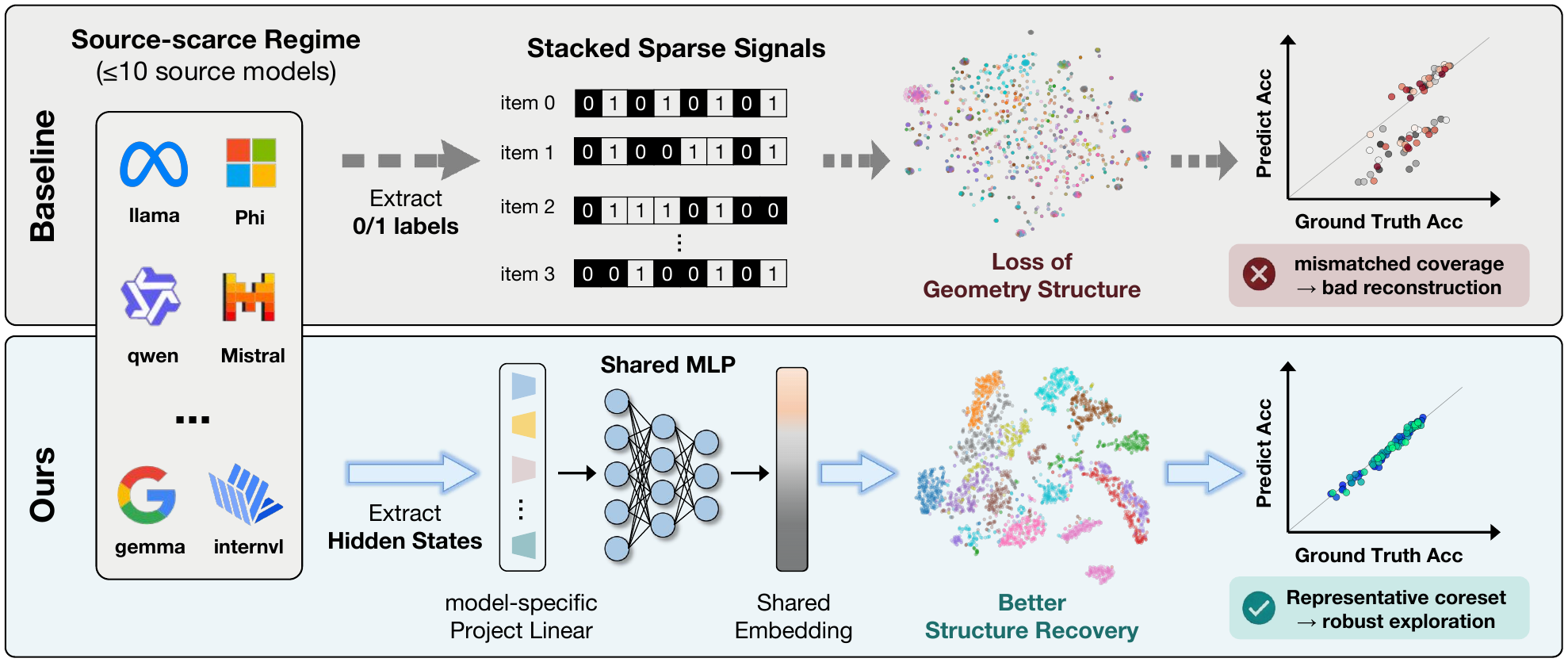}
  \caption{Comparison of item representation paradigms in source-scarce regimes. Top: Output-based methods rely on sparse 0/1 signals that fail to preserve the geometric structure of the item space. Bottom: \textsc{RepCore} aligns heterogeneous hidden states into a unified latent space to recover fine-grained item structures for robust coreset selection.}
  \label{fig:main_figure}
\end{figure*}

Large Language Models (LLMs) have demonstrated strong capabilities across diverse domains and reshaped modern AI research~\cite{achiam2023gpt,grattafiori2024llama,qwen2025qwen3,marjanovic2025deepseek}. Benchmark suites have therefore expanded in size, often reaching tens of thousands of items, as the community continually adds new tasks and test cases while preserving comparability over time. However, full-scale evaluation at this scale incurs substantial computational and financial costs, making frequent re-evaluation impractical during rapid model iteration~\cite{liang2023holistic,perlitz-etal-2024-efficient}. Consequently, recent work on \textit{benchmark compression} selects a small coreset and evaluates models on this subset to approximate full-benchmark scores and rankings~\cite{vivek2024anchor,pmlr-v235-maia-polo24a,yuan2025beyond}.

These methods generally follow a similar pipeline. They first collect output-level responses from many source models on the full benchmark, forming a model-by-item matrix of correctness or confidence. Each item is then represented by its response vector across models or by low-dimensional factors estimated from this matrix, for example through Item Response Theory, and these representations support both coreset selection and score extrapolation. This pipeline implicitly assumes broad source coverage. When only a handful of source models are available, as is often the case for newly released or long-tail benchmarks with limited historical evaluations, these output-derived item representations are estimated unreliably. As a result, the geometric structure of the item space collapses and many distinct items become indistinguishable, as shown in Figure~\ref{fig:main_figure}, leading to less representative coresets and larger extrapolation errors.

Motivated by this limitation, we investigate the internal representations produced during inference. Unlike discrete output labels, hidden states provide a high-dimensional and continuous view of the computation that generates an answer. Prior work has shown that these states encode rich internal features of uncertainty, problem difficulty, and task-dependent reasoning patterns~\cite{azaria2023internal,zou2025reptask,zhang2025reasoningmodelsknowtheyre,orgad2025llms,wang-etal-2025-faclens}. We hypothesize that hidden states offer a dense signal for characterizing items, but exploiting this signal requires aligning heterogeneous hidden spaces across architectures. To this end, we introduce \textsc{RepCore}, a framework that aligns heterogeneous hidden states into a unified latent space to recover fine-grained item structures.

\textsc{RepCore} utilizes a shared projection mechanism to map model-specific hidden states into aligned embeddings, recovering the geometric structure of the item space. As shown in the bottom panel of Figure~\ref{fig:main_figure}, aggregating these embeddings across source models reveals structured clusters associated with distinct reasoning patterns. Within this aggregated space, we employ a clustering-based strategy to select a diverse set of anchor items that serve as the training basis for a lightweight linear regressor. By extrapolating the target model's performance from these anchors, our approach yields accurate full-benchmark estimates and remains effective even with as few as ten source models.


We conduct experiments on five benchmarks spanning text-only and multimodal settings, using over two hundred models that include text-only and multimodal variants and cover both dense and mixture-of-experts architectures. Under matched coreset sizes with ten source models, our method consistently improves full-benchmark estimation accuracy and ranking consistency over competitive output-based baselines. Our contributions are summarized as follows, with related work deferred to Appendix~\ref{sec:Related_Work}:

\begin{itemize}[leftmargin=1.6em, itemsep=8.0pt, parsep=0.5pt, topsep=0.4pt]
\item We formalize benchmark compression under source scarcity and show that sparse output-level signals distort the geometry of the item space, degrading both coreset selection and performance extrapolation.
\item We propose \textsc{RepCore}, which maps model-specific hidden states into aligned embeddings and uses the resulting space to enable robust coreset selection and accurate extrapolation with few source models.
\item We provide empirical and structural evidence supporting internal representations. Our results demonstrate that the aligned embeddings contain separable components that reflect broad response tendencies and task-specific reasoning patterns.
\end{itemize}

\section{RepCore Approach}
\label{sec:method}

\begin{figure*}[!t]
  \centering
  \includegraphics[width=\textwidth]{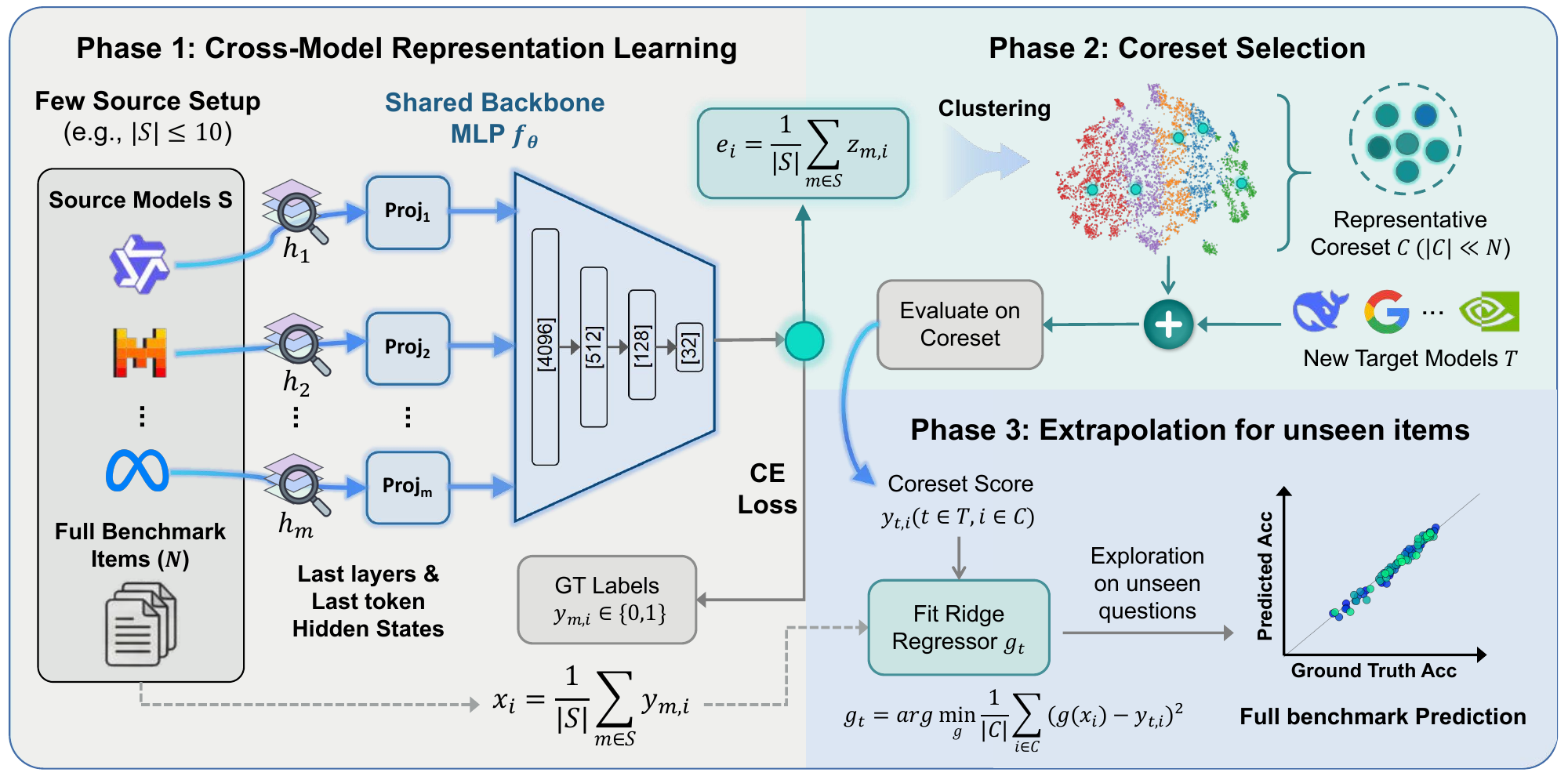}
  \caption{Overview of the \textsc{RepCore} framework. The pipeline proceeds in three phases: aligning heterogeneous hidden states into a unified latent space via model-specific projections and a shared MLP, selecting representative anchor items through consensus clustering, and extrapolating full-benchmark performance using a lightweight regressor.}
  \label{fig:method_overview}
  \vskip -0.15in
\end{figure*}

We propose \textsc{RepCore}, a framework designed to achieve high-fidelity benchmark compression in source-scarce regimes. As illustrated in Figure~\ref{fig:method_overview}, the pipeline operates in three consecutive phases. First, we learn aligned representations from model hidden states to recover the benchmark's geometric structure (\S\ref{sec:embedding}). Second, we perform consensus clustering in this latent space to construct a representative coreset (\S\ref{sec:coreset}). Finally, we employ a lightweight and robust extrapolator to map local coreset evaluations to full-benchmark performance estimations (\S\ref{sec:extrapolation}).

\subsection{Problem Setup}
\label{sec:setup}

Let $\mathcal{I}$ denote the comprehensive set of items in a benchmark. We consider a set of available source models $\mathcal{S}$ and a set of unseen target models $\mathcal{T}$, operating specifically within a source-scarce regime where $|\mathcal{S}|$ is strictly limited. We adopt the standard \textit{free-generation then binary-scoring} paradigm, where each source model $m \in \mathcal{S}$ generates a response for item $i \in \mathcal{I}$ that is deterministically parsed to yield a binary correctness label $y_{m,i} \in \{0,1\}$. Concurrently, we extract the hidden state $h_{m,i}$ associated with the generation. Our objective is to learn item representations that capture the relational structure of the benchmark, thereby guiding the selection of a minimal coreset $\mathcal{C} \subset \mathcal{I}$. By evaluating target models exclusively on $\mathcal{C}$, we aim to accurately estimate their performance on the full set $\mathcal{I}$.

\subsection{Cross-Model Representation Learning}
\label{sec:embedding}


Under the source-scarce regime (specifically $|\mathcal{S}| \leq 10$), the response matrix $Y$ contains only $|\mathcal{S}|$ rows and provides a highly limited view of inter-item relations. 
We therefore use hidden states as dense item-level signals to recover the benchmark's latent relational structure.

\paragraph{Hidden State Extraction} For each model-item pair $(m, i)$, we extract the final layer's hidden state of the final output token, denoted as $h_{m,i}$. Validated by recent interpretability studies \cite{burns2023discovering,wang2025latent,orgad2025llms,gekhman2025insideout}, these states encode the reasoning trajectory and decision signals, providing a high-fidelity information source superior to discrete correctness labels.

\paragraph{Cross-Model Alignment and Supervision} Source models exhibit significant architectural heterogeneity, possessing distinct hidden dimensions and feature spaces. 
To obtain comparable representations across models and reduce mismatch caused by heterogeneous hidden spaces, we first apply an independent model-specific linear projection layer $\mathrm{Proj}_m$ to each source model. 
For a source model $m$ with native hidden dimension $d_m$, this projection maps the hidden state into a shared intermediate space, i.e., $\mathrm{Proj}_m:\mathbb{R}^{d_m}\rightarrow\mathbb{R}^{4096}$. 
The projected features are then processed by a source-agnostic MLP $f_\theta$ shared across all source models, with layer widths $4096\rightarrow512\rightarrow128\rightarrow32$ (Appendix~\ref{app:mlp_ablation} further examines alternative MLP widths and depths). 
This pipeline acts as an information bottleneck, progressively compressing the high-dimensional input into an aligned representation:
\begin{equation}
    \setlength{\abovedisplayskip}{4pt}
    \setlength{\belowdisplayskip}{4pt}
    \setlength{\abovedisplayshortskip}{4pt}
    \setlength{\belowdisplayshortskip}{4pt}
    z_{m,i} = f_\theta\big(\mathrm{Proj}_m(h_{m,i})\big) \in \mathbb{R}^{d_z}
\end{equation}
We refer to $z_{m,i}$ as the aligned embedding of item $i$ under source model $m$.
The dimension $d_z$ determines the bottleneck capacity. We set $d_z = 32$ for all subsequent experiments, as we empirically observe that for most models, the effective rank of the aligned embedding matrix (capturing 99\% variance) remains below 20. Thus, 32 dimensions provide ample capacity to maintain representational richness while ensuring computational efficiency.

To explicitly extract signals directly tied to the final evaluation outcome, we apply a linear classifier $g_\phi$ to $z_{m,i}$ and optimize the entire network using Cross-Entropy loss to predict the ground-truth correctness $y_{m,i}$:
\begin{equation}
    \setlength{\abovedisplayskip}{4pt}
    \setlength{\belowdisplayskip}{4pt}
    \setlength{\abovedisplayshortskip}{4pt}
    \setlength{\belowdisplayshortskip}{4pt}
    \mathcal{L} = \sum_{m \in \mathcal{S}} \sum_{i \in \mathcal{I}_{\text{train}}} \mathrm{CE}\left(g_\phi(z_{m,i}), y_{m,i}\right)
\end{equation}
We define $\mathcal{I}_{\text{train}}$ by randomly partitioning the target benchmark $\mathcal{I}$ into training, validation, and test splits with proportions of 70\%, 10\%, and 20\%, respectively. This process treats the optimization as a representation learning task on the target distribution. Empirically, the classifier achieves high Area Under the Curve (AUC) scores peaking at approximately 0.9 on held-out test splits, confirming that the bottleneck features successfully distill validity signals. In this design, the classifier $g_\phi$ serves only as a training time supervision head, helping steer the shared latent space toward evaluation-relevant signals while reducing architecture specific variation. For the subsequent clustering phase, we utilize the MLP output $z_{m,i}$, as this aligned representation preserves richer information about the intrinsic relational structure of the full benchmark.

\subsection{Coreset Selection via Consensus Clustering}
\label{sec:coreset}
To synthesize a stable structural prior from the aligned representations, we compute the consensus embedding $e_i$ by averaging $z_{m,i}$ across all source models.
We use this consensus embedding as the item representation for coreset selection. Subsequently, we apply $L_2$-normalization to project the aggregated vectors onto a unit hypersphere:
\begin{equation}
    \setlength{\abovedisplayskip}{4pt}
    \setlength{\belowdisplayskip}{4pt}
    \setlength{\abovedisplayshortskip}{4pt}
    \setlength{\belowdisplayshortskip}{4pt}
    \tilde{e}_i = \frac{e_i}{\|e_i\|_2}, \quad \text{where} \quad e_i = \frac{1}{|\mathcal{S}|} \sum_{m \in \mathcal{S}} z_{m,i}
\end{equation}
This normalization emphasizes directional agreement and captures the geometric structure of the item space independent of vector magnitude.

Leveraging these normalized consensus embeddings, we partition the item space into $K$ disjoint clusters $\{\mathcal{G}_k\}_{k=1}^{K}$ based on directional similarity. Specifically, we optimize the partition to maximize the within-cluster cosine similarity:
\begin{equation}
    \setlength{\abovedisplayskip}{4pt}
    \setlength{\belowdisplayskip}{4pt}
    \setlength{\abovedisplayshortskip}{4pt}
    \setlength{\belowdisplayshortskip}{4pt}
    \{\mathcal{G}_k\}_{k=1}^{K} \leftarrow \arg\max_{\{\mathcal{G}_k\}} \sum_{k=1}^{K}\sum_{i\in\mathcal{G}_k} \cos(\tilde{e}_i, \mu_k)
\end{equation}
where $\mu_k$ denotes the centroid of the $k$-th cluster. We identify the anchor $c_k$ as the item maximizing the cosine similarity to the centroid:
\begin{equation}
    \setlength{\abovedisplayskip}{4pt}
    \setlength{\belowdisplayskip}{4pt}
    \setlength{\abovedisplayshortskip}{4pt}
    \setlength{\belowdisplayshortskip}{4pt}
    \mathcal{C}=\{c_k\}_{k=1}^{K}
    \quad \text{s.t.} \quad
    c_k=\arg\max_{i\in\mathcal{G}_k}\cos(\tilde{e}_i,\mu_k)
\end{equation}
The resulting coreset $\mathcal{C}$ effectively preserves the global relational structure of the benchmark while acting as a robust basis for the subsequent extrapolation phase.

\subsection{Extrapolation from Coreset to Full Benchmark}
\label{sec:extrapolation}

To evaluate an unseen target model $t \in \mathcal{T}$, we acquire its binary scores $\{y_{t,i}\}_{i \in \mathcal{C}}$ on the selected coreset. We then employ a Ridge regression model $g_t$ to extrapolate these local observations to the full benchmark.

Given the strictly limited coreset size, directly fitting target-specific regressors on full source-response vectors can lead to high-variance estimates.
To improve robustness, we instead construct a scalar feature $x_i$ for each item by averaging correctness across all source models:
\begin{equation}
    \setlength{\abovedisplayskip}{3pt}
    \setlength{\belowdisplayskip}{3pt}
    \setlength{\abovedisplayshortskip}{3pt}
    \setlength{\belowdisplayshortskip}{3pt}
    x_i = \frac{1}{|\mathcal{S}|} \sum_{m \in \mathcal{S}} y_{m,i} \in [0,1], \quad \forall i \in \mathcal{I}
\end{equation}
This averaged score serves as a stable anchor for each item in the source-scarce regime, while target-specific variation is captured by fitting a separate regressor $g_t$ to each model's responses on the selected coreset.
Appendix~\ref{app:extrapolation_feature} compares this scalar feature with an alternative based on the concatenated source response vector and shows that the averaged feature indeed provides more reliable ranking recovery under source scarcity.

We fit the regressor $g_t$ by minimizing the squared error on the coreset:
\begin{equation}
    \setlength{\abovedisplayskip}{3pt}
    \setlength{\belowdisplayskip}{3pt}
    \setlength{\abovedisplayshortskip}{3pt}
    \setlength{\belowdisplayshortskip}{3pt}
    g_t = \arg\min_{g} \frac{1}{|\mathcal{C}|} \sum_{i \in \mathcal{C}} \left(g(x_i) - y_{t,i}\right)^2
\end{equation}
The optimized regressor yields predictions $\hat{s}_{t,i} = g_t(x_i)$ for each unobserved item $i\in\mathcal{I}\setminus C$, providing a robust linear approximation for the unseen items.

Finally, the global accuracy is derived by aggregating the ground-truth performance on $\mathcal{C}$ with the predicted scores for the remaining items in $\mathcal{I} \setminus \mathcal{C}$. This formulation recovers both absolute performance and relative rankings while maintaining minimal computational overhead:
\begin{equation}
    \setlength{\abovedisplayskip}{4pt}
    \setlength{\belowdisplayskip}{4pt}
    \setlength{\abovedisplayshortskip}{4pt}
    \setlength{\belowdisplayshortskip}{4pt}
    \widehat{\mathrm{Acc}}(t) = \frac{1}{|\mathcal{I}|} \left( \sum_{i \in \mathcal{C}} y_{t,i} + \sum_{i \in \mathcal{I} \setminus \mathcal{C}} \hat{s}_{t,i} \right)
\end{equation}

\section{Main Experiments}
\label{sec:Experiment}
 
\begin{table*}[htbp]
    \centering
    \footnotesize
    \setlength{\tabcolsep}{3.5pt}
    \renewcommand{\arraystretch}{1.1}
    \caption{Main results. We report Spearman's rank correlation ($\rho$, $\uparrow$) and Mean Absolute Error (MAE, $\downarrow$) on target models. The budget $K$ denotes the number of items in the coreset. \textbf{Bold} and \uline{underlined} values denote the best and second-best results, respectively.} 
    \label{tab:main_budgetwise}
    \begin{tabular}{l l cc cc cc cc cc}
    \toprule
    \multirow{2}{*}{\textbf{Benchmark}} & \multirow{2}{*}{\textbf{Method}} &
    \multicolumn{2}{c}{\textbf{$K=10$}} & \multicolumn{2}{c}{\textbf{$K=20$}} & \multicolumn{2}{c}{\textbf{$K=30$}} &
    \multicolumn{2}{c}{\textbf{$K=40$}} & \multicolumn{2}{c}{\textbf{$K=50$}} \\
    \cmidrule(lr){3-4}\cmidrule(lr){5-6}\cmidrule(lr){7-8}\cmidrule(lr){9-10}\cmidrule(lr){11-12}
    & & $\rho \uparrow$ & \scriptsize MAE $\downarrow$ & $\rho \uparrow$ & \scriptsize MAE $\downarrow$ & $\rho \uparrow$ & \scriptsize MAE $\downarrow$ & $\rho \uparrow$ & \scriptsize MAE $\downarrow$ & $\rho \uparrow$ & \scriptsize MAE $\downarrow$ \\
    \midrule
    
    \multirow{4}{*}{\textbf{BBH}} 
    & \textsc{Random}       & 0.668 & 0.122 & 0.793 & 0.087 & \uline{0.852} & 0.069 & \uline{0.884} & 0.061 & \uline{0.905} & 0.054 \\
    & \textsc{AnchorPoints} & \uline{0.701} & 0.127 & \uline{0.799} & 0.097 & 0.835 & 0.087 & 0.854 & 0.081 & 0.867 & 0.078 \\
    & \textsc{GP-IRT}       & 0.681 & \textbf{0.093}$^{\star}$ & 0.786 & \textbf{0.068}$^{\star}$ & 0.832 & \uline{0.058} & 0.869 & \uline{0.052} & 0.901 & \textbf{0.045}$^{\star}$ \\
    & \textsc{RepCore}      & \textbf{0.718} & \uline{0.095} & \textbf{0.824} & \uline{0.069} & \textbf{0.870} & \textbf{0.057} & \textbf{0.898} & \textbf{0.049} & \textbf{0.913} & \textbf{0.045} \\
    \midrule
    
    \multirow{4}{*}{\textbf{GSM8K}} 
    & \textsc{Random}       & 0.702 & 0.087 & 0.815 & 0.060 & \uline{0.870} & \uline{0.049} & \uline{0.898} & \uline{0.042} & \uline{0.917} & 0.037 \\
    & \textsc{AnchorPoints} & 0.739 & 0.133 & 0.802 & 0.134 & 0.826 & 0.142 & 0.831 & 0.154 & 0.840 & 0.144 \\
    & \textsc{GP-IRT}       & \uline{0.750} & \textbf{0.069}$^{\star}$ & \uline{0.839} & \uline{0.057} & 0.857 & 0.051 & 0.865 & \uline{0.042} & 0.896 & \uline{0.034} \\
    & \textsc{RepCore}      & \textbf{0.763} & \uline{0.071} & \textbf{0.872} & \textbf{0.047} & \textbf{0.905} & \textbf{0.039} & \textbf{0.926} & \textbf{0.035} & \textbf{0.938} & \textbf{0.032} \\
    \midrule
    
    \multirow{4}{*}{\shortstack[l]{\textbf{SEED-Bench}\\\textbf{-2-Plus}}} 
    & \textsc{Random}       & 0.607 & 0.124 & \uline{0.728} & 0.088 & \uline{0.800} & 0.070 & \uline{0.836} & 0.060 & 0.861 & 0.053 \\
    & \textsc{AnchorPoints} & \textbf{0.653} & 0.139 & 0.715 & 0.116 & 0.747 & 0.110 & 0.771 & 0.108 & 0.782 & 0.105 \\
    & \textsc{GP-IRT}       & 0.594 & \uline{0.098} & 0.699 & \uline{0.063} & 0.780 & \textbf{0.050}$^{\star}$ & 0.824 & \textbf{0.043}$^{\star}$ & \uline{0.863} & \textbf{0.038}$^{\star}$ \\
    & \textsc{RepCore}      & \uline{0.642} & \textbf{0.084} & \textbf{0.755} & \textbf{0.060} & \textbf{0.819} & \uline{0.051} & \textbf{0.853} & \uline{0.045} & \textbf{0.874} & \uline{0.040} \\
    \bottomrule
    \end{tabular}
\end{table*}

\subsection{Experimental Setup}
\label{sec:exp_setup}

\paragraph{Benchmarks and Model Pools}
We evaluate \textsc{RepCore} on five benchmarks spanning text-only and multimodal domains. ARC Challenge \citep{clark2018arc} contains 1{,}172 science QA items evaluated by 96 models. BIG-Bench Hard (BBH) \citep{suzgun2022bbh} includes 6{,}511 challenging reasoning items across 101 models. GSM8K \citep{cobbe2021gsm8k} consists of 1{,}319 grade-school math word problems with 90 models. MMLU-Pro \citep{wang2024mmlupro} contains 12{,}032 multi-domain questions evaluated by 91 models. SEED-Bench-2-Plus \citep{li2024seedbench2plus} is a multimodal benchmark comprising 2{,}277 items and 93 models. A comprehensive list of models for each benchmark is provided in Appendix \ref{sec:model_pools}.

\paragraph{Implementation Details}
We conduct all evaluations in the source-scarce regime, defined by $|\mathcal{S}| = 10$. For each benchmark, we randomly sample 10 models to form the source set $\mathcal{S}$ and designate the remaining models as the target set $\mathcal{T}$. We report results averaged over 10 independent source/target splits. All model generations utilize greedy decoding to ensure reproducibility. To ensure fair comparison, all methods operate on identical splits and coreset sizes. Since \textsc{RepCore} additionally records hidden states during source-model inference, we quantify the resulting wall-clock cost and storage overhead in Appendix~\ref{app:cost_storage}.


\paragraph{Baselines and Metrics}
The main experiments compare \textsc{RepCore} with three primary output-based benchmark compression baselines. \textit{Random} selects the coreset via uniform random sampling. \textit{AnchorPoints} \citep{vivek2024anchor} represents items using their binary response vectors over $\mathcal{S}$ and selects representative anchors via clustering in this discrete space. \textit{tinyBenchmarks (GP-IRT)} \citep{pmlr-v235-maia-polo24a} employs a Gaussian Process-corrected Item Response Theory estimator to infer full-benchmark performance from the coreset evaluations.

To broaden the comparison with recent benchmark compression methods, we additionally report \textit{EffiEval} \citep{wang2025effieval} and \textit{TailoredBench} \citep{yuan2025beyond} as auxiliary baselines. These two methods are evaluated under the same settings as the main experiments, while the subsequent diagnostic analyses focus on the primary baselines that are more directly tied to the controlled factors under study.

For evaluation, we report two primary metrics on the target set $\mathcal{T}$: (i) \textit{Spearman's rank correlation} ($\rho$) between the predicted model rankings and the ground-truth rankings derived from full benchmark accuracy; (ii) \textit{Mean Absolute Error (MAE)} between the predicted and ground truth accuracies.

\subsection{Main Results}
\label{sec:main_results}

We evaluate benchmark compression performance in the source-scarce regime with $|\mathcal{S}|=10$. Table~\ref{tab:main_budgetwise} summarizes the results on BBH, GSM8K, and SEED-Bench-2-Plus. The remaining two benchmarks, ARC Challenge and MMLU-Pro, show consistent trends and are reported in Appendix~\ref{sec:additional_main_results}. We further provide auxiliary comparisons with EffiEval and TailoredBench in Appendix~\ref{app:add_baselines}.

\paragraph{\textsc{RepCore}: Effective Ranking and Reliable Extrapolation under Source Scarcity}
As shown in Table~\ref{tab:main_budgetwise}, \textsc{RepCore} consistently achieves stronger recovery of full-benchmark rankings. By leveraging consensus embeddings to recover the geometry of the item space, our method selects representative items even at small coreset sizes. For example, on GSM8K with $K=20$, \textsc{RepCore} attains a Spearman correlation of 0.872, surpassing \textsc{GP-IRT} at 0.839 and \textsc{AnchorPoints} at 0.802. In terms of absolute performance estimation, \textsc{RepCore} typically yields the lowest mean absolute error and improves steadily as $K$ increases, maintaining a clear margin over baselines. Furthermore, extended analyses in Appendix~\ref{app:stability_analysis} indicate that \textsc{RepCore} exhibits lower variance across experimental runs, demonstrating greater stability compared to baseline methods.

\paragraph{Analysis of Prediction Fidelity}
We observe that \textsc{GP-IRT} occasionally attains competitive MAE scores, marked with $^{\star}$. However, this aggregate metric can mask item-level errors through cancellation. To assess prediction fidelity at the item level, we report the \textit{Agreement} metric following \citet{vivek2024anchor}, which measures the proportion of exact matches between predicted and ground-truth correctness labels. As detailed in Appendix~\ref{sec:agreement_details}, \textsc{GP-IRT} exhibits substantially lower Agreement even when its MAE appears favorable. For instance, on BBH with $K=20$, while \textsc{GP-IRT} matches \textsc{RepCore} in MAE, its Agreement trails ours by approximately 9 percentage points. This gap indicates that \textsc{RepCore} better preserves the target model's item-level correctness pattern.


\section{Analysis of Aligned Representations}
\label{sec:analysis}

\begin{figure*}[htbp]
    \centering
    \begin{subfigure}[htbp]{0.45\linewidth}
        \centering
        \includegraphics[width=\linewidth]{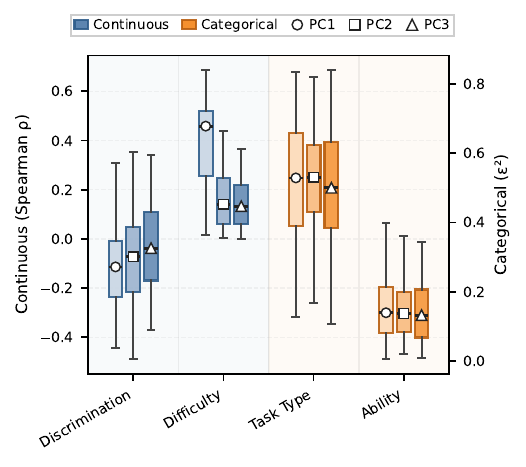}
        \caption{Global associations.}
        \label{fig:global_analyse}
    \end{subfigure}
    \hspace{0.2in}
    \begin{subfigure}[htbp]{0.45\linewidth}
        \centering
        \includegraphics[width=\linewidth]{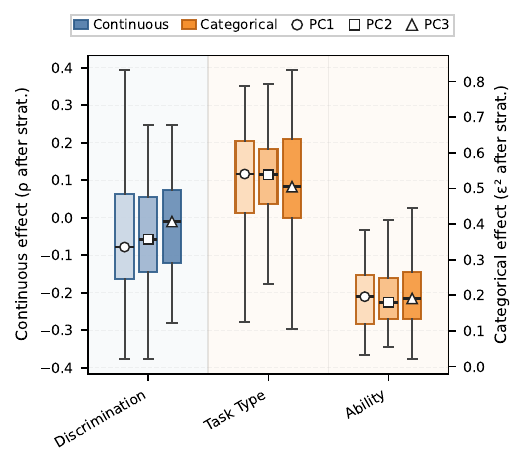}
        \caption{Stratified by difficulty.}
        \label{fig:stratified_analyse}
    \end{subfigure}
    \caption{Factor-association analysis of the \textsc{RepCore} latent space on BBH. Dual axes quantify associations: the left axis reports Spearman's $\rho$ for continuous factors (\textbf{blue}), while the right axis denotes effect size $\epsilon^2$ for categorical factors (\textbf{orange}). Markers indicate principal components (Circle: PC$_1$, Square: PC$_2$, Triangle: PC$_3$).
(a) \textbf{Global Analysis:} The primary axis (PC$_1$) is predominantly aligned with item difficulty ($\rho \approx 0.46$), while categorical task identities also show strong initial associations.
(b) \textbf{Stratified Analysis:} After controlling for difficulty through binning, fine-grained Task Type remains strongly clustered ($\epsilon^2 \approx 0.54$), and this persistence further extends to coarse Ability Mode ($\epsilon^2 \approx 0.20$), confirming that the latent space captures task-specific structure beyond scalar difficulty.}
\label{fig:pc_factor}
    \vskip -0.15in
\end{figure*}

While the previous sections demonstrate the effectiveness of \textsc{RepCore} in benchmark compression, a critical question remains regarding the interpretability of the learned representations: \textit{Does the latent space merely act as a proxy for item difficulty, or does it capture richer, task-specific structural information?} To investigate this question, we analyze the geometric structure of the latent space in Section~\ref{sec:structure} and evaluate the utility of its principal components in Section~\ref{sec:utility}.

\subsection{Geometric Definition and Statistical Protocol}
\label{sec:analysis_protocol}

We conduct our primary analysis on benchmark \textbf{Big-Bench Hard} to leverage its hierarchical taxonomy, which divides items into fine-grained \texttt{subtasks} and broader \texttt{ability} modes (see Appendix~\ref{sec:bbh_taxonomy_mapping} for detailed category definitions). This structure enables us to probe the granularity of the learned item representations.

\textbf{Analysis Protocol}
We analyze the geometric properties of aligned embeddings across 100 model snapshots (10 combos $\times$ 10 models, aligned with the main experiment). The protocol proceeds in three steps:

\paragraph{1. Embedding Projection and Decomposition}
For each model snapshot $m$ and item $i$, we first compute the aligned embedding $z_{m,i} = f_\theta (Proj_m(h_{m,i})) \in \mathbb{R}^{32}$. These vectors are then aggregated over all items to form the embedding matrix $Z_m \in \mathbb{R}^{|\mathcal{I}| \times 32}$, and we denote the empirical mean of these embeddings as $\bar{z}_m$. Next, we decompose the centered data via PCA to derive the $k$-th principal axis $w^{(k)}_m$. Finally, the projection score $\xi^{(k)}_{m,i}$ is computed by projecting the centered embedding onto the $k$-th principal axis: $\xi^{(k)}_{m,i} = \langle z_{m,i} - \bar{z}_m, w^{(k)}_m \rangle$.
Averaged over the analyzed BBH snapshots, the top three principal components explain 50.12\%, 16.21\%, and 9.21\% of the variance, respectively.

\paragraph{2. Interpretable Factors}
To investigate whether the learned representation recovers the intrinsic structural relationships among items in the benchmark, we evaluate the alignment between the projection scores $\xi_{m,i}^{(k)}$ and four ground-truth attributes. Detailed computational formulations and definitions for these attributes are provided in Appendix~\ref{sec:bbh_taxonomy_mapping} and \ref{sec:continuous_factors}:
\begin{itemize}[leftmargin=*, nosep, itemsep=2pt, before=\raggedright]
    \item \textbf{Continuous Factors:} Difficulty $\delta_i$ (pass rate) and Discrimination $\gamma_i$ (Kelley 27\% $D$).
    \item \textbf{Discrete Factors:} Subtask Identity $\tau_i$ (fine-grained) and Ability Mode $\alpha_i$ (coarse-grained).
\end{itemize}

\paragraph{3. Statistical Association} 
To rigorously quantify the alignment between distinct principal components and the item attributes defined above, we propose a dual-granularity evaluation protocol centered on a unified association operator $\mathcal{M}(\cdot)$. This operator is realized as Spearman’s $\rho$ for continuous attributes ($\delta, \gamma$) and Kruskal-Wallis $\epsilon^2$ for categorical ones ($\tau, \alpha$), thereby providing a consistent measure to evaluate the representational geometry across two distinct granularities: 
\begin{itemize}[leftmargin=*, nosep]
    \setlength{\abovedisplayskip}{4pt}
    \setlength{\belowdisplayskip}{4pt}
    \setlength{\abovedisplayshortskip}{4pt}
    \setlength{\belowdisplayshortskip}{4pt}

    \item \textbf{Global Analysis:} For a tested item attribute 
$\psi \in \{\delta,\gamma,\tau,\alpha\}$, let $\psi_i$ denote its value for item $i$. The macro-scale association is evaluated over the entire benchmark $\mathcal{I}$ to capture general trends:
    \begin{equation}
        \mathcal{A}_{\text{global}}^{(k,m)} = \mathcal{M}\left( \{ \xi_{m,i}^{(k)} \}_{i \in \mathcal{I}}, \{ \psi_i \}_{i \in \mathcal{I}} \right)
    \end{equation}
    
    \item \textbf{Stratified Analysis:} To isolate structural signals from the confounding effects of item difficulty, we partition $\mathcal{I}$ into $B=20$ equi-sized bins $\{\mathcal{I}_b\}_{b=1}^B$ based on $\delta$. Local statistics are then aggregated for the remaining attributes to derive a difficulty-invariant estimate:
    \begin{equation}
        \scalebox{0.9}{$
        \mathcal{A}_{\text{strat}}^{(k,m)} = \text{Agg} \left( \left\{ \mathcal{M} \left( \{ \xi_{m,i}^{(k)} \}_{i \in \mathcal{I}_b}, \{ \psi_i \}_{i \in \mathcal{I}_b} \right) \right\}_{b=1}^{B} \right)
        $}
    \end{equation}
\end{itemize}
where Agg$(\cdot)$ aggregates $p$-values using Stouffer's method and
effect sizes using statistic-specific procedures, including Fisher-z
transformation for Spearman correlations and sample-size-weighted averaging
for Kruskal-Wallis $\epsilon^2$. Full computational details are deferred to Appendix~\ref{sec:agg_operator}. 

To mitigate sampling variance, we aggregate these metrics over the full snapshot pool. Figure~\ref{fig:pc_factor} summarizes the resulting distributions via medians and interquartile ranges, offering a robust characterization of the representational geometry. Quantitatively, this characterization is statistically rigorous: both stratified $p$-values and FDR-corrected $q$-values consistently satisfy significance thresholds ($\alpha=0.05$) across snapshots, with medians often approaching floating-point limits (see Appendix ~\ref{sec:stat_dist} for full results).  


\subsection{Disentangling Difficulty and Structure}
\label{sec:structure}

\paragraph{Global View: The Dominance of Difficulty}
Globally (Figure~\ref{fig:global_analyse}), PC$_1$ emerges as the dominant axis of variation, manifesting a strong correlation with item difficulty ($\rho \approx 0.46$). While categorical factors such as \texttt{subtask type} exhibit substantial effect sizes ($\epsilon^2 \approx 0.53$), we observe that this alignment is potentially susceptible to difficulty confounding. Specifically, if items within the same subtask share homogeneous complexity levels, global clustering may be a misleading artifact of difficulty rather than a reflection of task-specific reasoning patterns. This possibility necessitates a more granular investigation to adjudicate whether the captured signals are truly structural.

\paragraph{Stratified View: Hierarchical Structural Persistence}
To verify the persistence of structural identity, we move to a difficulty-stratified analysis (Figure~\ref{fig:stratified_analyse}), which serves as a controlled setting to isolate task-specific signals from the aforementioned confounding effects. The results indicate that even within narrow difficulty bins, clustering by \texttt{subtask type} remains remarkably robust ($\epsilon^2 \approx 0.54$). This resilience confirms that the aligned representations $z_{m,i}$ encode fine-grained functional identities that transcend mere performance-based scalar signals. Moreover, this persistence further scales to the benchmark’s hierarchy, with the \texttt{ability mode} alignment ($\epsilon^2 \approx 0.20$) confirming the recovery of abstract reasoning domains. These results suggest that the latent manifold preserves high-level cognitive modalities beyond granular task labels.

\paragraph{The Role of Discrimination}
In contrast to the dominant structural geometry, item discrimination acts as a secondary background signal, maintaining a modest yet stable negative correlation with PC$_1$ across both global ($\rho \approx -0.11$) and stratified ($\rho \approx -0.08$) regimes. This suggests that while the bottleneck primarily optimizes for difficulty and structural features, it retains auxiliary information regarding the reliability of the item representations.

\begin{table}[htbp]
\centering
\caption{Principal component ablation on BBH for performance extrapolation across \ensuremath{K \in \{10,20,30,40,50\}}.}
\label{tab:svdpc}
\setlength{\tabcolsep}{4pt}
\vskip -0.08in
\begin{small}
\begin{sc}
\begin{tabular}{lccccc}
\toprule
$\mathbf{\rho} \uparrow$ & $K=10$ & $K=20$ & $K=30$ & $K=40$ & $K=50$ \\
\midrule
PC$_1$       & 0.665 & 0.793 & 0.857 & 0.888 & 0.907 \\
PC$_2$       & 0.662 & 0.794 & 0.853 & 0.887 & 0.907 \\
PC$_3$       & \underline{0.689} & 0.812 & 0.859 & 0.882 & 0.900 \\
PC$_{1,2}$   & \underline{0.689} & 0.812 & \underline{0.871} & \textbf{0.898} & \textbf{0.917} \\
PC$_{1,2,3}$ & 0.686 & \underline{0.820} & \textbf{0.875} & \underline{0.896} & \underline{0.914} \\
All  & \textbf{0.718} & \textbf{0.824} & 0.870 & \textbf{0.898} & 0.913 \\
\midrule
\scriptsize \textbf{MAE} $\downarrow$ & $K=10$ & $K=20$ & $K=30$ & $K=40$ & $K=50$ \\
\midrule
PC$_1$       & \underline{0.101} & 0.074 & 0.060 & 0.053 & 0.048 \\
PC$_2$       & \underline{0.101} & \underline{0.070} & \underline{0.058} & \underline{0.050} & \textbf{0.045} \\
PC$_3$       & 0.103 & 0.072 & 0.060 & 0.052 & 0.047 \\
PC$_{1,2}$   & 0.104 & 0.072 & 0.059 & 0.051 & \textbf{0.045} \\
PC$_{1,2,3}$ & 0.104 & 0.073 & \underline{0.058} & 0.051 & \underline{0.046} \\
All  & \textbf{0.095} & \textbf{0.069} & \textbf{0.057} & \textbf{0.049} & \textbf{0.045} \\
\bottomrule
\end{tabular}
\end{sc}
\end{small}
\vskip -0.15in
\end{table}

\subsection{Utility of the Interpretable Subspace}
\label{sec:utility}

Building on the structural insights from the spectral analysis, we further evaluate how these interpretable dimensions contribute to the core task of performance extrapolation. Table~\ref{tab:svdpc} presents the ablation results on the BBH benchmark, where we compare the predictive efficacy of individual principal components and low-rank subspaces against the full aligned representations across varying coreset budgets.

\paragraph{Beyond Scalar Difficulty} 
While PC$_1$ provides a robust predictive baseline by capturing item difficulty, the inclusion of PC$_2$ and PC$_3$ consistently enhances estimation fidelity. This improvement reinforces our geometric findings: higher-order components capture essential structural context including task identity and reasoning modalities. By leveraging these signals, the regressor can adjust for model-specific capabilities that a standard difficulty curve cannot account for.

\paragraph{Robustness and Signal Concentration} 
A comparison with the full aligned embedding space reveals a distinct performance crossover. In extremely data-scarce regimes where $K=10$, the full aligned embedding slightly outperforms the PC$_{1,2}$ subspace. However, as the coreset size grows to $K \ge 30$, the PC$_{1,2}$ subspace demonstrates superior robustness and eventually surpasses the full aligned embedding space. This shift suggests that while high-dimensional features may offer some utility at minimal sample sizes, they also introduce predictive instability. Conversely, the low-rank subspace concentrates the most generalizable signals to provide a stable basis for extrapolation without relying on the noise inherent in the high-dimensional manifold.

\begin{table*}[htbp]
    \centering
    \setlength{\tabcolsep}{10pt}
    \caption{Performance averaged over five benchmarks for varying source pool sizes with $K=30$.}
    \label{tab:ablation_quantity}
    \begin{tabular*}{0.85\textwidth}{@{\extracolsep{\fill}} l cc cc cc cc @{}}
    \toprule
    \multirow{2}{*}{\textbf{Method}} & \multicolumn{2}{c}{\textbf{$|\mathcal{S}| = 5$}} & \multicolumn{2}{c}{\textbf{$|\mathcal{S}| = 10$}} & \multicolumn{2}{c}{\textbf{$|\mathcal{S}| = 15$}} & \multicolumn{2}{c}{\textbf{$|\mathcal{S}| = 20$}} \\
    \cmidrule(lr){2-3} \cmidrule(lr){4-5} \cmidrule(lr){6-7} \cmidrule(lr){8-9}
    & $\rho \uparrow$ & \scriptsize \textbf{MAE} $\downarrow$ & $\rho \uparrow$ & \scriptsize \textbf{MAE} $\downarrow$ & $\rho \uparrow$ & \scriptsize \textbf{MAE} $\downarrow$ & $\rho \uparrow$ & \scriptsize \textbf{MAE} $\downarrow$ \\
    \midrule
    \textsc{AnchorPoints} & 0.640 & 0.210 & 0.809 & 0.116 & \uline{0.848} & 0.091 & \uline{0.861} & 0.080 \\
    \textsc{GP-IRT}       & \uline{0.822} & \uline{0.057} & \uline{0.836} & \uline{0.053} & 0.843 & \textbf{0.049} & 0.828 & \uline{0.050} \\
    \textbf{\textsc{RepCore}} & \textbf{0.860} & \textbf{0.052} & \textbf{0.868} & \textbf{0.049} & \textbf{0.860} & \textbf{0.049} & \textbf{0.864} & \textbf{0.049} \\
    \bottomrule
    \end{tabular*}
    \vskip -0.15in
\end{table*}

\section{Ablations and Additional Analyses}  
\label{sec:detailed_analysis}




\paragraph{Continuous Representations Enable Superior Coreset Selection}
To validate the necessity of our representation learning module, we conduct a controlled experiment by substituting the consensus embeddings with binary correctness vectors during the clustering phase, while maintaining the identical downstream extrapolation mechanism. Table~\ref{tab:ablation_component} reports the average performance with $K=30$. The results indicate that relying on discrete correctness signals yields inferior ranking correlation and higher estimation error. This performance gap confirms that the continuous embedding space preserves fine-grained item distinctions essential for representative sampling, which are not captured by binary correctness signals. Detailed per-dataset results are provided in Appendix~\ref{sec:discrete_ablation}. 

\begingroup
\setlength{\abovecaptionskip}{5pt}
\setlength{\belowcaptionskip}{3pt}
\begin{center}
\setlength{\tabcolsep}{8pt}
\captionof{table}{Ablation on clustering inputs averaged over five benchmarks with $K=30$.}
\label{tab:ablation_component}
\begin{tabular}{l c c}
\toprule
\textbf{Clustering Input} & \textbf{$\rho \uparrow$} & \scriptsize \textbf{MAE} $\downarrow$ \\
\midrule
Correctness Vectors & 0.850 & 0.055 \\
\textbf{Consensus Embeddings} & \textbf{0.868} & \textbf{0.049} \\
\bottomrule
\end{tabular}
\vskip -0.16in
\end{center}
\endgroup


\paragraph{Aligned Representations Are Data-Efficient in Source-Scarce Regimes}
We evaluate the data efficiency of our framework by varying the source pool size $|\mathcal{S}|$. Table~\ref{tab:ablation_quantity} presents the performance metrics as the number of source models increases from 5 to 20 with $K=30$. \textsc{RepCore} demonstrates strong stability even in highly resource-constrained scenarios. Under the most restricted setting with only five source models, output-based methods, including \textsc{AnchorPoints}, exhibit a substantial performance drop. In contrast, \textsc{RepCore} achieves a Spearman correlation of 0.860, which significantly outperforms baselines. Furthermore, the performance of our framework stabilizes rapidly as the source pool expands. This saturation indicates that the aligned space effectively recovers the geometry of the item space under minimal supervision, thereby reducing the dependency on extensive source model pools. Detailed per-dataset results are provided in Appendix~\ref{sec:source_num_ablation}.
\vskip -0.2in
\paragraph{Impact of Source Model Composition}
We investigate how source pool diversity affects compression quality. Table~\ref{tab:source_composition} indicates that a diverse composition of model families and capability ranges yields the most robust performance. Relying on a single model family introduces architectural biases, as evidenced by the performance drop in single-family pools. Similarly, homogeneous pools restricted to a narrow capability range prove suboptimal, as they fail to simultaneously maintain high ranking correlation and low estimation error. These findings suggest that incorporating diverse sources mitigates architectural and capability biases, allowing the aligned representations to better capture the geometry of the item space that supports reliable extrapolation. Detailed per-dataset results are provided in Appendix~\ref{sec:family_ablation} and Appendix~\ref{sec:capability_ablation}.

\begingroup
\setlength{\abovecaptionskip}{5pt}
\setlength{\belowcaptionskip}{3pt}
\begin{center}
\setlength{\tabcolsep}{6pt}
\vskip -0.05in
\captionof{table}{Performance comparison under varied source compositions averaged across five benchmarks at $K=30$.}
\label{tab:source_composition}
\begin{tabular}{l l cc}
\toprule
\textbf{Composition} & \textbf{Setting} & \textbf{$\rho \uparrow$} & \textbf{\scriptsize MAE $\downarrow$} \\
\midrule
\multirow{3}{*}{\textit{Model Family}}
& Family A$^{\dag}$ & \uline{0.848} & \uline{0.057} \\
& Family B$^{\ddag}$ & 0.686 & 0.058 \\
& Diverse (Ours) & \textbf{0.868} & \textbf{0.049} \\
\midrule
\multirow{3}{*}{\textit{Capability}}
& Strong Only & 0.778 & 0.092 \\
& Weak Only & \uline{0.799} & \uline{0.055} \\
& Diverse (Ours) & \textbf{0.868} & \textbf{0.049} \\
\bottomrule
\multicolumn{4}{l}{\scriptsize $^{\dag}$ Qwen / Ovis; $^{\ddag}$ Llama / InternVL.} \\
\end{tabular}
\vskip -0.20in
\end{center}
\endgroup


\paragraph{Scalability to Larger Coreset Budgets}
We further examine scalability by increasing the coreset budget to $K=100$, 150, and 200. As shown in Table~\ref{tab:large_budget}, \textsc{RepCore} improves steadily as $K$ increases and maintains strong ranking and estimation accuracy relative to baselines. These results indicate that the learned item space continues to support effective coreset selection at larger budgets. Detailed per-dataset results are provided in Appendix~\ref{app:coreset_size_generalization}.

\begingroup
\setlength{\abovecaptionskip}{5pt}
\setlength{\belowcaptionskip}{3pt}
\begin{center}
    \small
    \renewcommand{\arraystretch}{1.3}
    \setlength{\tabcolsep}{4pt}
    \captionof{table}{Average performance across benchmarks with larger coreset budgets.}
    \label{tab:large_budget}
    \begin{tabular}{l cc cc cc}
    \toprule
    \multirow{2}{*}{\textbf{Method}} & \multicolumn{2}{c}{\textbf{$K=100$}} & \multicolumn{2}{c}{\textbf{$K=150$}} & \multicolumn{2}{c}{\textbf{$K=200$}} \\
    \cmidrule(lr){2-3} \cmidrule(lr){4-5} \cmidrule(lr){6-7}
    & $\rho \uparrow$ & \scriptsize \textbf{MAE} $\downarrow$ & $\rho \uparrow$ & \scriptsize \textbf{MAE} $\downarrow$ & $\rho \uparrow$ & \scriptsize \textbf{MAE} $\downarrow$ \\
    \midrule
    \textsc{Random}       & 0.947 & 0.033 & 0.964 & 0.027 & 0.972 & 0.023 \\
    \hdashline
    \makecell[l]{\textsc{Anchor}\\\textsc{Points}} & 0.833 & 0.111 & 0.832 & 0.107 & 0.836 & 0.099 \\
    \hdashline
    \textsc{GP-IRT}       & \uline{0.951} & \textbf{0.027} & \uline{0.966} & \textbf{0.023} & \uline{0.973} & \textbf{0.021} \\
    \hdashline
    \textsc{RepCore}      & \textbf{0.954} & \uline{0.028} & \textbf{0.967} & \textbf{0.023} & \textbf{0.975} & \textbf{0.021} \\
    \bottomrule
    \end{tabular}
    \vskip -0.18in
\end{center}
\endgroup


\paragraph{Robustness to Temporal Distribution Shift}
Rapid progress in LLMs induces temporal shifts in model behaviors. To simulate leaderboard-style evaluation, we sort models by their release time and use the oldest 30 models as the historical source candidate pool. For each benchmark, we conduct 10 trials, each sampling 10 source models from this pool and holding out all remaining models as newer targets. As shown in Table~\ref{tab:temporal_shift}, \textsc{RepCore} achieves the strongest ranking performance with competitive estimation error, suggesting that the learned item space remains effective as the model pool evolves. Detailed per-dataset results are provided in Appendix~\ref{sec:temporal_drift}. 

\begingroup
\setlength{\abovecaptionskip}{5pt}
\setlength{\belowcaptionskip}{3pt}
\begin{center}
    \vskip -0.08in
    \setlength{\tabcolsep}{8pt}
    \captionof{table}{Temporal generalization under a chronological split at $K=30$, averaged over 10 trials across five benchmarks.}
    \label{tab:temporal_shift}
    \vskip -0.05in
    \begin{tabular}{l cc}
    \toprule
    \textbf{Method} & \textbf{$\rho \uparrow$} & \textbf{\scriptsize MAE $\downarrow$} \\
    \midrule
    \textsc{Random} & \uline{0.845} & 0.060 \\
    \textsc{AnchorPoints} & 0.797 & 0.110 \\
    \textsc{GP-IRT} & 0.835 & \textbf{0.049} \\
    \textbf{\textsc{RepCore}} & \textbf{0.853} & \uline{0.050} \\
    \bottomrule
    \end{tabular}
    \vskip -0.20in
\end{center}
\endgroup

\section{Discussion of Method Boundaries}
\label{sec:scope_boundary}

\subsection{Dependence on Hidden States}
\label{sec:limitations}

\begin{table}[htbp]
\centering
\caption{Masking results under partial source hidden-state unavailability.
Results are averaged over five benchmarks and five coreset budgets, with output-based baselines included for reference.}
\label{tab:closed_source_masking}
\small
\setlength{\tabcolsep}{5pt}
\renewcommand{\arraystretch}{1.05}
\begin{tabular*}{\columnwidth}{@{\extracolsep{\fill}}lcccc@{}}
\toprule
\multirow{2}{*}{\textbf{Masked Sources}} 
& \multicolumn{2}{c}{\textbf{Random Mask}} 
& \multicolumn{2}{c}{\textbf{Advdist Mask}} \\
\cmidrule(lr){2-3}\cmidrule(lr){4-5}
& $\rho \uparrow$ & MAE $\downarrow$ & $\rho \uparrow$ & MAE $\downarrow$ \\
\midrule
0\%  & 0.842 & 0.054 & \multicolumn{2}{c}{--} \\
10\% & 0.842 & 0.058 & 0.840 & 0.058 \\
20\% & 0.839 & 0.058 & 0.836 & 0.060 \\
30\% & 0.837 & 0.058 & 0.828 & 0.061 \\
40\% & 0.837 & 0.059 & 0.820 & 0.062 \\
50\% & 0.833 & 0.059 & 0.805 & 0.065 \\
\midrule
Random Subset & 0.818 & 0.070 & \multicolumn{2}{c}{--} \\
AnchorPoints & 0.791 & 0.120 & \multicolumn{2}{c}{--} \\
tinyBenchmarks (GP-IRT) & 0.811 & 0.057 & \multicolumn{2}{c}{--} \\
TailoredBench & 0.747 & 0.075 & \multicolumn{2}{c}{--} \\
EffiEval & 0.762 & 0.111 & \multicolumn{2}{c}{--} \\
\bottomrule
\end{tabular*}
\end{table}

\textsc{RepCore} assumes access to hidden states of the source models used for coreset construction. This restricts the direct use of closed-source or API-only models as representation sources, although their output responses can still be used. To further probe this boundary, we simulate partially accessible source pools by masking hidden states for a subset of sources while retaining their benchmark responses. For masked sources, we approximate item representations from accessible sources with similar output behavior and use the resulting pseudo-representations for coreset selection. 

Table~\ref{tab:closed_source_masking} reports two masking methods. Random Mask removes source hidden states uniformly at random, while Advdist Mask provides a harder stress test by first masking sources that are most distinct from the rest according to their output correctness vectors. The pseudo-representation construction and the distinctness score used by Advdist Mask are detailed in Appendix~\ref{app:closed_source_masking}.

Under random masking, performance degrades mildly and remains above output-based baselines in ranking correlation even when 50\% of hidden states are unavailable. The Advdist setting reveals a clearer boundary: the advantage narrows when the inaccessible sources provide distinctive behavioral coverage. Overall, \textsc{RepCore} is robust to random or redundant missingness, but benefits most when hidden states are available for behaviorally diverse sources.

\subsection{Beyond Binary-Scored Benchmarks}
\label{sec:extensions}
In this paper, we primarily  focus on binary-scored benchmarks, where each item is parsed
into a correctness label and the alignment objective uses cross-entropy loss.
This scope matches common free-generation QA evaluation protocols, but excludes
settings with continuous item-level outcomes, such as graded scores or
reward-model judgments. This is an empirical scope limitation rather than a
fundamental restriction of \textsc{RepCore}. Since coreset selection relies on
hidden state geometry, it does not inherently require binary labels. For
continuous metrics, Eq.~(2) can be replaced with a regression loss such as MSE,
and the target-specific extrapolator can be
trained to predict continuous item-level scores instead of binary correctness.

\section{Conclusion and Future Work}
\label{sec:conclusion}

In this work, we introduce \textsc{RepCore}, a framework leveraging aligned representations to identify representative benchmark subsets. Our evaluation confirms that \textsc{RepCore} significantly outperforms output-based baselines in ranking correlation and estimation accuracy, while maintaining robust performance under source model scarcity and temporal distribution shifts. For future directions, we aim to extend this paradigm to closed-source models, thereby establishing a unified standard for open and proprietary systems. Furthermore, we aim to extend \textsc{RepCore} to more general benchmark settings beyond binary-scored evaluation and explore the cross-domain transferability of aligned representations, enabling efficient coreset
selection for broader evaluation protocols and specialized domains.

\newpage
\section*{Acknowledgements}
This work is supported by the Beijing Natural Science Foundation (Grant Nos. 4262065, 4222037, and L181010).
\section*{Impact Statement}
This paper presents work whose goal is to advance the field of Machine
Learning. There are many potential societal consequences of our work, none of which we feel must be specifically highlighted here.
\bibliography{example_paper}
\bibliographystyle{icml2026}

\newpage
\appendix
\onecolumn

\section{Related Work}
\label{sec:Related_Work}
 
\subsection{Benchmark Compression}
\textit{Benchmark Compression} aims to select a minimal subset of items that preserves full-benchmark rankings and enables accurate score estimation. Existing methods can be categorized based on the signals they utilize: prediction outputs versus input/internal features.

\paragraph{Prediction-Based Coreset Selection.}
The dominant paradigm relies on the correlation of model outputs (i.e., accuracy) across test items. \textsc{AnchorPoints}~\cite{vivek2024anchor} and \textsc{TailoredBench}~\cite{yuan2025beyond} represent each item by its ``prediction pattern''—a vector of correctness scores derived from a population of source models. By applying clustering to these patterns, they identify items that maximize information gain. Similarly, \textsc{TinyBenchmarks}~\cite{pmlr-v235-maia-polo24a} adopts Item Response Theory (IRT) to fit latent discrimination and difficulty parameters from response patterns. A critical limitation of these methods is the ``cold start'' problem: they require a diverse pool of source models to construct robust prediction patterns. In low-resource regimes (e.g., few source models), the sparse binary signal (correct/incorrect) is often insufficient to capture fine-grained dataset structures.

\paragraph{Feature-Based and Heuristic Filtering.}
Alternative approaches utilize static semantic features or internal model statistics. \textsc{SMART}~\cite{jain2024improving} filters datasets by removing samples that are semantically redundant (measured via textual embedding distance) or statistically ``easy'' for models. More recently, \textsc{EffiEval}~\cite{wang2025effieval} proposes a training-free selection method that maximizes the coverage of activated neurons, positing that diverse activation patterns correspond to diverse model capabilities. While \textsc{EffiEval} leverages internal signals, it focuses on neuron-level coverage for sampling rather than aligning latent spaces for structural recovery. Our work advances this direction by exploiting the continuous hidden states to characterize item structure, providing a denser signal than discrete outputs while avoiding the need for massive source model populations.

\subsection{Internal Representations and Difficulty Perception}
Beyond their use in compression, hidden states have been extensively studied for their ability to encode task-level properties.

\paragraph{Probing for Factuality.}
Representation engineering has shown that LLM hidden states contain linearly separable directions corresponding to truthfulness. Methods like SAPLMA~\cite{azaria2023internal} and \textsc{FacLens}~\cite{wang-etal-2025-faclens} train probes to detect hallucinations or predict correctness before generation is complete, suggesting models possess intrinsic meta-cognition regarding their knowledge boundaries~\cite{ma2025large,orgad2025llms}.

\paragraph{Encoding Difficulty and Structure.}
Recent findings indicate that hidden representations also implicitly encode the complexity of input queries. \citet{zhu2025llm} and \citet{garg2024reasoning} demonstrate that the difficulty of a question can be estimated directly from its hidden states or attention patterns, often providing a more reliable signal of model-specific difficulty than the model's own explicit judgments. Unlike prior works that use these signals primarily for difficulty estimation or inference-time intervention~\cite{li2024inference}, we propose \textsc{RepCore} to leverage these high-dimensional, density-rich signals for benchmark compression. By treating hidden states as a continuous proxy for item characteristics, we can recover dataset structures more effectively than methods relying on sparse discrete labels.

\section{Methodological Details}
\subsection{Aggregation Operator for Stratified Associations}
\label{sec:agg_operator}

This section formalizes the aggregation operator $\text{Agg}(\cdot)$ used in the stratified analysis protocol
(Section~\ref{sec:analysis_protocol}).
Let $\{\mathcal{I}_b\}_{b=1}^{B}$ denote the difficulty-stratified partition of the benchmark item set $\mathcal{I}$.
For each bin $\mathcal{I}_b$, the within-bin procedure $\mathcal{M}(\cdot)$ produces a pair
$(\eta_b, p_b)$, where $\eta_b$ is an effect-size statistic and $p_b$ is the corresponding $p$-value (bins with undefined outputs are excluded from aggregation).
Let $n_b$ be the number of valid items contributing to $\mathcal{M}(\cdot)$ in bin $b$.
To account for varying bin sample sizes in the aggregation, we assign each bin the weight $w_b = \sqrt{n_b}$.

\paragraph{Aggregation of $p$-values (weighted Stouffer)}
Let $p_b$ denote the $p$-value computed on bin $b$. We convert $p_b$ to a normal score $s_b$ by
\begin{equation}
s_b=
\begin{cases}
\Phi^{-1}(1-p_b), & \text{Kruskal--Wallis (one-sided)},\\[2pt]
\Phi^{-1}\!\left(1-\frac{p_b}{2}\right), & \text{Spearman (two-sided)}.
\end{cases}
\end{equation}
The weighted Stouffer statistic is
\begin{equation}
Z_{\mathrm{st}}=\frac{\sum_b w_b s_b}{\sqrt{\sum_b w_b^2}}.
\end{equation}
Accordingly, the stratified $p$-value is
\begin{equation}
p_{\text{strat}}=
\begin{cases}
1-\Phi(Z_{\mathrm{st}}), & \text{Kruskal--Wallis},\\[2pt]
2\!\left(1-\Phi(|Z_{\mathrm{st}}|)\right), & \text{Spearman},
\end{cases}
\end{equation}
where $\Phi(\cdot)$ denotes the standard normal CDF.

\paragraph{Aggregation of effect sizes}
Effect sizes are aggregated separately for continuous and discrete attributes.

\begin{itemize}[leftmargin=*, nosep, itemsep=2pt, before=\raggedright]
    \item \textbf{Continuous attributes (Spearman).}
    Let $\rho_b$ denote the Spearman correlation computed on bin $b$.
    Correlations are combined via the Fisher transformation with inverse-variance weights:
    \begin{equation}
        \zeta
        = \frac{\sum_b (n_b - 3)\,\operatorname{arctanh}(\rho_b)}{\sum_b (n_b - 3)},
    \end{equation}
    \begin{equation}
        \rho_c = \tanh(\zeta),
    \end{equation}
    where $(n_b-3)$ is the standard inverse-variance weight under Fisher's transformation.
    The aggregated correlation $\rho_c$ is reported as the stratified effect size.

    \item \textbf{Discrete attributes (Kruskal--Wallis).}
    Let $H_b$ denote the Kruskal--Wallis statistic computed on bin $b$,
    using $n_b$ valid items across $g_b$ retained groups.
    The within-bin effect size is quantified by epsilon-squared:
    \begin{equation}
        \epsilon_b^2 = \max\!\left\{0,\; \frac{H_b - g_b + 1}{n_b - g_b}\right\}.
    \end{equation}
    The stratified effect size is the mean of within-bin estimates, weighted by their respective sample sizes $n_b$:
    \begin{equation}
        \epsilon_{\text{strat}}^2
        = \frac{\sum_b n_b\,\epsilon_b^2}{\sum_b n_b}.
    \end{equation}
\end{itemize}

\paragraph{Multiple testing control}
Let $\Psi$ denote the set of tested item attributes.
For each fixed principal component index $k$, we apply Benjamini--Hochberg FDR correction to the collection of
stratified $p$-values $\{p_{\text{strat}}(\psi)\}_{\psi\in\Psi}$.  
The resulting per-component adjusted values are reported as $q_{\text{strat,pcwise}}$. 

\subsection{Detailed Statistical Distributions of Aligned Representations}
\label{sec:stat_dist}

To complement the summary statistics provided in the main text, we report the
detailed distribution of attribute-aggregated significance metrics across the
100 model snapshots. For each principal component, global significance statistics
are aggregated over all tested item attributes
$\psi \in \{\delta,\gamma,\tau,\alpha\}$. For stratified statistics, difficulty
$\delta$ is used only to construct the bins and is therefore excluded from the
attribute aggregation; the reported stratified metrics are aggregated over
$\psi \in \{\gamma,\tau,\alpha\}$. Table~\ref{tab:p_value_dist} presents the
median, 25th percentile, and 75th percentile of the raw global $p$-values and
the stratified significance metrics, including both raw $p$-values and
FDR-corrected $q$-values ($q_{\mathrm{strat}}$).

As observed, the associations are overwhelmingly significant. For the difficulty-stratified analysis, the median $q_{\text{strat}}$ values for the first three principal components (PC$_1$--PC$_3$) are consistently below $10^{-190}$, approaching the numerical precision limits of standard floating-point representations. These results confirm that the geometric alignment captured by \textsc{RepCore} is statistically robust and not an artifact of specific source model sampling.

\begin{table}[h]
\centering
\caption{Distribution of attribute-aggregated statistical significance metrics.
We report the empirical distribution (median, $25^{\mathrm{th}}$, and
$75^{\mathrm{th}}$ percentiles) of significance tests across 100 model snapshots.
Global metrics are aggregated over all tested item attributes
$\psi \in \{\delta,\gamma,\tau,\alpha\}$ and reported as raw $p$-values from
the corresponding association tests. For stratified metrics, difficulty
$\delta$ is used only for binning, the aggregation is performed over
$\psi \in \{\gamma,\tau,\alpha\}$, and both raw $p$-values and FDR-corrected
$q$-values ($q_{\text{strat}}$) are reported. Note that values displayed as $0$ indicate probabilities below the numerical precision limit (underflow).}
\label{tab:p_value_dist}
\vskip 0.15in
\begin{small}
\resizebox{\textwidth}{!}{
\begin{tabular}{llccccc}
\toprule
\textbf{Mode} & \textbf{Component} & \textbf{Metric} & \textbf{Median} & \textbf{25\% Percentile} & \textbf{75\% Percentile} & \textbf{IQR} \\
\midrule
\multirow{3}{*}{Global}
& PC1 & $p$ & $4.16 \times 10^{-249}$ & $0$ & $6.96 \times 10^{-70}$ & $6.96 \times 10^{-70}$ \\
& PC2 & $p$ & $2.07 \times 10^{-129}$ & $0$ & $2.02 \times 10^{-30}$ & $2.02 \times 10^{-30}$ \\
& PC3 & $p$ & $7.46 \times 10^{-109}$ & $0$ & $3.96 \times 10^{-26}$ & $3.96 \times 10^{-26}$ \\
\midrule
\multirow{6}{*}{Stratified}
& \multirow{2}{*}{PC1}
  & $p$ & $1.27 \times 10^{-204}$ & $0$ & $3.41 \times 10^{-65}$ & $3.41 \times 10^{-65}$ \\
& & $q_{\text{strat}}$ & $3.18 \times 10^{-204}$ & $0$ & $5.69 \times 10^{-65}$ & $5.69 \times 10^{-65}$ \\
\cmidrule{2-7}
& \multirow{2}{*}{PC2}
  & $p$ & $1.76 \times 10^{-193}$ & $0$ & $4.56 \times 10^{-60}$ & $4.56 \times 10^{-60}$ \\
& & $q_{\text{strat}}$ & $4.39 \times 10^{-193}$ & $0$ & $5.88 \times 10^{-60}$ & $5.88 \times 10^{-60}$ \\
\cmidrule{2-7}
& \multirow{2}{*}{PC3}
  & $p$ & $5.25 \times 10^{-203}$ & $0$ & $1.57 \times 10^{-59}$ & $1.57 \times 10^{-59}$ \\
& & $q_{\text{strat}}$ & $1.31 \times 10^{-202}$ & $0$ & $1.96 \times 10^{-59}$ & $1.96 \times 10^{-59}$ \\
\bottomrule
\end{tabular}
}
\end{small}
\vskip -0.1in
\end{table}

\subsection{BBH Taxonomy and Ability Mapping}
\label{sec:bbh_taxonomy_mapping}

Table~\ref{tab:bbh_task_to_ability} summarizes the BBH subtask grouping used in our analysis. Since BBH contains heterogeneous subtasks with different reasoning demands, we consolidate the fine-grained task names into five broader ability modes (Logical, Language, Mathematical, Spatial/Temporal, and Knowledge/Domain). This mapping provides a consistent level of abstraction for reporting and interpreting stratified results at the ability level, while retaining the original BBH taxonomy at the subtask level.

\begin{table*}[htbp]
    \centering
    \small
    \renewcommand{\arraystretch}{1.4} 
    \caption{\textbf{BBH taxonomy and task grouping.}}
    \label{tab:bbh_task_to_ability}
    \begin{tabular}{l >{\raggedright\arraybackslash}p{11.5cm}} 
        \toprule
        \textbf{Ability} & \textbf{Subtasks} \\
        \midrule
        \textbf{Logical} & 
        boolean\_expressions, dyck\_languages, formal\_fallacies, web\_of\_lies, logical\_deduction\_\{three, five, seven\}\_objects \\ \addlinespace[5pt]
        
        \textbf{Language} & 
        disambiguation\_qa, hyperbaton, ruin\_names, salient\_translation\_error\_detection, snarks \\ \addlinespace[5pt]
        
        \textbf{Mathematical} & 
        geometric\_shapes, multistep\_arithmetic\_two, object\_counting, word\_sorting \\ \addlinespace[5pt]
        
        \textbf{Spatial/Temporal} & 
        date\_understanding, navigate, temporal\_sequences, tracking\_shuffled\_objects\_\{three, five, seven\}\_objects \\ \addlinespace[5pt]
        
        \textbf{Knowledge/Domain} & 
        causal\_judgement, movie\_recommendation, penguins\_in\_a\_table, reasoning\_about\_colored\_objects, sports\_understanding \\
        \bottomrule
    \end{tabular}
\end{table*}

\subsection{Computation of Continuous Factors}
\label{sec:continuous_factors}

All continuous item factors are computed from responses of the full evaluated model set
$\mathcal{U}=\mathcal{S}\cup\mathcal{T}$.
Let $a_{m,i}\in\{0,1\}$ indicate whether model $m\in\mathcal{U}$ answers item $i$ correctly.

\paragraph{Difficulty ($\delta_i$).}
Item difficulty is defined as the pass rate across the full evaluated models:
\begin{equation}
\delta_i=\frac{1}{|\mathcal{U}|}\sum_{m\in\mathcal{U}} a_{m,i}.
\end{equation}
Lower values of $\delta_i$ correspond to more difficult items.

\paragraph{Discrimination ($\gamma_i$; Kelley 27\% $D$)}
Item discrimination is defined using Kelley’s 27\% high--low criterion.
Each model is first assigned an overall benchmark score
\begin{equation}
r_m=\frac{1}{|\mathcal{I}|}\sum_{j\in\mathcal{I}} a_{m,j},
\end{equation}
where $\mathcal{I}$ denotes the item set of the benchmark.
Let $q=\max\{1,\lfloor 0.27|\mathcal{U}|\rfloor\}$, and let $\mathcal{U}_{\mathrm{high}}$ and $\mathcal{U}_{\mathrm{low}}$
denote the top-$q$ and bottom-$q$ models under $r_m$, respectively.
For each item $i$, the high/low-group pass rates and the Kelley discrimination score are defined as
\begin{equation}
\begin{aligned}
\pi_i^{\mathrm{high}} &= \frac{1}{|\mathcal{U}_{\mathrm{high}}|}\sum_{m\in\mathcal{U}_{\mathrm{high}}} a_{m,i}, \\
\pi_i^{\mathrm{low}} &= \frac{1}{|\mathcal{U}_{\mathrm{low}}|}\sum_{m\in\mathcal{U}_{\mathrm{low}}} a_{m,i}, \\
\gamma_i &= D_i = \pi_i^{\mathrm{high}} - \pi_i^{\mathrm{low}}.
\end{aligned}
\end{equation}
A higher $\gamma_i$ signifies a more pronounced separation between models with high and low performance.

\section{More Experimental Results}
\subsection{Detailed Item-level Agreement Results}
\label{sec:agreement_details}

Motivated by an observation in the main results that \textsc{GP-IRT} can achieve deceptively low MAE, Table~\ref{tab:agreement_full} reports an \textbf{item-level Agreement} metric across budgets $K\in\{10,20,30,40,50\}$, measuring the fraction of items whose predicted correctness labels exactly match the ground truth. Agreement reveals that a low MAE does not necessarily imply faithful extrapolation: MAE can be artificially reduced when a method shrinks predictions toward the global mean or benefits from error cancellation.

Across all benchmarks and budgets, \textsc{RepCore} consistently achieves the best Agreement, indicating that our predictions preserve the fine-grained per-item error pattern rather than merely matching averages. Therefore, we treat Agreement as a complementary fidelity check and caution against using MAE alone as the primary indicator of extrapolation quality.

\begin{table*}[htbp]
    \centering
    \footnotesize
    \setlength{\tabcolsep}{10pt}
    \renewcommand{\arraystretch}{1.05}
    \caption{Item-level Agreement across budgets. Agreement ($\uparrow$) over $K\in\{10,20,30,40,50\}$; \textsc{Random} is excluded as it does not extrapolate beyond the coreset. \textbf{Bold} and \underline{underline} denote the best and second-best results.}
    \label{tab:agreement_full}
    \begin{tabular}{llccccc}
        \toprule
        \textbf{Benchmark} & \textbf{Method} & \textbf{$K=10$} & \textbf{$K=20$} & \textbf{$K=30$} & \textbf{$K=40$} & \textbf{$K=50$} \\
        \midrule

        \multirow{3}{*}{\textbf{ARC-Challenge}}
        & \textsc{AnchorPoints} & 0.760 & 0.741 & 0.736 & 0.730 & 0.731 \\
        & \textsc{GP-IRT}       & \underline{0.811} & \underline{0.806} & \underline{0.806} & \underline{0.822} & \underline{0.838} \\
        & \textsc{RepCore}      & \textbf{0.848} & \textbf{0.858} & \textbf{0.862} & \textbf{0.865} & \textbf{0.867} \\
        \midrule

        \multirow{3}{*}{\textbf{BBH}}
        & \textsc{AnchorPoints} & 0.612 & 0.626 & 0.630 & 0.634 & 0.634 \\
        & \textsc{GP-IRT}       & \underline{0.618} & \underline{0.631} & \underline{0.645} & \underline{0.672} & \underline{0.693} \\
        & \textsc{RepCore}      & \textbf{0.697} & \textbf{0.714} & \textbf{0.721} & \textbf{0.725} & \textbf{0.727} \\
        \midrule

        \multirow{3}{*}{\textbf{GSM8K}}
        & \textsc{AnchorPoints} & 0.767 & 0.756 & 0.747 & 0.733 & 0.745 \\
        & \textsc{GP-IRT}       & \underline{0.820} & \underline{0.824} & \underline{0.835} & \underline{0.851} & \underline{0.863} \\
        & \textsc{RepCore}      & \textbf{0.863} & \textbf{0.876} & \textbf{0.881} & \textbf{0.884} & \textbf{0.886} \\
        \midrule

        \multirow{3}{*}{\textbf{MMLU\_Pro}}
        & \textsc{AnchorPoints} & 0.603 & 0.627 & 0.629 & 0.629 & 0.631 \\
        & \textsc{GP-IRT}       & \underline{0.611} & \underline{0.651} & \underline{0.685} & \underline{0.713} & \underline{0.730} \\
        & \textsc{RepCore}      & \textbf{0.726} & \textbf{0.744} & \textbf{0.751} & \textbf{0.754} & \textbf{0.756} \\
        \midrule

        \multirow{3}{*}{\shortstack[l]{\textbf{SEED-Bench}\\\textbf{-2-Plus}}}
        & \textsc{AnchorPoints} & \underline{0.654} & \underline{0.667} & \underline{0.670} & 0.669 & 0.671 \\
        & \textsc{GP-IRT}       & 0.594 & 0.618 & \underline{0.670} & \underline{0.718} & \underline{0.749} \\
        & \textsc{RepCore}      & \textbf{0.769} & \textbf{0.784} & \textbf{0.791} & \textbf{0.794} & \textbf{0.797} \\
        \bottomrule
    \end{tabular}
\end{table*}

\subsection{Additional Main Results on ARC-Challenge and MMLU-Pro}
\label{sec:additional_main_results}

Table~\ref{tab:main_budgetwise_new} extends the main-budget evaluation to ARC-Challenge and MMLU-Pro under the same source-scarce protocol, reporting Spearman's $\rho$ and MAE across budgets $K\in\{10,20,30,40,50\}$. The overall trend matches the main text: \textsc{RepCore} improves monotonically with $K$ and achieves the strongest ranking recovery once the budget leaves the extremely small regime, delivering the best $\rho$ on both benchmarks for $K\ge 20$.

In terms of MAE, \textsc{RepCore} remains competitive across budgets and often excels at small-to-moderate $K$, while \textsc{GP-IRT} can appear slightly better on MMLU-Pro at larger $K$. However, as discussed in Appendix~\ref{sec:agreement_details}, MAE can be reduced by averaging effects and error cancellation. Together with Table~\ref{tab:agreement_full}, these results suggest that \textsc{RepCore} achieves reliable ranking recovery while better preserving per-item correctness.

\begin{table*}[htbp]  
    \centering  
    \footnotesize
    \setlength{\tabcolsep}{3.5pt}
    \renewcommand{\arraystretch}{1.1}
    \caption{Additional main results. We report Spearman's rank correlation ($\rho$) and Mean Absolute Error (MAE). \textbf{Bold} and \underline{underline} indicate the best and second-best results among competitive methods, respectively.}
    \label{tab:main_budgetwise_new}
    \begin{tabular}{l l cc cc cc cc cc}
    \toprule
    \multirow{2}{*}{\textbf{Benchmark}} & \multirow{2}{*}{\textbf{Method}} &
    \multicolumn{2}{c}{\textbf{$K=10$}} & \multicolumn{2}{c}{\textbf{$K=20$}} & \multicolumn{2}{c}{\textbf{$K=30$}} &
    \multicolumn{2}{c}{\textbf{$K=40$}} & \multicolumn{2}{c}{\textbf{$K=50$}} \\
    \cmidrule(lr){3-4}\cmidrule(lr){5-6}\cmidrule(lr){7-8}\cmidrule(lr){9-10}\cmidrule(lr){11-12}
    & & $\rho \uparrow$ & \scriptsize MAE $\downarrow$ & $\rho \uparrow$ & \scriptsize MAE $\downarrow$ & $\rho \uparrow$ & \scriptsize MAE $\downarrow$ & $\rho \uparrow$ & \scriptsize MAE $\downarrow$ & $\rho \uparrow$ & \scriptsize MAE $\downarrow$ \\
    \midrule
    
    \multirow{4}{*}{\textbf{ARC-Challenge}} 
    & \textsc{Random}       & 0.681 & 0.089 & 0.801 & 0.061 & \underline{0.856} & \underline{0.051} & \underline{0.888} & 0.043 & \underline{0.908} & 0.038 \\
    & \textsc{AnchorPoints} & \textbf{0.724} & 0.132 & 0.767 & 0.136 & 0.792 & 0.138 & 0.797 & 0.145 & 0.796 & 0.143 \\
    & \textsc{GP-IRT}       & 0.706 & \underline{0.076} & \underline{0.808} & \underline{0.057} & 0.845 & 0.052 & 0.866 & \underline{0.043} & 0.884 & \underline{0.036} \\
    & \textsc{RepCore}      & \underline{0.707} & \textbf{0.072} & \textbf{0.820} & \textbf{0.050} & \textbf{0.871} & \textbf{0.041} & \textbf{0.900} & \textbf{0.036} & \textbf{0.921} & \textbf{0.032} \\
    \midrule
    
    \multirow{4}{*}{\textbf{MMLU\_Pro}} 
    & \textsc{Random}       & 0.702 & 0.124 & 0.812 & 0.088 & 0.865 & 0.071 & 0.895 & 0.062 & 0.914 & 0.055 \\
    & \textsc{AnchorPoints} & \textbf{0.754} & 0.140 & 0.800 & 0.107 & 0.846 & 0.102 & 0.861 & 0.095 & 0.867 & 0.095 \\
    & \textsc{GP-IRT}       & 0.646 & \underline{0.101} & \underline{0.806} & \textbf{0.065}$^{\star}$ & \underline{0.866} & \textbf{0.053}$^{\star}$ & \underline{0.899} & \textbf{0.046}$^{\star}$ & \underline{0.920} & \textbf{0.041}$^{\star}$ \\
    & \textsc{RepCore}      & \underline{0.732} & \textbf{0.091} & \textbf{0.829} & \underline{0.067} & \textbf{0.875} & \underline{0.055} & \textbf{0.906} & \underline{0.048} & \textbf{0.924} & \underline{0.043} \\
    \bottomrule
    \end{tabular}
\end{table*}

\subsection{Additional Baselines Results}
\label{app:add_baselines}

\begin{table*}[htbp]
    \centering
    \footnotesize
    \small
    \setlength{\tabcolsep}{2.2pt}
    \renewcommand{\arraystretch}{1.1}
    \caption{Mean performance: \textsc{RepCore} (ours) vs.\ EffiEval vs.\ TailoredBench Spearman $\rho$ ($\uparrow$) and MAE ($\downarrow$) averaged across target-model evaluations. \textbf{Bold} marks the best and \uline{underline} the second best per column.}
    \label{tab:add_baselines}
    \begin{tabular}{l l cc cc cc cc cc cc}
    \toprule
    \multirow{2}{*}{\textbf{Benchmark}} & \multirow{2}{*}{\textbf{Condition}} &
    \multicolumn{2}{c}{\textbf{$K=10$}} & \multicolumn{2}{c}{\textbf{$K=20$}} & \multicolumn{2}{c}{\textbf{$K=30$}} &
    \multicolumn{2}{c}{\textbf{$K=40$}} & \multicolumn{2}{c}{\textbf{$K=50$}} & \multicolumn{2}{c}{\textbf{Avg}} \\
    \cmidrule(lr){3-4}\cmidrule(lr){5-6}\cmidrule(lr){7-8}\cmidrule(lr){9-10}\cmidrule(lr){11-12}\cmidrule(lr){13-14}
    & & $\rho \uparrow$ & \scriptsize MAE $\downarrow$ & $\rho \uparrow$ & \scriptsize MAE $\downarrow$ & $\rho \uparrow$ & \scriptsize MAE $\downarrow$ & $\rho \uparrow$ & \scriptsize MAE $\downarrow$ & $\rho \uparrow$ & \scriptsize MAE $\downarrow$ & $\rho \uparrow$ & \scriptsize MAE $\downarrow$\\
    \midrule
    \multirow{3}{*}{\textbf{ARC-Challenge}}
          & \textsc{EffiEval} & \textbf{0.7520} & {0.0902} & \textbf{0.8230} & {\uline{0.0638}} & \uline{0.8640} & {\uline{0.0524}} & \uline{0.8900} & {\uline{0.0465}} & \uline{0.9180} & {\uline{0.0370}} & \textbf{0.8494} & {\uline{0.0580}} \\
          & \textsc{TailoredBench} & \uline{0.7110} & {\textbf{0.0717}} & 0.7460 & {0.0705} & 0.7540 & {0.0706} & 0.7550 & {0.0728} & 0.7530 & {0.0713} & 0.7438 & {0.0714} \\
          & \textsc{RepCore} & 0.7070 & {\uline{0.0720}} & \uline{0.8200} & {\textbf{0.0500}} & \textbf{0.8710} & {\textbf{0.0410}} & \textbf{0.9000} & {\textbf{0.0360}} & \textbf{0.9210} & {\textbf{0.0320}} & \uline{0.8438} & {\textbf{0.0462}} \\
    \midrule
    \multirow{3}{*}{\textbf{BBH}}
          & \textsc{EffiEval} & 0.4590 & {0.1390} & 0.5490 & {0.1130} & 0.6550 & {0.1030} & 0.7100 & {0.0915} & 0.7650 & {0.0824} & 0.6276 & {0.1058} \\
          & \textsc{TailoredBench} & \uline{0.6870} & {\uline{0.0990}} & \uline{0.7590} & {\uline{0.0795}} & \uline{0.7850} & {\uline{0.0748}} & \uline{0.7830} & {\uline{0.0743}} & \uline{0.7850} & {\uline{0.0737}} & \uline{0.7598} & {\uline{0.0803}} \\
          & \textsc{RepCore} & \textbf{0.7180} & {\textbf{0.0950}} & \textbf{0.8240} & {\textbf{0.0690}} & \textbf{0.8700} & {\textbf{0.0570}} & \textbf{0.8980} & {\textbf{0.0490}} & \textbf{0.9130} & {\textbf{0.0450}} & \textbf{0.8446} & {\textbf{0.0630}} \\
    \midrule
    \multirow{3}{*}{\textbf{GSM8K}}
          & \textsc{EffiEval} & 0.7090 & {0.0973} & \uline{0.7970} & {0.0831} & \uline{0.8350} & {0.0932} & \uline{0.8640} & {0.0950} & \uline{0.8790} & {0.0856} & \uline{0.8168} & {0.0908} \\
          & \textsc{TailoredBench} & \uline{0.7410} & {\textbf{0.0644}} & 0.7670 & {\uline{0.0765}} & 0.7750 & {\uline{0.0750}} & 0.7740 & {\uline{0.0789}} & 0.7770 & {\uline{0.0770}} & 0.7668 & {\uline{0.0744}} \\
          & \textsc{RepCore} & \textbf{0.7630} & {\uline{0.0710}} & \textbf{0.8720} & {\textbf{0.0470}} & \textbf{0.9050} & {\textbf{0.0390}} & \textbf{0.9260} & {\textbf{0.0350}} & \textbf{0.9380} & {\textbf{0.0320}} & \textbf{0.8808} & {\textbf{0.0448}} \\
    \midrule
    \multirow{3}{*}{\textbf{MMLU-Pro}}
          & \textsc{EffiEval} & 0.6060 & {0.2170} & 0.7380 & {0.1860} & 0.7850 & {0.1790} & 0.8050 & {0.1760} & 0.8010 & {0.1680} & 0.7470 & {0.1852} \\
          & \textsc{TailoredBench} & \uline{0.7090} & {\uline{0.0917}} & \uline{0.7810} & {\uline{0.0752}} & \uline{0.8150} & {\uline{0.0677}} & \uline{0.8170} & {\uline{0.0666}} & \uline{0.8220} & {\uline{0.0656}} & \uline{0.7888} & {\uline{0.0734}} \\
          & \textsc{RepCore} & \textbf{0.7320} & {\textbf{0.0910}} & \textbf{0.8290} & {\textbf{0.0670}} & \textbf{0.8750} & {\textbf{0.0550}} & \textbf{0.9060} & {\textbf{0.0480}} & \textbf{0.9240} & {\textbf{0.0430}} & \textbf{0.8532} & {\textbf{0.0608}} \\
    \midrule
    \multirow{3}{*}{\shortstack[l]{\textbf{SEED-Bench}\\\textbf{-2-Plus}}}
          & \textsc{EffiEval} & \textbf{0.6760} & {0.1890} & \textbf{0.7760} & {0.1170} & \uline{0.8130} & {0.1010} & \uline{0.8300} & {0.0852} & \uline{0.8570} & {0.0731} & \textbf{0.7904} & {0.1131} \\
          & \textsc{TailoredBench} & 0.6200 & {\textbf{0.0810}} & 0.6570 & {\uline{0.0788}} & 0.6970 & {\uline{0.0737}} & 0.7080 & {\uline{0.0728}} & 0.7060 & {\uline{0.0729}} & 0.6776 & {\uline{0.0758}} \\
          & \textsc{RepCore} & \uline{0.6420} & {\uline{0.0840}} & \uline{0.7550} & {\textbf{0.0600}} & \textbf{0.8190} & {\textbf{0.0510}} & \textbf{0.8530} & {\textbf{0.0450}} & \textbf{0.8740} & {\textbf{0.0400}} & \uline{0.7886} & {\textbf{0.0560}} \\
    \midrule
    \multirow{3}{*}{\textbf{Average}}
          & \textsc{EffiEval} & 0.6404 & {0.1465} & 0.7366 & {0.1126} & \uline{0.7904} & {0.1057} & \uline{0.8198} & {0.0988} & \uline{0.8440} & {0.0892} & \uline{0.7662} & {0.1106} \\
          & \textsc{TailoredBench} & \uline{0.6936} & {\textbf{0.0816}} & \uline{0.7420} & {\uline{0.0761}} & 0.7652 & {\uline{0.0724}} & 0.7674 & {\uline{0.0731}} & 0.7686 & {\uline{0.0721}} & 0.7474 & {\uline{0.0750}} \\
          & \textsc{RepCore} & \textbf{0.7124} & {\uline{0.0826}} & \textbf{0.8200} & {\textbf{0.0586}} & \textbf{0.8680} & {\textbf{0.0486}} & \textbf{0.8966} & {\textbf{0.0426}} & \textbf{0.9140} & {\textbf{0.0384}} & \textbf{0.8422} & {\textbf{0.0542}} \\
    \bottomrule
    \end{tabular}
\end{table*}

Table~\ref{tab:add_baselines} compares EffiEval and TailoredBench with \textsc{RepCore} across all five benchmarks and five coreset budgets. These two methods are included as auxiliary baselines to provide a broader comparison with recent benchmark compression methods, but are not further used in diagnostic ablations because their assumptions can be fragile under the source-scarce setting studied here.

TailoredBench includes an explicit source-model filtering stage before subset construction. In the source-scarce setting, this stage further narrows the already limited source pool, leaving the method with less usable source information than other baselines. EffiEval, on the other hand, relies on the transferability of activation-based similarity across models, whose reliability can fluctuate considerably in heterogeneous source pools with different architectures and model families. These source-scarce failure modes are also reflected in their limited competitiveness in Table~\ref{tab:add_baselines}. Therefore, to control for factors unrelated to the diagnostic variables under study, we report their full end-to-end results here but do not include them in the subsequent ablation analyses.

\subsection{Stability Comparison: \textsc{RepCore} vs. Baselines}
\label{app:stability_analysis}

\begin{figure*}[!htbp]
    \centering
    
    \includegraphics[width=0.9\textwidth]{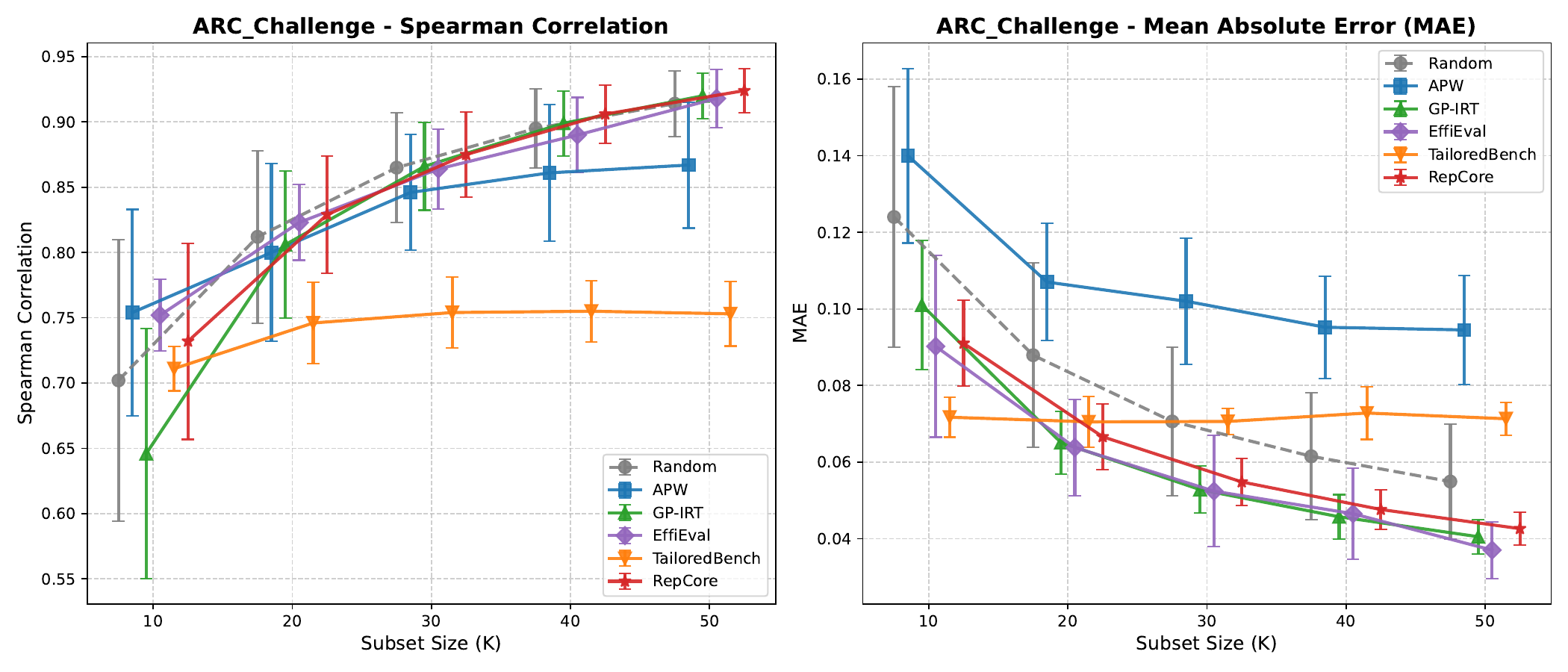}
    \vskip 0.2in
    
    \includegraphics[width=0.9\textwidth]{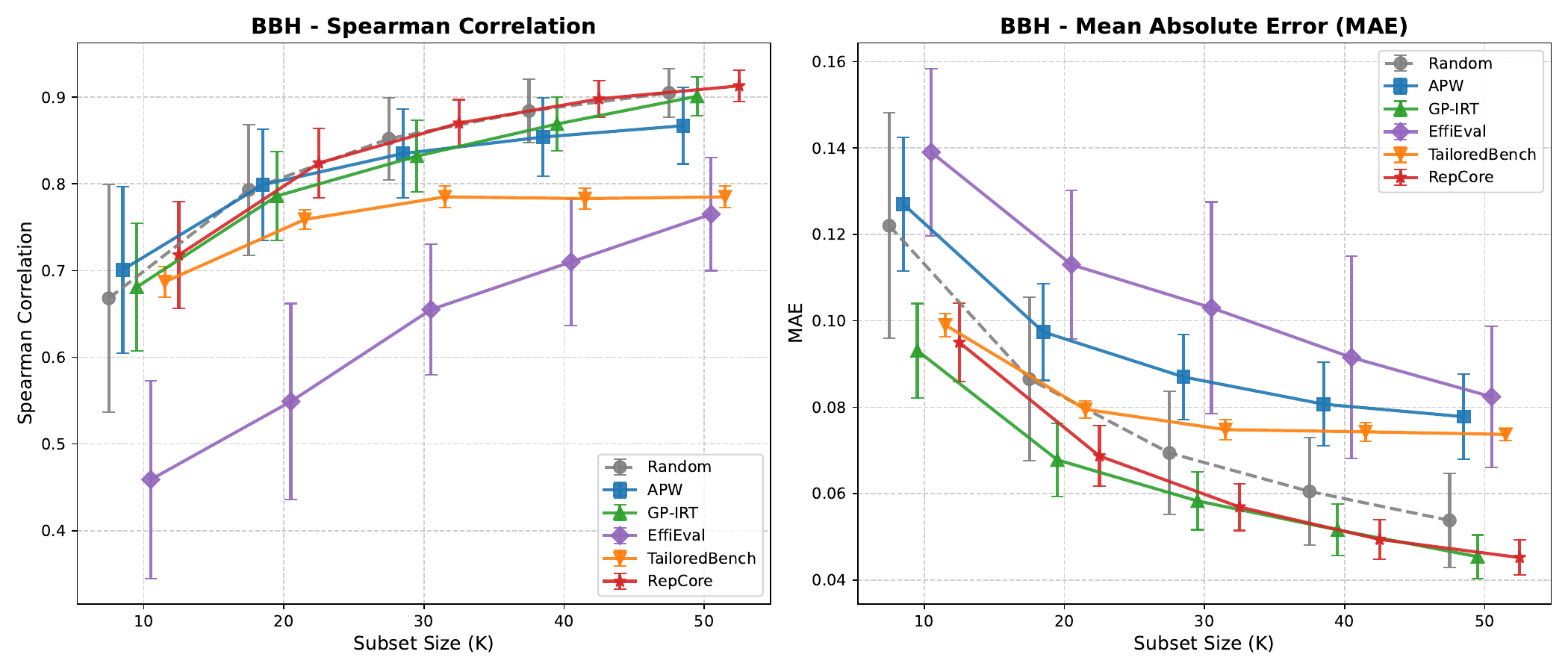}
    \vskip 0.2in
    
    \includegraphics[width=0.9\textwidth]{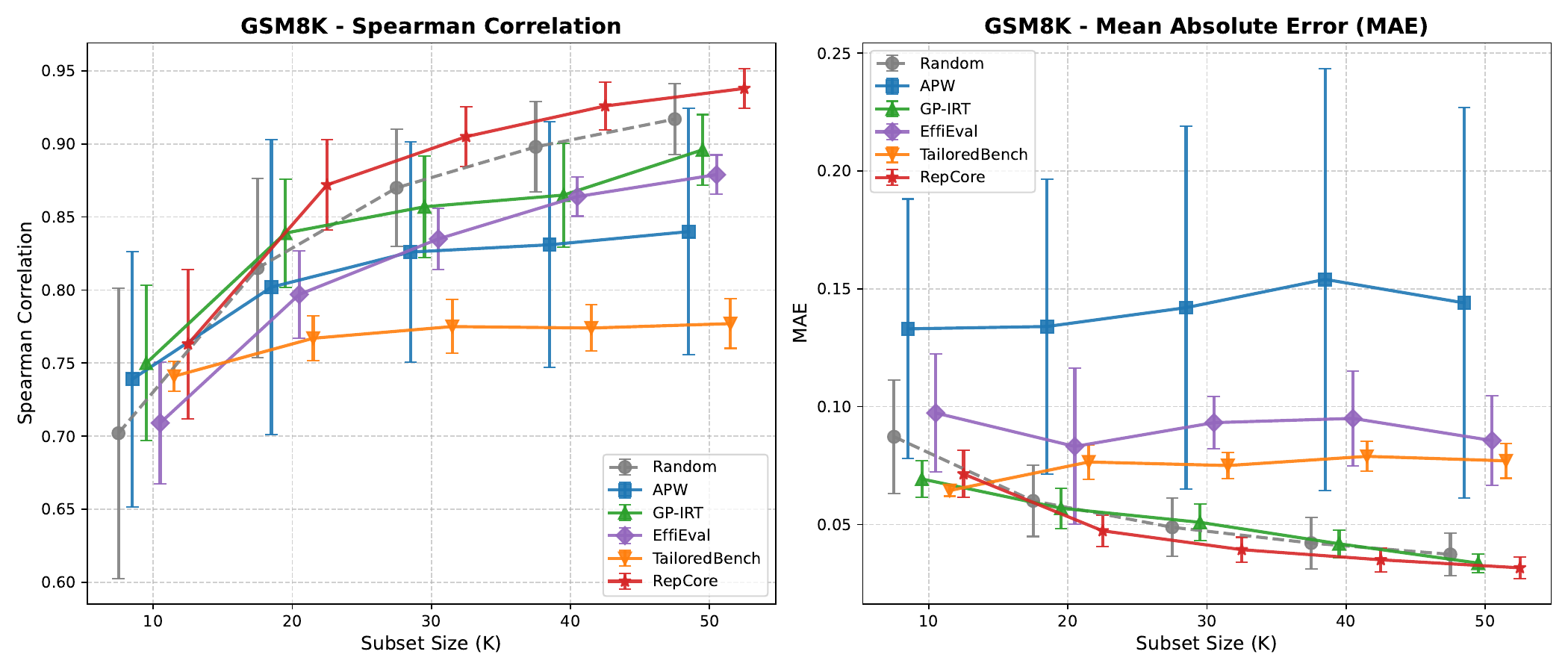}
    
    \caption{Stability Analysis (Part I): ARC, BBH, and GSM8K. The charts display Spearman Correlation (left) and MAE (right) with standard deviation error bars. Note the stability of \textsc{RepCore} across varying coreset sizes.}
    \label{fig:stability_part1}
\end{figure*}

\begin{figure*}[!htbp]
    \centering
    
    \includegraphics[width=0.9\textwidth]{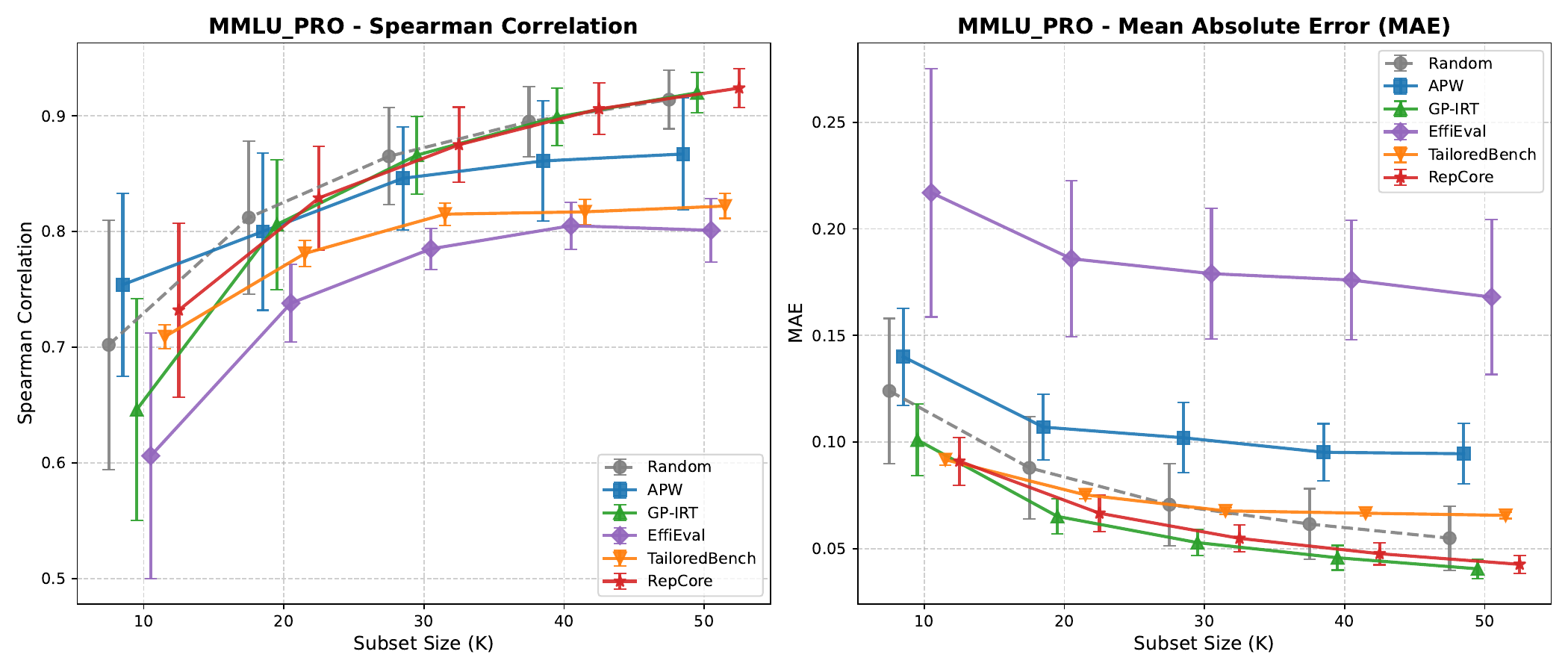}
    \vskip 0.2in
    
    \includegraphics[width=0.9\textwidth]{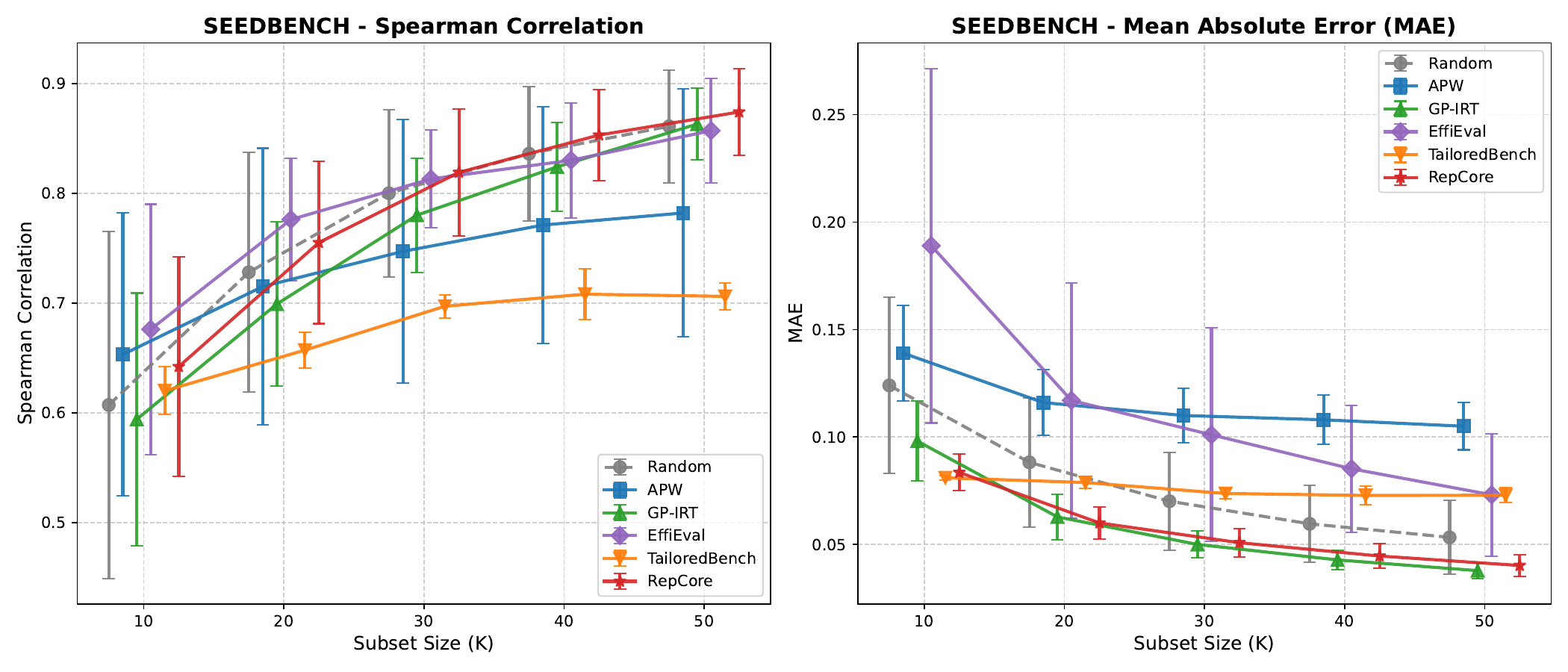}
    
    \caption{Stability Analysis (Part II): MMLU-PRO and SEEDBENCH. Comparisons on larger-scale benchmarks similarly demonstrate the robustness of the proposed method.}
    \label{fig:stability_part2}
\end{figure*}

In this section, we further examine the stability of different compression methods by reporting the Combo Std across 10 random source-model combinations, with each combination averaged over 50 random anchor selections. Figures~\ref{fig:stability_part1} and~\ref{fig:stability_part2} visualize the mean performance and standard deviation on all five benchmarks, split into two parts for readability. In each benchmark, the left panel reports Spearman's rank correlation and the right panel reports MAE across coreset sizes. The error bars indicate variation across runs. Overall, \textsc{RepCore} (red star) achieves consistently strong performance with relatively small variance compared to baselines, providing further evidence for the effectiveness and robustness of our method.

\subsection{Computational Cost and Storage Overhead}
\label{app:cost_storage}

In this section, we report wall-clock timing and storage measurements under the same model pool and experimental configuration as the main experiments to quantify the additional cost introduced by \textsc{RepCore}. 
Following the structure of the pipeline, we decompose the cost into three parts: running each source model on the full benchmark, applying each method's compression procedure after source inference, and storing the hidden states used by \textsc{RepCore}.
The first part is common to all methods that rely on source models, since they must evaluate the source models on the full benchmark to construct the response matrix for coreset selection and score extrapolation. 
\textsc{RepCore} reuses the same source inference runs. Its only modification during this stage is to register a forward hook that records hidden states during the forward pass. All timing measurements in this section are obtained on a server equipped with NVIDIA H20 96GB GPUs.

\paragraph{Source-model inference cost.}
We measure the average wall-clock time for a single source model to complete inference on each benchmark, with and without registering the forward hook used to capture hidden states. 
As shown in Table~\ref{tab:cost_source_inference}, the hook introduces no meaningful additional overhead: the average runtime increases from 9852.8s to 9892.5s, resulting approximately 0.4\% relative overhead.

\begin{table}[!htbp]
\centering
\caption{Source-model inference time with and without hidden-state extraction. 
The numbers report the average wall-clock time for a single source model to complete inference on one benchmark.}
\label{tab:cost_source_inference}
\small
\setlength{\tabcolsep}{12pt}
\begin{tabular}{ccc}
\toprule
 & Hook (s) & No hook (s) \\
\midrule
Average & 9892.5 & 9852.8 \\
\bottomrule
\end{tabular}
\end{table}



\paragraph{\textsc{RepCore} compression cost after source inference.}
After source-model outputs and hidden states have been collected, \textsc{RepCore} runs its compression procedure, including representation alignment, consensus embedding construction, clustering, and fitting the lightweight extrapolator. 
This stage takes 479.7s on average over the five benchmarks.

\paragraph{Hidden-state storage.}
The additional storage required by \textsc{RepCore} comes from saving source-model hidden states. 
As shown in Table~\ref{tab:cost_storage}, the storage remains modest across benchmarks, ranging from 0.11GB on ARC-Challenge to 1.25GB on MMLU-Pro.

\begin{table}[!htbp]
\centering
\caption{Storage overhead for saved hidden states.}
\label{tab:cost_storage}
\small
\setlength{\tabcolsep}{7pt}
\begin{tabular}{cccccc}
\toprule
Dataset & ARC-C & BBH & GSM8K & MMLU-Pro & SEED-Bench \\
\midrule
Size (GB) & 0.11 & 0.67 & 0.13 & 1.25 & 0.17 \\
\bottomrule
\end{tabular}
\end{table}


\subsection{Discrete-Representation Ablation}
\label{sec:discrete_ablation}

Table~\ref{tab:ablation_discrete_full} presents a controlled ablation that isolates the contribution of continuous aligned representations in the coreset selection stage. We replace our consensus embeddings with concatenated binary correctness vectors and perform clustering in this discrete space (\textsc{0/1 K-means+Ridge}), while keeping the downstream extrapolation stage identical. This design ensures that any performance difference is attributable to the representation used for summarizing item structure rather than the predictor itself.

Across all datasets, \textsc{RepCore} consistently yields higher (or comparable) Spearman correlation and lower MAE for most budgets. While the discrete baseline can be competitive on MAE at the smallest budget for a few cases, \textsc{RepCore} achieves more stable improvements as $K$ increases, suggesting that the continuous embedding geometry captures fine-grained inter-item relations that binary outcome vectors fail to preserve, thereby enabling more reliable and transferable coreset selection.

\begin{table*}[!htbp]
    \centering
    \footnotesize
    \setlength{\tabcolsep}{3.5pt}
    \renewcommand{\arraystretch}{1.08}
   \caption{Discrete-representation ablation. Per-dataset results comparing \textsc{RepCore} with a discrete clustering variant (\textsc{0/1 K-means+Ridge}). We report Spearman's $\rho$ ($\uparrow$) and MAE ($\downarrow$) across budgets $K\in\{10,20,30,40,50\}$. \textbf{Bold} indicates the best result.}  
    \label{tab:ablation_discrete_full}
    \begin{tabular}{l l cc cc cc cc cc}
        \toprule
        \multirow{2}{*}{\textbf{Benchmark}} & \multirow{2}{*}{\textbf{Method}} & 
        \multicolumn{2}{c}{\textbf{$K=10$}} & \multicolumn{2}{c}{\textbf{$K=20$}} & \multicolumn{2}{c}{\textbf{$K=30$}} & 
        \multicolumn{2}{c}{\textbf{$K=40$}} & \multicolumn{2}{c}{\textbf{$K=50$}} \\
        \cmidrule(lr){3-4}\cmidrule(lr){5-6}\cmidrule(lr){7-8}\cmidrule(lr){9-10}\cmidrule(lr){11-12}
        & & $\rho \uparrow$ & \scriptsize MAE $\downarrow$ & $\rho \uparrow$ & \scriptsize MAE $\downarrow$ & $\rho \uparrow$ & \scriptsize MAE $\downarrow$ & $\rho \uparrow$ & \scriptsize MAE $\downarrow$ & $\rho \uparrow$ & \scriptsize MAE $\downarrow$ \\
        \midrule

        \multirow{2}{*}{\textbf{ARC-Challenge}} 
        & \textsc{0/1 K-means+Ridge}    & 0.685 & \textbf{0.0719} & 0.803 & 0.0578 & 0.850 & 0.0512 & 0.874 & 0.0485 & 0.894 & 0.0449 \\
        & \textsc{RepCore}              & \textbf{0.707} & 0.0720 & \textbf{0.820} & \textbf{0.0500} & \textbf{0.871} & \textbf{0.0409} & \textbf{0.900} & \textbf{0.0358} & \textbf{0.921} & \textbf{0.0321} \\
        \midrule

        \multirow{2}{*}{\textbf{BBH}} 
        & \textsc{0/1 K-means+Ridge}    & 0.717 & 0.0958 & 0.821 & 0.0713 & 0.862 & 0.0604 & 0.884 & 0.0533 & 0.896 & 0.0487 \\
        & \textsc{RepCore}              & \textbf{0.718} & \textbf{0.0950} & \textbf{0.824} & \textbf{0.0687} & \textbf{0.870} & \textbf{0.0569} & \textbf{0.898} & \textbf{0.0494} & \textbf{0.913} & \textbf{0.0452} \\
        \midrule

        \multirow{2}{*}{\textbf{GSM8K}} 
        & \textsc{0/1 K-means+Ridge}    & 0.754 & 0.0745 & 0.845 & 0.0579 & 0.886 & 0.0514 & 0.909 & 0.0471 & 0.922 & 0.0442 \\
        & \textsc{RepCore}              & \textbf{0.763} & \textbf{0.0714} & \textbf{0.872} & \textbf{0.0473} & \textbf{0.905} & \textbf{0.0393} & \textbf{0.926} & \textbf{0.0350} & \textbf{0.938} & \textbf{0.0316} \\
        \midrule

        \multirow{2}{*}{\textbf{SEED-Bench-2-Plus}} 
        & \textsc{0/1 K-means+Ridge}    & 0.514 & 0.0860 & 0.697 & 0.0656 & 0.786 & 0.0566 & 0.829 & 0.0511 & 0.857 & 0.0473 \\
        & \textsc{RepCore}              & \textbf{0.642} & \textbf{0.0835} & \textbf{0.755} & \textbf{0.0600} & \textbf{0.819} & \textbf{0.0508} & \textbf{0.853} & \textbf{0.0446} & \textbf{0.874} & \textbf{0.0402} \\
        \midrule

        \multirow{2}{*}{\textbf{MMLU-Pro}} 
        & \textsc{0/1 K-means+Ridge}    & 0.725 & \textbf{0.0874} & 0.824 & 0.0685 & 0.865 & 0.0575 & 0.890 & 0.0505 & 0.904 & 0.0462 \\
        & \textsc{RepCore}              & \textbf{0.732} & 0.0910 & \textbf{0.829} & \textbf{0.0666} & \textbf{0.875} & \textbf{0.0548} & \textbf{0.906} & \textbf{0.0476} & \textbf{0.924} & \textbf{0.0426} \\
        \bottomrule
    \end{tabular}
\end{table*}

\subsection{Sensitivity to MLP Architecture}
\label{app:mlp_ablation}

To examine whether \textsc{RepCore} is sensitive to the specific width and depth of the shared MLP backbone, we conduct an architecture ablation on BBH. 
We compare the default backbone with wider, deeper, and shallower variants, with all other components kept unchanged. 
Table~\ref{tab:mlp_ablation_bbh} reports the results across different coreset budgets.

\begin{table*}[htbp]
    \centering
    \footnotesize
    \setlength{\tabcolsep}{3.5pt}
    \renewcommand{\arraystretch}{1.1}
    \caption{Results on BBH under alternative shared-backbone configurations. Spearman's $\rho$ ($\uparrow$) and MAE ($\downarrow$) are reported across coreset budgets $K \in {10,20,30,40,50}$. The first row is the original \textsc{RepCore} setting; the remaining rows are different MLP hidden-layer configurations (values denote hidden sizes). \textbf{Bold} marks the best and \uline{underline} the second best per column.}
    \label{tab:mlp_ablation_bbh}
    \begin{tabular}{l cc cc cc cc cc cc}
    \toprule
    \multirow{2}{*}{\textbf{MLP hidden dims}} &
    \multicolumn{2}{c}{\textbf{$K=10$}} & \multicolumn{2}{c}{\textbf{$K=20$}} & \multicolumn{2}{c}{\textbf{$K=30$}} &
    \multicolumn{2}{c}{\textbf{$K=40$}} & \multicolumn{2}{c}{\textbf{$K=50$}} & \multicolumn{2}{c}{\textbf{Avg}} \\
    \cmidrule(lr){2-3}\cmidrule(lr){4-5}\cmidrule(lr){6-7}\cmidrule(lr){8-9}\cmidrule(lr){10-11}\cmidrule(lr){12-13}
    & $\rho$ & {MAE} & $\rho$ & {MAE} & $\rho$ & {MAE} & $\rho$ & {MAE} & $\rho$ & {MAE} & $\rho$ & {MAE} \\
    \midrule
    \textsc{RepCore (ours)} & 0.718 & {\textbf{0.095}} & \textbf{0.824} & {\textbf{0.069}} & \textbf{0.870} & {\uline{0.057}} & \uline{0.898} & {\uline{0.049}} & 0.913 & {0.045} & \uline{0.845} & {\uline{0.063}} \\
    \textsc{1024-512-128-32} & \textbf{0.723} & {\uline{0.095}} & \uline{0.822} & {0.070} & \uline{0.868} & {0.057} & 0.894 & {0.050} & 0.911 & {0.045} & 0.844 & {0.064} \\
    \textsc{1024-512-32} & \uline{0.719} & {0.095} & 0.821 & {0.070} & 0.868 & {\textbf{0.056}} & \textbf{0.900} & {\textbf{0.048}} & \textbf{0.918} & {\textbf{0.043}} & \textbf{0.845} & {\textbf{0.062}} \\
    \textsc{512-32} & 0.710 & {0.096} & 0.816 & {\uline{0.069}} & 0.866 & {0.057} & 0.897 & {0.050} & \uline{0.917} & {\uline{0.045}} & 0.841 & {0.063} \\
    \bottomrule
    \end{tabular}
\end{table*}

Overall, the results are highly similar across different MLP architectures, with no significant performance gap among the default, wider, deeper, and shallower variants. 
This suggests that \textsc{RepCore} is not sensitive to moderate changes in the width or depth of the shared MLP backbone.

\subsection{Extrapolation Feature Ablation}
\label{app:extrapolation_feature}
To examine the design choice of using the averaged source correctness $x_i$ as the extrapolation feature, we compare it with an alternative that uses the concatenated source correctness vector as input to the same ridge regressor. 
The scalar variant, denoted as $\mathrm{label}_{\mathrm{avg}}$, uses $x_i=\frac{1}{|\mathcal{S}|}\sum_{m\in\mathcal{S}}y_{m,i}$, whereas $\mathrm{label}_{\mathrm{vec}}$ uses $(y_{m_1,i},\ldots,y_{m_{|\mathcal{S}|},i})\in\mathbb{R}^{|\mathcal{S}|}$, where $\mathcal{S}=\{m_1,\ldots,m_{|\mathcal{S}|}\}$. 
All other components are kept unchanged. 
As shown in Table~\ref{tab:extrapolation_feature_ablation}, the scalar feature consistently improves ranking recovery while maintaining comparable MAE, supporting its use as a stable low-variance item-level anchor in the source-scarce regime.

\begin{table}[htbp]
\centering
\caption{Extrapolation feature ablation averaged over five benchmarks.
$\mathrm{label}_{\mathrm{avg}}$ uses the scalar averaged source correctness $x_i$ as the input feature, whereas $\mathrm{label}_{\mathrm{vec}}$ uses the full source correctness vector. Each cell reports Spearman's $\rho$ / MAE.}
\label{tab:extrapolation_feature_ablation}
\small
\setlength{\tabcolsep}{5pt}
\begin{tabular}{lccccc}
\toprule
Feature & K=10 & K=20 & K=30 & K=40 & K=50 \\
\midrule
$\mathrm{label}_{\mathrm{vec}}$ & 0.567 / 0.081 & 0.694 / 0.060 & 0.788 / 0.050 & 0.835 / 0.044 & 0.861 / 0.040 \\
$\mathrm{label}_{\mathrm{avg}}$ & 0.642 / 0.084 & 0.755 / 0.060 & 0.819 / 0.051 & 0.853 / 0.045 & 0.874 / 0.040 \\
\bottomrule
\end{tabular}
\end{table}

\subsection{Detailed Results on Source Model Quantity}
\label{sec:source_num_ablation}

In this section, we provide a comprehensive breakdown of the  analysis presented in Section~\ref{sec:Experiment}. To rigorously evaluate the sensitivity of each method to the availability of source models, we vary the source pool size $|\mathcal{S}| \in \{5, 10, 15, 20\}$ across the full spectrum of computational budgets $K \in \{10, 20, 30, 40, 50\}$. The detailed performance metrics for each budget setting are reported in Tables~\ref{tab:ablation_quantity_K10}-\ref{tab:ablation_quantity_K50}.

The results consistently demonstrate the robustness of \textsc{RepCore} in source-scarce regimes. As evidenced across all budget levels, output-based baselines such as \textsc{AnchorPoints} exhibit sharp performance degradation when $|\mathcal{S}|=5$, suggesting a heavy reliance on extensive statistical sampling to estimate item difficulty. In contrast, \textsc{RepCore} maintains high ranking correlations even with minimal supervision, validating our hypothesis that the geometric structure of the representation space can be effectively aligned with limited source signals. Furthermore, performance saturation is observed as $|\mathcal{S}|$ approaches 20, indicating that our framework captures the essential topology of the item space early.


\begin{table*}[htbp]
\centering
\footnotesize
\setlength{\tabcolsep}{4pt}
\caption{Detailed results with budget $K=10$ under varying source pool sizes $|\mathcal{S}|$. We report Spearman's rank correlation ($\rho$, $\uparrow$) and Mean Absolute Error (MAE, $\downarrow$). \textbf{Bold} and \underline{underlined} values denote the best and second-best results, respectively.}
\label{tab:ablation_quantity_K10}
\begin{tabular}{l cc cc cc cc cc }
\toprule
\multirow{2}{*}{\textbf{Method}} & \multicolumn{2}{c}{\textbf{ARC-C}} & \multicolumn{2}{c}{\textbf{BBH}} & \multicolumn{2}{c}{\textbf{GSM8K}} & \multicolumn{2}{c}{\textbf{SeedBench}} & \multicolumn{2}{c}{\textbf{MMLU-Pro}} \\
\cmidrule(lr){2-3} \cmidrule(lr){4-5} \cmidrule(lr){6-7} \cmidrule(lr){8-9} \cmidrule(lr){10-11}
 & $\rho \uparrow$ & \scriptsize MAE $\downarrow$ & $\rho \uparrow$ & \scriptsize MAE $\downarrow$ & $\rho \uparrow$ & \scriptsize MAE $\downarrow$ & $\rho \uparrow$ & \scriptsize MAE $\downarrow$ & $\rho \uparrow$ & \scriptsize MAE $\downarrow$ \\
\midrule
\multicolumn{11}{c}{\textbf{$|\mathcal{S}| = 5$}} \\
\midrule
\textsc{AnchorPoints} & \underline{0.632} & 0.216 & 0.614 & 0.176 & 0.714 & 0.210 & 0.513 & 0.228 & 0.673 & 0.187 \\
\textsc{GP-IRT} & 0.609 & \underline{0.088} & \underline{0.639} & \underline{0.106} & \underline{0.729} & \underline{0.079} & \underline{0.581} & \underline{0.095} & \underline{0.703} & \underline{0.115} \\
\textsc{RepCore} & \textbf{0.704} & \textbf{0.077} & \textbf{0.706} & \textbf{0.104} & \textbf{0.749} & \textbf{0.074} & \textbf{0.590} & \textbf{0.088} & \textbf{0.744} & \textbf{0.096} \\
\midrule
\multicolumn{11}{c}{\textbf{$|\mathcal{S}| = 10$}} \\
\midrule
\textsc{AnchorPoints} & \textbf{0.724} & 0.132 & \underline{0.701} & 0.127 & 0.739 & 0.133 & \textbf{0.653} & 0.139 & \textbf{0.754} & 0.140 \\
\textsc{GP-IRT} & 0.706 & \underline{0.076} & 0.681 & \textbf{0.093} & \underline{0.750} & \textbf{0.069} & 0.594 & \underline{0.098} & 0.646 & \underline{0.101} \\
\textsc{RepCore} & \underline{0.707} & \textbf{0.072} & \textbf{0.718} & \underline{0.095} & \textbf{0.763} & \underline{0.071} & \underline{0.642} & \textbf{0.084} & \underline{0.732} & \textbf{0.091} \\
\midrule
\multicolumn{11}{c}{\textbf{$|\mathcal{S}| = 15$}} \\
\midrule
\textsc{AnchorPoints} & 0.562 & 0.126 & \textbf{0.765} & 0.117 & \textbf{0.784} & 0.110 & \textbf{0.715} & 0.123 & \textbf{0.798} & 0.128 \\
\textsc{GP-IRT} & \textbf{0.719} & \underline{0.081} & \underline{0.674} & \textbf{0.090} & 0.734 & \underline{0.073} & 0.607 & \underline{0.092} & 0.612 & \underline{0.098} \\
\textsc{RepCore} & \underline{0.699} & \textbf{0.074} & 0.672 & \underline{0.098} & \underline{0.763} & \textbf{0.070} & \underline{0.619} & \textbf{0.091} & \underline{0.687} & \textbf{0.095} \\
\midrule
\multicolumn{11}{c}{\textbf{$|\mathcal{S}| = 20$}} \\
\midrule
\textsc{AnchorPoints} & \textbf{0.683} & 0.106 & \textbf{0.787} & 0.115 & \underline{0.762} & 0.098 & \textbf{0.737} & 0.124 & \textbf{0.809} & 0.121 \\
\textsc{GP-IRT} & 0.638 & \underline{0.082} & 0.620 & \textbf{0.096} & 0.711 & \underline{0.070} & 0.616 & \underline{0.091} & 0.566 & \underline{0.106} \\
\textsc{RepCore} & \underline{0.671} & \textbf{0.074} & \underline{0.717} & \underline{0.099} & \textbf{0.787} & \textbf{0.065} & \underline{0.638} & \textbf{0.086} & \underline{0.682} & \textbf{0.093} \\
\bottomrule
\end{tabular}
\end{table*}

\begin{table*}[htbp]
\centering
\footnotesize
\setlength{\tabcolsep}{4pt}
\caption{Detailed  results with budget $K=20$ under varying source pool sizes $|\mathcal{S}|$. We report Spearman's rank correlation ($\rho$, $\uparrow$) and Mean Absolute Error (MAE, $\downarrow$). \textbf{Bold} and \underline{underlined} values denote the best and second-best results, respectively.}
\label{tab:ablation_quantity_K20}
\begin{tabular}{l cc cc cc cc cc }
\toprule
\multirow{2}{*}{\textbf{Method}} & \multicolumn{2}{c}{\textbf{ARC-C}} & \multicolumn{2}{c}{\textbf{BBH}} & \multicolumn{2}{c}{\textbf{GSM8K}} & \multicolumn{2}{c}{\textbf{SeedBench}} & \multicolumn{2}{c}{\textbf{MMLU-Pro}} \\
\cmidrule(lr){2-3} \cmidrule(lr){4-5} \cmidrule(lr){6-7} \cmidrule(lr){8-9} \cmidrule(lr){10-11}
 & $\rho \uparrow$ & \scriptsize MAE $\downarrow$ & $\rho \uparrow$ & \scriptsize MAE $\downarrow$ & $\rho \uparrow$ & \scriptsize MAE $\downarrow$ & $\rho \uparrow$ & \scriptsize MAE $\downarrow$ & $\rho \uparrow$ & \scriptsize MAE $\downarrow$ \\
\midrule
\multicolumn{11}{c}{\textbf{$|\mathcal{S}| = 5$}} \\
\midrule
\textsc{AnchorPoints} & 0.647 & 0.260 & 0.619 & 0.156 & 0.722 & 0.230 & 0.539 & 0.220 & 0.678 & 0.172 \\
\textsc{GP-IRT} & \underline{0.726} & \underline{0.077} & \underline{0.730} & \underline{0.078} & \underline{0.820} & \underline{0.053} & \underline{0.683} & \underline{0.071} & \underline{0.808} & \textbf{0.067} \\
\textsc{RepCore} & \textbf{0.818} & \textbf{0.053} & \textbf{0.808} & \textbf{0.075} & \textbf{0.845} & \textbf{0.051} & \textbf{0.731} & \textbf{0.065} & \textbf{0.830} & \underline{0.069} \\
\midrule
\multicolumn{11}{c}{\textbf{$|\mathcal{S}| = 10$}} \\
\midrule
\textsc{AnchorPoints} & 0.767 & 0.136 & \underline{0.799} & 0.097 & 0.794 & 0.132 & \underline{0.741} & 0.119 & \underline{0.800} & 0.107 \\
\textsc{GP-IRT} & \underline{0.808} & \underline{0.057} & 0.786 & \textbf{0.068} & \underline{0.839} & \underline{0.057} & 0.699 & \underline{0.063} & 0.806 & \textbf{0.065} \\
\textsc{RepCore} & \textbf{0.820} & \textbf{0.050} & \textbf{0.824} & \underline{0.069} & \textbf{0.872} & \textbf{0.047} & \textbf{0.755} & \textbf{0.060} & \textbf{0.829} & \underline{0.067} \\
\midrule
\multicolumn{11}{c}{\textbf{$|\mathcal{S}| = 15$}} \\
\midrule
\textsc{AnchorPoints} & 0.813 & 0.107 & \textbf{0.840} & 0.086 & \underline{0.846} & 0.102 & \textbf{0.796} & 0.096 & \textbf{0.848} & 0.094 \\
\textsc{GP-IRT} & \textbf{0.826} & \underline{0.054} & 0.779 & \textbf{0.068} & 0.828 & \underline{0.051} & \underline{0.719} & \textbf{0.061} & 0.755 & \underline{0.069} \\
\textsc{RepCore} & \underline{0.822} & \textbf{0.051} & \underline{0.807} & \underline{0.072} & \textbf{0.856} & \textbf{0.050} & 0.713 & \underline{0.063} & \underline{0.838} & \textbf{0.067} \\
\midrule
\multicolumn{11}{c}{\textbf{$|\mathcal{S}| = 20$}} \\
\midrule
\textsc{AnchorPoints} & \underline{0.817} & 0.092 & \textbf{0.861} & 0.084 & 0.831 & 0.088 & \textbf{0.820} & 0.092 & \textbf{0.861} & 0.088 \\
\textsc{GP-IRT} & 0.774 & \underline{0.055} & 0.775 & \textbf{0.067} & \underline{0.842} & \textbf{0.049} & 0.716 & \textbf{0.059} & 0.687 & \underline{0.078} \\
\textsc{RepCore} & \textbf{0.823} & \textbf{0.051} & \underline{0.844} & \underline{0.071} & \textbf{0.870} & \underline{0.051} & \underline{0.743} & \underline{0.061} & \underline{0.827} & \textbf{0.065} \\
\bottomrule
\end{tabular}
\end{table*}

\begin{table*}[htbp]
\centering
\footnotesize
\setlength{\tabcolsep}{4pt}
\caption{Detailed  results with budget $K=30$ under varying source pool sizes $|\mathcal{S}|$. We report Spearman's rank correlation ($\rho$, $\uparrow$) and Mean Absolute Error (MAE, $\downarrow$). \textbf{Bold} and \underline{underlined} values denote the best and second-best results, respectively.}
\label{tab:ablation_quantity_K30}
\begin{tabular}{l cc cc cc cc cc }
\toprule
\multirow{2}{*}{\textbf{Method}} & \multicolumn{2}{c}{\textbf{ARC-C}} & \multicolumn{2}{c}{\textbf{BBH}} & \multicolumn{2}{c}{\textbf{GSM8K}} & \multicolumn{2}{c}{\textbf{SeedBench}} & \multicolumn{2}{c}{\textbf{MMLU-Pro}} \\
\cmidrule(lr){2-3} \cmidrule(lr){4-5} \cmidrule(lr){6-7} \cmidrule(lr){8-9} \cmidrule(lr){10-11}
 & $\rho \uparrow$ & \scriptsize MAE $\downarrow$ & $\rho \uparrow$ & \scriptsize MAE $\downarrow$ & $\rho \uparrow$ & \scriptsize MAE $\downarrow$ & $\rho \uparrow$ & \scriptsize MAE $\downarrow$ & $\rho \uparrow$ & \scriptsize MAE $\downarrow$ \\
\midrule
\multicolumn{11}{c}{\textbf{$|\mathcal{S}| = 5$}} \\
\midrule
\textsc{AnchorPoints} & 0.656 & 0.259 & 0.627 & 0.155 & 0.718 & 0.260 & 0.535 & 0.212 & 0.664 & 0.163 \\
\textsc{GP-IRT} & \underline{0.813} & \underline{0.068} & \underline{0.806} & \underline{0.062} & \underline{0.864} & \textbf{0.042} & \underline{0.774} & \underline{0.055} & \underline{0.853} & \textbf{0.057} \\
\textsc{RepCore} & \textbf{0.868} & \textbf{0.044} & \textbf{0.860} & \textbf{0.060} & \textbf{0.900} & \underline{0.044} & \textbf{0.790} & \textbf{0.053} & \textbf{0.882} & \textbf{0.057} \\
\midrule
\multicolumn{11}{c}{\textbf{$|\mathcal{S}| = 10$}} \\
\midrule
\textsc{AnchorPoints} & 0.792 & 0.138 & \underline{0.835} & 0.087 & 0.832 & 0.125 & \underline{0.794} & 0.108 & 0.846 & 0.102 \\
\textsc{GP-IRT} & \underline{0.845} & \underline{0.052} & 0.832 & \underline{0.058} & \underline{0.857} & \underline{0.051} & 0.780 & \textbf{0.050} & \underline{0.866} & \textbf{0.053} \\
\textsc{RepCore} & \textbf{0.871} & \textbf{0.041} & \textbf{0.870} & \textbf{0.057} & \textbf{0.905} & \textbf{0.039} & \textbf{0.819} & \underline{0.051} & \textbf{0.875} & \underline{0.055} \\
\midrule
\multicolumn{11}{c}{\textbf{$|\mathcal{S}| = 15$}} \\
\midrule
\textsc{AnchorPoints} & 0.822 & 0.104 & \textbf{0.873} & 0.076 & \underline{0.887} & 0.098 & \textbf{0.838} & 0.085 & \underline{0.873} & 0.084 \\
\textsc{GP-IRT} & \textbf{0.877} & \textbf{0.040} & 0.843 & \textbf{0.056} & 0.872 & \underline{0.043} & \underline{0.787} & \textbf{0.048} & 0.837 & \underline{0.056} \\
\textsc{RepCore} & \underline{0.875} & \underline{0.041} & \underline{0.857} & \underline{0.058} & \textbf{0.898} & \textbf{0.041} & 0.783 & \underline{0.052} & \textbf{0.886} & \textbf{0.054} \\
\midrule
\multicolumn{11}{c}{\textbf{$|\mathcal{S}| = 20$}} \\
\midrule
\textsc{AnchorPoints} & \underline{0.835} & 0.094 & \textbf{0.886} & 0.074 & 0.874 & 0.084 & \textbf{0.864} & 0.076 & \textbf{0.889} & 0.079 \\
\textsc{GP-IRT} & 0.833 & \underline{0.043} & 0.858 & \textbf{0.054} & \underline{0.882} & \underline{0.042} & 0.768 & \textbf{0.049} & 0.798 & \underline{0.063} \\
\textsc{RepCore} & \textbf{0.878} & \textbf{0.041} & \underline{0.880} & \underline{0.055} & \textbf{0.909} & \textbf{0.041} & \underline{0.799} & \underline{0.050} & \underline{0.880} & \textbf{0.053} \\
\bottomrule
\end{tabular}
\end{table*}

\begin{table*}[htbp]
\centering
\footnotesize
\setlength{\tabcolsep}{4pt}
\caption{Detailed  results with budget $K=40$ under varying source pool sizes $|\mathcal{S}|$. We report Spearman's rank correlation ($\rho$, $\uparrow$) and Mean Absolute Error (MAE, $\downarrow$). \textbf{Bold} and \underline{underlined} values denote the best and second-best results, respectively.}
\label{tab:ablation_quantity_K40}
\begin{tabular}{l cc cc cc cc cc }
\toprule
\multirow{2}{*}{\textbf{Method}} & \multicolumn{2}{c}{\textbf{ARC-C}} & \multicolumn{2}{c}{\textbf{BBH}} & \multicolumn{2}{c}{\textbf{GSM8K}} & \multicolumn{2}{c}{\textbf{SeedBench}} & \multicolumn{2}{c}{\textbf{MMLU-Pro}} \\
\cmidrule(lr){2-3} \cmidrule(lr){4-5} \cmidrule(lr){6-7} \cmidrule(lr){8-9} \cmidrule(lr){10-11}
 & $\rho \uparrow$ & \scriptsize MAE $\downarrow$ & $\rho \uparrow$ & \scriptsize MAE $\downarrow$ & $\rho \uparrow$ & \scriptsize MAE $\downarrow$ & $\rho \uparrow$ & \scriptsize MAE $\downarrow$ & $\rho \uparrow$ & \scriptsize MAE $\downarrow$ \\
\midrule
\multicolumn{11}{c}{\textbf{$|\mathcal{S}| = 5$}} \\
\midrule
\textsc{AnchorPoints} & 0.650 & 0.269 & 0.618 & \underline{0.150} & 0.714 & 0.227 & 0.539 & \underline{0.215} & 0.413 & 0.182 \\
\textsc{GP-IRT} & \underline{0.817} & \underline{0.058} & \underline{0.853} & \textbf{0.054} & \underline{0.892} & \textbf{0.037} & \underline{0.829} & \textbf{0.047} & \underline{0.885} & \textbf{0.050} \\
\textsc{RepCore} & \textbf{0.894} & \textbf{0.039} & \textbf{0.887} & \textbf{0.054} & \textbf{0.922} & \underline{0.040} & \textbf{0.834} & \textbf{0.047} & \textbf{0.906} & \underline{0.051} \\
\midrule
\multicolumn{11}{c}{\textbf{$|\mathcal{S}| = 10$}} \\
\midrule
\textsc{AnchorPoints} & 0.797 & 0.145 & 0.854 & 0.081 & 0.864 & 0.117 & 0.818 & 0.103 & 0.861 & 0.095 \\
\textsc{GP-IRT} & \underline{0.866} & \underline{0.043} & \underline{0.869} & \underline{0.052} & \underline{0.865} & \underline{0.042} & \underline{0.824} & \textbf{0.043} & \underline{0.899} & \textbf{0.046} \\
\textsc{RepCore} & \textbf{0.900} & \textbf{0.036} & \textbf{0.898} & \textbf{0.049} & \textbf{0.926} & \textbf{0.035} & \textbf{0.853} & \underline{0.045} & \textbf{0.906} & \underline{0.048} \\
\midrule
\multicolumn{11}{c}{\textbf{$|\mathcal{S}| = 15$}} \\
\midrule
\textsc{AnchorPoints} & 0.843 & 0.102 & \textbf{0.896} & 0.068 & \underline{0.903} & 0.090 & \textbf{0.865} & 0.082 & \underline{0.894} & 0.081 \\
\textsc{GP-IRT} & \underline{0.901} & \textbf{0.035} & 0.887 & \textbf{0.047} & 0.897 & \textbf{0.036} & \underline{0.841} & \textbf{0.041} & 0.887 & \textbf{0.046} \\
\textsc{RepCore} & \textbf{0.903} & \underline{0.036} & \underline{0.894} & \underline{0.050} & \textbf{0.922} & \textbf{0.036} & 0.822 & \underline{0.045} & \textbf{0.908} & \underline{0.047} \\
\midrule
\multicolumn{11}{c}{\textbf{$|\mathcal{S}| = 20$}} \\
\midrule
\textsc{AnchorPoints} & 0.767 & 0.096 & \underline{0.906} & 0.065 & 0.892 & 0.080 & \textbf{0.883} & 0.070 & \underline{0.904} & 0.076 \\
\textsc{GP-IRT} & \underline{0.867} & \underline{0.037} & 0.894 & \textbf{0.046} & \underline{0.900} & \underline{0.038} & 0.813 & \textbf{0.042} & 0.864 & \underline{0.052} \\
\textsc{RepCore} & \textbf{0.903} & \textbf{0.035} & \textbf{0.911} & \underline{0.047} & \textbf{0.928} & \textbf{0.037} & \underline{0.831} & \underline{0.045} & \textbf{0.906} & \textbf{0.047} \\
\bottomrule
\end{tabular}
\end{table*}

\begin{table*}[htbp]
\centering
\footnotesize
\setlength{\tabcolsep}{4pt}
\caption{Detailed  results with budget $K=50$ under varying source pool sizes $|\mathcal{S}|$. We report Spearman's rank correlation ($\rho$, $\uparrow$) and Mean Absolute Error (MAE, $\downarrow$). \textbf{Bold} and \underline{underlined} values denote the best and second-best results, respectively.}
\label{tab:ablation_quantity_K50}
\begin{tabular}{l cc cc cc cc cc }
\toprule
\multirow{2}{*}{\textbf{Method}} & \multicolumn{2}{c}{\textbf{ARC-C}} & \multicolumn{2}{c}{\textbf{BBH}} & \multicolumn{2}{c}{\textbf{GSM8K}} & \multicolumn{2}{c}{\textbf{SeedBench}} & \multicolumn{2}{c}{\textbf{MMLU-Pro}} \\
\cmidrule(lr){2-3} \cmidrule(lr){4-5} \cmidrule(lr){6-7} \cmidrule(lr){8-9} \cmidrule(lr){10-11}
 & $\rho \uparrow$ & \scriptsize MAE $\downarrow$ & $\rho \uparrow$ & \scriptsize MAE $\downarrow$ & $\rho \uparrow$ & \scriptsize MAE $\downarrow$ & $\rho \uparrow$ & \scriptsize MAE $\downarrow$ & $\rho \uparrow$ & \scriptsize MAE $\downarrow$ \\
\midrule
\multicolumn{11}{c}{\textbf{$|\mathcal{S}| = 5$}} \\
\midrule
\textsc{AnchorPoints} & 0.636 & 0.255 & 0.622 & 0.151 & 0.718 & 0.246 & 0.498 & 0.215 & 0.647 & 0.162 \\
\textsc{GP-IRT} & \underline{0.819} & \underline{0.047} & \underline{0.880} & \textbf{0.049} & \underline{0.913} & \textbf{0.032} & \textbf{0.863} & \textbf{0.042} & \underline{0.906} & \textbf{0.045} \\
\textsc{RepCore} & \textbf{0.917} & \textbf{0.035} & \textbf{0.903} & \underline{0.050} & \textbf{0.936} & \underline{0.036} & \textbf{0.863} & \underline{0.043} & \textbf{0.922} & \underline{0.046} \\
\midrule
\multicolumn{11}{c}{\textbf{$|\mathcal{S}| = 10$}} \\
\midrule
\textsc{AnchorPoints} & 0.796 & 0.143 & 0.867 & 0.078 & 0.880 & 0.116 & 0.835 & 0.098 & 0.867 & 0.094 \\
\textsc{GP-IRT} & \underline{0.884} & \underline{0.036} & \underline{0.901} & \textbf{0.045} & \underline{0.896} & \underline{0.034} & \underline{0.863} & \textbf{0.038} & \underline{0.920} & \textbf{0.041} \\
\textsc{RepCore} & \textbf{0.921} & \textbf{0.032} & \textbf{0.913} & \textbf{0.045} & \textbf{0.938} & \textbf{0.032} & \textbf{0.874} & \underline{0.040} & \textbf{0.924} & \underline{0.043} \\
\midrule
\multicolumn{11}{c}{\textbf{$|\mathcal{S}| = 15$}} \\
\midrule
\textsc{AnchorPoints} & 0.830 & 0.118 & \textbf{0.904} & 0.065 & 0.913 & 0.089 & \textbf{0.877} & 0.078 & 0.906 & 0.079 \\
\textsc{GP-IRT} & \underline{0.919} & \textbf{0.030} & \textbf{0.911} & \textbf{0.042} & \underline{0.916} & \textbf{0.031} & \underline{0.870} & \textbf{0.037} & \underline{0.912} & \textbf{0.041} \\
\textsc{RepCore} & \textbf{0.921} & \underline{0.032} & \textbf{0.911} & \underline{0.045} & \textbf{0.936} & \underline{0.032} & 0.849 & \underline{0.041} & \textbf{0.926} & \underline{0.042} \\
\midrule
\multicolumn{11}{c}{\textbf{$|\mathcal{S}| = 20$}} \\
\midrule
\textsc{AnchorPoints} & 0.850 & 0.087 & \underline{0.919} & 0.059 & 0.904 & 0.079 & \textbf{0.893} & 0.066 & \underline{0.916} & 0.074 \\
\textsc{GP-IRT} & \underline{0.898} & \textbf{0.032} & 0.915 & \textbf{0.041} & \underline{0.910} & \underline{0.034} & 0.852 & \textbf{0.036} & 0.896 & \underline{0.045} \\
\textsc{RepCore} & \textbf{0.921} & \textbf{0.032} & \textbf{0.927} & \underline{0.043} & \textbf{0.941} & \textbf{0.033} & \underline{0.859} & \underline{0.040} & \textbf{0.924} & \textbf{0.041} \\
\bottomrule
\end{tabular}
\end{table*}

\subsection{Detailed Results of Source Composition Ablation}
\label{sec:family_ablation}
In this section, we provide the detailed breakdown of the source model composition ablation study across five datasets: ARC-Challenge, BBH, GSM8K, SeedBench, and MMLU-Pro. Table~\ref{tab:full_ablation_results} presents the Ranking Correlation (Spearman $\rho$) and Estimation Error (MAE) across coreset budgets $K \in \{10,20,30,40,50\}$.

\begin{table}[htbp]
    \centering
    \setlength{\parskip}{0.5cm}
    \caption{Detailed ablation study on source model composition across five datasets. We compare single model families (e.g., Qwen, Llama, Ovis, InternVL) against a \textbf{Diverse} source composition (Ours). We report Spearman’s rank correlation ($\rho$, $\uparrow$) and Mean Absolute Error (MAE, $\downarrow$). \textbf{Bold} and \underline{underlined} values denote the best and second-best results, respectively.}
    \label{tab:full_ablation_results}

    \begin{subtable}{0.9\textwidth}
        \centering
        \caption{ARC-Challenge}
        \resizebox{\textwidth}{!}{
        \begin{tabular}{lccccc|ccccc}
            \toprule
             & \multicolumn{5}{c|}{$\boldsymbol{\rho}$ $\uparrow$} & \multicolumn{5}{c}{\textbf{MAE} $\downarrow$} \\
            \textbf{Source Pool} & K=10 & K=20 & K=30 & K=40 & K=50 & K=10 & K=20 & K=30 & K=40 & K=50 \\
            \midrule
            Qwen & 0.658 & 0.800 & 0.853 & 0.888 & 0.908 & 0.081 & \underline{0.055} & \underline{0.046} & \underline{0.041} & \underline{0.037} \\
            Llama & \textbf{0.725} & \textbf{0.833} & \textbf{0.879} & \textbf{0.904} & \underline{0.921} & \underline{0.078} & 0.056 & 0.047 & 0.042 & 0.038 \\
            Diverse (Ours) & \underline{0.707} & \underline{0.820} & \underline{0.871} & \underline{0.900} & \textbf{0.921} & \textbf{0.072} & \textbf{0.050} & \textbf{0.041} & \textbf{0.036} & \textbf{0.032} \\
            \bottomrule
        \end{tabular}
        }
        
    \end{subtable}
    
    \begin{subtable}{0.9\textwidth}
        \centering
        \caption{BBH}
        \resizebox{\textwidth}{!}{
        \begin{tabular}{lccccc|ccccc}
            \toprule
             & \multicolumn{5}{c|}{$\boldsymbol{\rho}$ $\uparrow$} & \multicolumn{5}{c}{\textbf{MAE} $\downarrow$} \\
            \textbf{Source Pool} & K=10 & K=20 & K=30 & K=40 & K=50 & K=10 & K=20 & K=30 & K=40 & K=50 \\
            \midrule
            Qwen & 0.651 & 0.759 & 0.814 & 0.842 & 0.862 & 0.119 & 0.085 & 0.068 & 0.060 & 0.054 \\
            Llama & \underline{0.661} & \underline{0.811} & \underline{0.862} & \underline{0.888} & \underline{0.904} & \underline{0.099} & \underline{0.072} & \underline{0.060} & \underline{0.054} & \underline{0.048} \\
            Diverse (Ours) & \textbf{0.718} & \textbf{0.824} & \textbf{0.870} & \textbf{0.898} & \textbf{0.913} & \textbf{0.095} & \textbf{0.069} & \textbf{0.057} & \textbf{0.049} & \textbf{0.045} \\
            \bottomrule
        \end{tabular}
        }
        
    \end{subtable}
    
    \begin{subtable}{0.9\textwidth}
        \centering
        \caption{GSM8K}
        \resizebox{\textwidth}{!}{
        \begin{tabular}{lccccc|ccccc}
            \toprule
             & \multicolumn{5}{c|}{$\boldsymbol{\rho}$ $\uparrow$} & \multicolumn{5}{c}{\textbf{MAE} $\downarrow$} \\  
            \textbf{Source Pool} & K=10 & K=20 & K=30 & K=40 & K=50 & K=10 & K=20 & K=30 & K=40 & K=50 \\
            \midrule
            Qwen & 0.739 & 0.832 & 0.885 & 0.913 & 0.928 & 0.090 & \underline{0.065} & \underline{0.057} & \underline{0.052} & \underline{0.048} \\
            Llama & \underline{0.754} & \underline{0.854} & \underline{0.898} & \underline{0.919} & \underline{0.934} & \underline{0.087} & 0.067 & 0.060 & 0.054 & 0.050 \\
            Diverse (Ours) & \textbf{0.763} & \textbf{0.872} & \textbf{0.905} & \textbf{0.926} & \textbf{0.938} & \textbf{0.071} & \textbf{0.047} & \textbf{0.039} & \textbf{0.035} & \textbf{0.032} \\
            \bottomrule
        \end{tabular}
        }
        
    \end{subtable}
    
    \begin{subtable}{0.9\textwidth}
        \centering
        \caption{SeedBench}
        \resizebox{\textwidth}{!}{
        \begin{tabular}{lccccc|ccccc}
            \toprule
             & \multicolumn{5}{c|}{$\boldsymbol{\rho}$ $\uparrow$} & \multicolumn{5}{c}{\textbf{MAE} $\downarrow$} \\
            \textbf{Source Pool} & K=10 & K=20 & K=30 & K=40 & K=50 & K=10 & K=20 & K=30 & K=40 & K=50 \\
            \midrule
            Ovis & 0.597 & \textbf{0.756} & \textbf{0.826} & \textbf{0.864} & \textbf{0.883} & \underline{0.093} & \underline{0.063} & \underline{0.054} & \underline{0.049} & \underline{0.045} \\
            InternVL & \underline{0.618} & 0.741 & 0.802 & 0.837 & 0.859 & 0.094 & 0.073 & 0.065 & 0.059 & 0.055 \\
            Diverse (Ours) & \textbf{0.642} & \underline{0.755} & \underline{0.819} & \underline{0.853} & \underline{0.874} & \textbf{0.084} & \textbf{0.060} & \textbf{0.051} & \textbf{0.045} & \textbf{0.040} \\
            \bottomrule
        \end{tabular}
        }
        
    \end{subtable}
    
    \begin{subtable}{0.9\textwidth}
        \centering
        \caption{MMLU-Pro}
        \resizebox{\textwidth}{!}{
        \begin{tabular}{lccccc|ccccc}
            \toprule
             & \multicolumn{5}{c|}{$\boldsymbol{\rho}$ $\uparrow$} & \multicolumn{5}{c}{\textbf{MAE} $\downarrow$} \\
            \textbf{Source Pool} & K=10 & K=20 & K=30 & K=40 & K=50 & K=10 & K=20 & K=30 & K=40 & K=50 \\
            \midrule
            Qwen & 0.703 & 0.816 & 0.860 & 0.888 & 0.903 & 0.101 & 0.074 & 0.059 & 0.052 & 0.047 \\
            Llama & \underline{0.716} & \underline{0.823} & \textbf{0.876} & \underline{0.902} & \underline{0.921} & \underline{0.097} & \underline{0.070} & \underline{0.057} & \underline{0.050} & \underline{0.045} \\
            Diverse (Ours) & \textbf{0.732} & \textbf{0.829} & \underline{0.875} & \textbf{0.906} & \textbf{0.924} & \textbf{0.091} & \textbf{0.067} & \textbf{0.055} & \textbf{0.048} & \textbf{0.043} \\
            \bottomrule
        \end{tabular}
        }
    \end{subtable}
\end{table}

\subsection{Detailed Results of Source Capability Range Ablation}
\label{sec:capability_ablation}
In this section, we present the complete experimental results analyzing the impact of source model capability on compression performance. Table~\ref{tab:capability_ablation_results} provides a dataset-wise breakdown of the Ranking Correlation (Spearman $\rho$) and Estimation Error (MAE) across coreset budgets $K \in \{10,20,30,40,50\}$.

We compare three distinct source composition strategies:
\begin{itemize}[itemsep=2pt, topsep=2pt, parsep=0pt]
    \item \textbf{Strong:} Pools consisting exclusively of the top-10 strongest models.
    \item \textbf{Weak:} Pools consisting exclusively of the bottom-10 weakest models.
    \item \textbf{Diverse (Ours):} A heterogeneous mix spanning the full capability spectrum.
\end{itemize}

The detailed results reaffirm the analysis in the main text: The \textbf{Diverse} strategy consistently achieves the best trade-off between ranking accuracy and estimation stability across all five datasets.

\begin{table}[!htbp]
    \centering
    \setlength{\parskip}{0.5cm}
    \caption{Detailed ablation study on source capability range across five datasets. We compare homogeneous source pools consisting of only \textbf{Strong} or \textbf{Weak} models against a \textbf{Diverse} source composition (Ours). We report Spearman’s rank correlation ($\boldsymbol{\rho}$) and Mean Absolute Error (MAE). \textbf{Bold} and \underline{underlined} values denote the best and second-best results, respectively.}
    \label{tab:capability_ablation_results}

    \begin{subtable}{0.9\textwidth}
        \centering
        \caption{ARC-Challenge}
        \resizebox{\textwidth}{!}{
        \begin{tabular}{lccccc|ccccc}
            \toprule
             & \multicolumn{5}{c|}{$\boldsymbol{\rho}$ $\uparrow$} & \multicolumn{5}{c}{\textbf{MAE} $\downarrow$} \\
            \textbf{Source Pool} & K=10 & K=20 & K=30 & K=40 & K=50 & K=10 & K=20 & K=30 & K=40 & K=50 \\
            \midrule
            Strong & 0.479 & 0.659 & \underline{0.785} & 0.843 & 0.868 & 0.125 & 0.144 & 0.154 & 0.144 & 0.128 \\
            Weak & \underline{0.536} & \underline{0.713} & 0.782 & \underline{0.844} & \underline{0.873} & \underline{0.092} & \underline{0.055} & \underline{0.043} & \underline{0.037} & \textbf{0.032} \\
            Diverse (Ours) & \textbf{0.707} & \textbf{0.820} & \textbf{0.871} & \textbf{0.900} & \textbf{0.921} & \textbf{0.072} & \textbf{0.050} & \textbf{0.041} & \textbf{0.036} & \underline{0.032} \\
            \bottomrule
        \end{tabular}
        }
        
    \end{subtable}

    \begin{subtable}{0.9\textwidth}
        \centering
        \caption{BBH}
        \resizebox{\textwidth}{!}{
        \begin{tabular}{lccccc|ccccc}
            \toprule
             & \multicolumn{5}{c|}{$\boldsymbol{\rho}$ $\uparrow$} & \multicolumn{5}{c}{\textbf{MAE} $\downarrow$} \\
            \textbf{Source Pool} & K=10 & K=20 & K=30 & K=40 & K=50 & K=10 & K=20 & K=30 & K=40 & K=50 \\
            \midrule
            Strong & 0.502 & 0.705 & 0.770 & 0.785 & 0.823 & 0.148 & 0.116 & 0.073 & 0.066 & 0.062 \\
            Weak & \underline{0.654} & \underline{0.758} & \underline{0.817} & \underline{0.855} & \underline{0.890} & \underline{0.098} & \underline{0.075} & \underline{0.061} & \underline{0.054} & \underline{0.050} \\
            Diverse (Ours) & \textbf{0.718} & \textbf{0.824} & \textbf{0.870} & \textbf{0.898} & \textbf{0.913} & \textbf{0.095} & \textbf{0.069} & \textbf{0.057} & \textbf{0.049} & \textbf{0.045} \\
            \bottomrule
        \end{tabular}
        }
        
    \end{subtable}

    \begin{subtable}{0.9\textwidth}
        \centering
        \caption{GSM8K}
        \resizebox{\textwidth}{!}{
        \begin{tabular}{lccccc|ccccc}
            \toprule
             & \multicolumn{5}{c|}{$\boldsymbol{\rho}$ $\uparrow$} & \multicolumn{5}{c}{\textbf{MAE} $\downarrow$} \\
            \textbf{Source Pool} & K=10 & K=20 & K=30 & K=40 & K=50 & K=10 & K=20 & K=30 & K=40 & K=50 \\
            \midrule
            Strong & \underline{0.718} & 0.812 & \underline{0.858} & \underline{0.888} & \underline{0.908} & 0.120 & 0.094 & 0.079 & 0.065 & 0.056 \\
            Weak & 0.612 & \underline{0.817} & 0.839 & 0.874 & 0.905 & \underline{0.074} & \underline{0.049} & \underline{0.040} & \textbf{0.035} & \textbf{0.030} \\
            Diverse (Ours) & \textbf{0.763} & \textbf{0.872} & \textbf{0.905} & \textbf{0.926} & \textbf{0.938} & \textbf{0.071} & \textbf{0.047} & \textbf{0.039} & \underline{0.035} & \underline{0.032} \\
            \bottomrule
        \end{tabular}
        }
        
    \end{subtable}

    \begin{subtable}{0.9\textwidth}
        \centering
        \caption{SeedBench}
        \resizebox{\textwidth}{!}{
        \begin{tabular}{lccccc|ccccc}
            \toprule
             & \multicolumn{5}{c|}{$\boldsymbol{\rho}$ $\uparrow$} & \multicolumn{5}{c}{\textbf{MAE} $\downarrow$} \\
            \textbf{Source Pool} & K=10 & K=20 & K=30 & K=40 & K=50 & K=10 & K=20 & K=30 & K=40 & K=50 \\
            \midrule
            Strong & 0.626 & 0.694 & \underline{0.748} & \underline{0.802} & \underline{0.846} & \underline{0.089} & \underline{0.079} & 0.073 & 0.063 & 0.057 \\
            Weak & \textbf{0.683} & \underline{0.696} & 0.741 & 0.793 & 0.833 & 0.114 & 0.083 & \underline{0.063} & \underline{0.052} & \underline{0.051} \\
            Diverse (Ours) & \underline{0.642} & \textbf{0.755} & \textbf{0.819} & \textbf{0.853} & \textbf{0.874} & \textbf{0.084} & \textbf{0.060} & \textbf{0.051} & \textbf{0.045} & \textbf{0.040} \\
            \bottomrule
        \end{tabular}
        }
        
    \end{subtable}

    \begin{subtable}{0.9\textwidth}
        \centering
        \caption{MMLU-Pro}
        \resizebox{\textwidth}{!}{
        \begin{tabular}{lccccc|ccccc}
            \toprule
             & \multicolumn{5}{c|}{$\boldsymbol{\rho}$ $\uparrow$} & \multicolumn{5}{c}{\textbf{MAE} $\downarrow$} \\
            \textbf{Source Pool} & K=10 & K=20 & K=30 & K=40 & K=50 & K=10 & K=20 & K=30 & K=40 & K=50 \\
            \midrule
            Strong & 0.482 & 0.544 & 0.728 & 0.776 & 0.800 & \underline{0.119} & 0.094 & 0.079 & 0.081 & 0.070 \\
            Weak & \underline{0.614} & \underline{0.743} & \underline{0.815} & \underline{0.871} & \underline{0.896} & 0.119 & \underline{0.082} & \underline{0.067} & \underline{0.056} & \underline{0.049} \\
            Diverse (Ours) & \textbf{0.732} & \textbf{0.829} & \textbf{0.875} & \textbf{0.906} & \textbf{0.924} & \textbf{0.091} & \textbf{0.067} & \textbf{0.055} & \textbf{0.048} & \textbf{0.043} \\
            \bottomrule
        \end{tabular}
        }
    \end{subtable}
\end{table}

\subsection{Detailed Results on Coreset Size Generalization}
\label{app:coreset_size_generalization}

In this section, we provide the comprehensive numerical results for the coreset size generalization experiments. We evaluate the robustness of different selection methods across three coreset sizes (100, 150, and 200 points). Table~\ref{tab:coreset_detailed} presents the detailed breakdown, with Panel A showing the Spearman's rank correlation ($\rho$) and Panel B showing the Mean Absolute Error (MAE) on five benchmarks.

\begin{table*}[!htbp]
\caption{Detailed generalization results with varying coreset sizes. We report Spearman's $\rho$ and MAE on five benchmarks. The best results are \textbf{bolded} and the second best are \underline{underlined}.}
\label{tab:coreset_detailed}
\begin{center}
\begin{tabular}{llccccc}
\toprule
Size & Method & ARC & BBH & GSM8K & MMLU & SeedBench \\
\midrule
\multicolumn{7}{c}{\textbf{Panel A: $\rho$ ($\uparrow$)}} \\
\midrule
\multirow{4}{*}{100} & \textsc{Random} & 0.953 & \underline{0.948} & \underline{0.958} & 0.953 & 0.921 \\
 & \textsc{APW} & 0.791 & 0.883 & 0.838 & 0.881 & 0.770 \\
 & \textsc{GP-IRT} & \underline{0.957} & \underline{0.948} & 0.955 & \textbf{0.960} & \textbf{0.934} \\
 & \textsc{RepCore} & \textbf{0.960} & \textbf{0.954} & \textbf{0.967} & \underline{0.959} & \underline{0.928} \\
\midrule
\multirow{4}{*}{150} & \textsc{Random} & 0.969 & \underline{0.964} & \underline{0.972} & 0.968 & 0.947 \\
 & \textsc{APW} & 0.777 & 0.888 & 0.829 & 0.884 & 0.781 \\
 & \textsc{GP-IRT} & \underline{0.972} & 0.961 & 0.968 & \textbf{0.972} & \textbf{0.957} \\
 & \textsc{RepCore} & \textbf{0.973} & \textbf{0.967} & \textbf{0.977} & \underline{0.971} & \underline{0.948} \\
\midrule
\multirow{4}{*}{200} & \textsc{Random} & 0.977 & \underline{0.972} & \underline{0.979} & 0.975 & \underline{0.959} \\
 & \textsc{APW} & 0.784 & 0.890 & 0.824 & 0.887 & 0.796 \\
 & \textsc{GP-IRT} & \underline{0.979} & 0.969 & 0.973 & \underline{0.977} & \textbf{0.967} \\
 & \textsc{RepCore} & \textbf{0.981} & \textbf{0.973} & \textbf{0.982} & \textbf{0.978} & \underline{0.959} \\
\midrule
\midrule
\multicolumn{7}{c}{\textbf{Panel B: MAE ($\downarrow$)}} \\
\midrule
\multirow{4}{*}{100} & \textsc{Random} & 0.0267 & 0.0378 & 0.0261 & 0.0385 & 0.0377 \\
 & \textsc{APW} & 0.1430 & 0.0716 & 0.1480 & 0.0885 & 0.1040 \\
 & \textsc{GP-IRT} & \underline{0.0228} & \textbf{0.0318} & \underline{0.0237} & \textbf{0.0282} & \textbf{0.0274} \\
 & \textsc{RepCore} & \textbf{0.0227} & \underline{0.0328} & \textbf{0.0233} & \underline{0.0314} & \underline{0.0288} \\
\midrule
\multirow{4}{*}{150} & \textsc{Random} & 0.0211 & 0.0306 & \underline{0.0206} & 0.0312 & 0.0303 \\
 & \textsc{APW} & 0.1350 & 0.0704 & 0.1460 & 0.0866 & 0.0953 \\
 & \textsc{GP-IRT} & \underline{0.0202} & \underline{0.0289} & 0.0215 & \textbf{0.0236} & \textbf{0.0222} \\
 & \textsc{RepCore} & \textbf{0.0189} & \textbf{0.0271} & \textbf{0.0194} & \underline{0.0261} & \underline{0.0239} \\
\midrule
\multirow{4}{*}{200} & \textsc{Random} & \underline{0.0175} & \underline{0.0264} & \underline{0.0176} & 0.0271 & 0.0255 \\
 & \textsc{APW} & 0.1210 & 0.0685 & 0.1330 & 0.0855 & 0.0850 \\
 & \textsc{GP-IRT} & 0.0189 & 0.0270 & 0.0204 & \textbf{0.0213} & \textbf{0.0194} \\
 & \textsc{RepCore} & \textbf{0.0168} & \textbf{0.0239} & \textbf{0.0173} & \underline{0.0231} & \underline{0.0216} \\
\bottomrule
\end{tabular}
\end{center}
\end{table*}

\subsection{Robustness to Temporal Drift}
\label{sec:temporal_drift}

This section reports the complete results for the temporal distribution shift setting described in Section~\ref{sec:detailed_analysis}. Table~\ref{tab:temporal_drift_detailed} summarizes Spearman's $\rho$ and MAE across coreset budgets $K \in \{10,20,30,40,50\}$ on five benchmarks. The results are consistent with the trend observed in the main text.

\begin{table*}[!htbp]
\caption{Complete results under temporal distribution shift.
Spearman's $\rho$ ($\uparrow$) and MAE ($\downarrow$) are reported across coreset budgets $K \in \{10,20,30,40,50\}$. The best results are \textbf{bolded} and the second best are \underline{underlined}.}
\label{tab:temporal_drift_detailed}
\begin{center}
\begin{tabular}{llccccc}
\toprule
K & Method & ARC & BBH & GSM8K & MMLU Pro & SeedBench \\
\midrule
\multicolumn{7}{c}{\textbf{Panel A: $\rho$ ($\uparrow$)}} \\
\midrule
\multirow{4}{*}{10} & \textsc{Random} & 0.676 & 0.706 & 0.714 & \textbf{0.735} & 0.555 \\
 & \textsc{APW} & \textbf{0.705} & \underline{0.716} & \underline{0.758} & 0.707 & \textbf{0.577} \\
 & \textsc{GP-IRT} & \underline{0.698} & 0.704 & 0.724 & \underline{0.725} & 0.549 \\
 & \textsc{RepCore} & 0.673 & \textbf{0.719} & \textbf{0.772} & 0.716 & \underline{0.574} \\
\midrule
\multirow{4}{*}{20} & \textsc{Random} & 0.791 & \underline{0.817} & 0.822 & \textbf{0.842} & \underline{0.674} \\
 & \textsc{APW} & 0.779 & 0.774 & 0.815 & 0.803 & 0.648 \\
 & \textsc{GP-IRT} & \textbf{0.822} & 0.786 & \underline{0.826} & 0.825 & \textbf{0.679} \\
 & \textsc{RepCore} & \underline{0.800} & \textbf{0.844} & \textbf{0.870} & \underline{0.831} & 0.666 \\
\midrule
\multirow{4}{*}{30} & \textsc{Random} & 0.848 & \underline{0.868} & \underline{0.870} & \textbf{0.890} & \textbf{0.747} \\
 & \textsc{APW} & 0.806 & 0.819 & 0.838 & 0.830 & 0.691 \\
 & \textsc{GP-IRT} & \textbf{0.865} & 0.845 & 0.863 & 0.870 & 0.733 \\
 & \textsc{RepCore} & \underline{0.861} & \textbf{0.885} & \textbf{0.908} & \underline{0.878} & \underline{0.735} \\
\midrule
\multirow{4}{*}{40} & \textsc{Random} & 0.880 & \underline{0.896} & \underline{0.898} & \textbf{0.912} & \textbf{0.792} \\
 & \textsc{APW} & 0.820 & 0.827 & 0.855 & 0.843 & 0.728 \\
 & \textsc{GP-IRT} & \underline{0.886} & 0.886 & 0.893 & 0.899 & \underline{0.791} \\
 & \textsc{RepCore} & \textbf{0.896} & \textbf{0.909} & \textbf{0.927} & \underline{0.905} & 0.779 \\
\midrule
\multirow{4}{*}{50} & \textsc{Random} & 0.901 & \underline{0.914} & \underline{0.917} & \textbf{0.928} & \underline{0.820} \\
 & \textsc{APW} & 0.820 & 0.836 & 0.858 & 0.867 & 0.772 \\
 & \textsc{GP-IRT} & \underline{0.908} & 0.906 & 0.916 & 0.922 & \textbf{0.828} \\
 & \textsc{RepCore} & \textbf{0.914} & \textbf{0.920} & \textbf{0.942} & \underline{0.923} & 0.812 \\
\midrule
\midrule
\multicolumn{7}{c}{\textbf{Panel B: MAE ($\downarrow$)}} \\
\midrule
\multirow{4}{*}{10} & \textsc{Random} & 0.0832 & 0.1180 & 0.0833 & 0.1220 & 0.1180 \\
 & \textsc{APW} & 0.1140 & 0.1400 & 0.1250 & 0.1420 & 0.1220 \\
 & \textsc{GP-IRT} & \textbf{0.0671} & \underline{0.1060} & \textbf{0.0711} & \underline{0.1030} & \underline{0.0934} \\
 & \textsc{RepCore} & \underline{0.0742} & \textbf{0.0994} & \underline{0.0724} & \textbf{0.0993} & \textbf{0.0912} \\
\midrule
\multirow{4}{*}{20} & \textsc{Random} & 0.0591 & 0.0837 & 0.0576 & 0.0865 & 0.0840 \\
 & \textsc{APW} & 0.1100 & 0.1180 & 0.1250 & 0.1190 & 0.0992 \\
 & \textsc{GP-IRT} & \textbf{0.0476} & \textbf{0.0703} & \underline{0.0519} & \underline{0.0711} & \textbf{0.0614} \\
 & \textsc{RepCore} & \underline{0.0479} & \underline{0.0726} & \textbf{0.0516} & \textbf{0.0694} & \underline{0.0648} \\
\midrule
\multirow{4}{*}{30} & \textsc{Random} & 0.0484 & 0.0690 & 0.0469 & 0.0704 & 0.0671 \\
 & \textsc{APW} & 0.1130 & 0.1070 & 0.1290 & 0.1120 & 0.0912 \\
 & \textsc{GP-IRT} & \underline{0.0398} & \underline{0.0592} & \underline{0.0434} & \textbf{0.0565} & \textbf{0.0477} \\
 & \textsc{RepCore} & \textbf{0.0386} & \textbf{0.0581} & \textbf{0.0414} & \underline{0.0574} & \underline{0.0525} \\
\midrule
\multirow{4}{*}{40} & \textsc{Random} & 0.0418 & 0.0587 & 0.0406 & 0.0610 & 0.0573 \\
 & \textsc{APW} & 0.1130 & 0.1030 & 0.1270 & 0.1080 & 0.0863 \\
 & \textsc{GP-IRT} & \underline{0.0352} & \textbf{0.0500} & \textbf{0.0357} & \textbf{0.0492} & \textbf{0.0400} \\
 & \textsc{RepCore} & \textbf{0.0325} & \underline{0.0507} & \underline{0.0366} & \underline{0.0498} & \underline{0.0460} \\
\midrule
\multirow{4}{*}{50} & \textsc{Random} & 0.0368 & 0.0525 & 0.0356 & 0.0550 & 0.0537 \\
 & \textsc{APW} & 0.1110 & 0.1000 & 0.1220 & 0.0945 & 0.0854 \\
 & \textsc{GP-IRT} & \underline{0.0312} & \textbf{0.0441} & \textbf{0.0307} & \textbf{0.0429} & \textbf{0.0361} \\
 & \textsc{RepCore} & \textbf{0.0296} & \underline{0.0461} & \underline{0.0326} & \underline{0.0453} & \underline{0.0411} \\
\bottomrule
\end{tabular}
\end{center}
\end{table*}

\subsection{Source Hidden-State Masking Analysis}
\label{app:closed_source_masking}

This section details the mixed-source simulation used in Section~\ref{sec:limitations}. We first describe the pseudo-representation construction for masked sources and the two masking methods, then report the full masking results together with additional analyses of robustness and sensitivity.

\subsubsection{Pseudo-representation construction.}
Let $O$ denote the sources with accessible hidden states and $M$ denote the sources whose hidden states are masked, with $S=O\cup M$. For each masked source $m\in M$, we approximate its item representation using accessible sources with similar output behavior. Specifically, we compute the output-behavior similarity and normalize it as
\begin{equation}
\mathrm{sim}(m,o)=\cos(y_m,y_o), 
\qquad
\alpha_{m,o}=\mathrm{softmax}_{o\in O}\bigl(\mathrm{sim}(m,o)\bigr),
\end{equation}
where $y_m$ and $y_o$ are binary correctness vectors over the full benchmark. We then construct the pseudo-representation of masked source $m$ on item $i$ as
\begin{equation}
\hat{z}_{m,i}=\sum_{o\in O}\alpha_{m,o} \cdot z_{o,i}.
\end{equation}
The consensus embedding used for clustering combines real aligned embeddings from accessible sources with pseudo-representations for masked sources:
\begin{equation}
e_i=\frac{1}{|S|}
\left(
\sum_{o\in O}z_{o,i}
+
\sum_{m\in M}\hat{z}_{m,i}
\right).
\end{equation}
This replacement is applied only during coreset selection, while the extrapolation stage remains unchanged. Thus, the pseudo-representation is a practical approximation for partially accessible source pools rather than a substitute for true hidden-state access.

\subsubsection{Masking methods.}
We consider two masking methods: Random Mask removes source hidden states uniformly at random and averages over 10 masking draws; Advdist Mask instead targets the sources that are least redundant with the rest of the pool, providing a stricter stress test. For each source model $s$, we compute its behavioral distinctness score as
\begin{equation}
d(s)=\frac{1}{|S|-1}\sum_{s'\neq s}\left(1-\cos(y_s,y_{s'})\right),
\end{equation}
where $y_s$ denotes the binary correctness vector of source model $s$ over the full benchmark. Advdist Mask removes sources in descending order of $d(s)$, so the masked sources are expected to be the hardest to approximate from the remaining accessible sources.

\subsubsection{Full masking results analysis}
This part provides the full breakdown of the hidden state unavailable source experiments. 
Tables~\ref{tab:masking_mean_10pct}--\ref{tab:masking_mean_50pct} report mean performance under masking ratios from 10\% to 50\%, covering five benchmarks and coreset budgets $K\in\{10,20,30,40,50\}$. Each table reports Spearman's $\rho$ and MAE averaged over 10 source-model combinations and 50 random anchor selections. For \textsc{Random Pseudo}, results are further averaged over 10 stochastic masking draws; \textsc{RepCore} and \textsc{Advdist Pseudo} are deterministic under each source-model combination. The \textsc{Best Baseline} row denotes the strongest output-based baseline under the corresponding setting.

Tables~\ref{tab:masking_combo_std_10pct}--\ref{tab:masking_combo_std_50pct} report $Combo_{std}$, measuring variation across the 10 source-model combinations. For \textsc{Random Pseudo}, each combination-level score is first averaged over 10 random masking draws before computing $Combo_{std}$. Since this additional averaging structurally reduces variance, its standard deviations are reported as reference values. In contrast, \textsc{RepCore} and \textsc{Advdist Pseudo} are deterministic for each source-model combination, so their $Combo_{std}$ values are directly comparable. Table~\ref{tab:masking_draw_std_combined} further reports $Draw_{std}$ for \textsc{Random Pseudo}, isolating the sensitivity to which source models are randomly masked.

\paragraph{Random masking remains competitive.}
Under random masking, the pseudo-representation strategy degrades gradually as the masking ratio increases. Across 125 breakdown cells, corresponding to 5 benchmarks $\times$ 5 budgets $\times$ 5 masking ratios, \textsc{Random Pseudo} outperforms the strongest output-based baseline in 89 cells. The number of wins remains high across masking ratios, with 20, 19, 18, 19, and 13 wins out of 25 cells from 10\% to 50\% masking, respectively. Even at 50\% masking, its average Spearman correlation remains above all output-based baselines. This suggests that randomly removed source representations are often approximable from the remaining accessible sources, especially when the remaining pool still covers similar model behaviors.

\paragraph{Adversarial masking exposes the boundary of pseudo-representations.}
The flat trend under random masking does not hold when the unavailable sources are selected adversarially. \textsc{Advdist Pseudo} masks the most output-distinct sources first and therefore provides a stricter stress test for mixed source pools. Its degradation is small at 10\%--20\% masking, becomes visible from 30\%, and grows substantially at 50\%. Correspondingly, the win rate against the strongest output-based baseline drops from 19/25 cells at 10\% masking to 13/25 at 30\% and 4/25 at 50\%. MAE shows the same pattern: from 30\% masking onward, \textsc{Advdist Pseudo} has higher MAE than \textsc{Random Pseudo} in all 25 per-benchmark $\times$ per-budget cells. This suggests that pseudo-representations are less reliable when the masked sources are behaviorally distinctive, since the item representations are harder to approximate from the remaining accessible sources using output-similarity weights.

\paragraph{Sensitivity varies across benchmarks.}
The effect of adversarial masking is benchmark-dependent. At 50\% masking, the Spearman gap between \textsc{Random Pseudo} and \textsc{Advdist Pseudo}, averaged across budgets, follows
BBH $(+.048)$ $>$ MMLU-Pro $(+.033)$ $>$ SEED-Bench-2-Plus $(+.025)$ $>$ ARC-Challenge $(+.022)$ $>$ GSM8K $(+.012)$. The MAE gap follows a similar trend, with MMLU-Pro $(+.008)$ and BBH $(+.008)$ showing the largest degradation, followed by SEED-Bench-2-Plus $(+.007)$, ARC-Challenge $(+.005)$, and GSM8K $(+.003)$. These patterns suggest that task-heterogeneous benchmarks are more sensitive to the loss of behaviorally distinctive source models, as their broader task coverage makes the remaining accessible sources less likely to provide good approximations across all item groups.

Taken together, these results support a bounded view of mixed source pools: pseudo-representations provide a useful partial-access heuristic under random or moderate missingness, but the approximation quality degrades when the hidden states of behaviorally distinctive sources become unavailable.

\begin{table*}[!htbp]
    \centering
    \footnotesize
    \setlength{\tabcolsep}{2.2pt}
    \renewcommand{\arraystretch}{1.1}

    \captionof{table}{\textbf{Mean performance under 10\% masking.} Spearman $\rho$ ($\uparrow$) and MAE ($\downarrow$) are averaged over 10 source-model combinations $\times$ 50 random anchor selections. For Random Pseudo, scores are further averaged over 10 stochastic masking draws; \textsc{RepCore} and \textsc{Advdist Pseudo} are deterministic under each source-model combination. \textbf{Bold} indicates the best and \uline{underline} the second best per column.}
    \label{tab:masking_mean_10pct}

    \begin{tabular}{l l cc cc cc cc cc cc}
    \toprule
    \multirow{2}{*}{\textbf{Benchmark}} & \multirow{2}{*}{\textbf{Condition}} &
    \multicolumn{2}{c}{\textbf{$K=10$}} & \multicolumn{2}{c}{\textbf{$K=20$}} & \multicolumn{2}{c}{\textbf{$K=30$}} &
    \multicolumn{2}{c}{\textbf{$K=40$}} & \multicolumn{2}{c}{\textbf{$K=50$}} & \multicolumn{2}{c}{\textbf{Avg}} \\
    \cmidrule(lr){3-4}\cmidrule(lr){5-6}\cmidrule(lr){7-8}\cmidrule(lr){9-10}\cmidrule(lr){11-12}\cmidrule(lr){13-14}
    & & $\rho$ & {MAE} & $\rho$ & {MAE} & $\rho$ & {MAE} & $\rho$ & {MAE} & $\rho$ & {MAE} & $\rho$ & {MAE} \\
    \midrule

    \multirow{4}{*}{\textbf{ARC-Challenge}}
          & \textsc{REPCORE} & 0.7070 & {\uline{0.0720}} & 0.8200 & {\textbf{0.0500}} & 0.8710 & {\textbf{0.0409}} & 0.9000 & {\textbf{0.0358}} & \uline{0.9210} & {\textbf{0.0321}} & 0.8438 & {\textbf{0.0462}} \\
          & \textsc{Random Pseudo} & 0.7267 & {0.0754} & \textbf{0.8358} & {\uline{0.0543}} & \textbf{0.8821} & {\uline{0.0449}} & \textbf{0.9083} & {\uline{0.0390}} & \textbf{0.9241} & {\uline{0.0349}} & \textbf{0.8554} & {\uline{0.0497}} \\
          & \textsc{Advdist Pseudo} & \uline{0.7466} & {0.0797} & \uline{0.8303} & {0.0567} & \uline{0.8744} & {0.0467} & \uline{0.9018} & {0.0410} & 0.9188 & {0.0376} & \uline{0.8544} & {0.0523} \\
          & \textsc{Best Baseline} & \textbf{0.7520} & {\textbf{0.0717}} & 0.8230 & {0.0573} & 0.8640 & {0.0505} & 0.8900 & {0.0432} & 0.9180 & {0.0357} & 0.8494 & {0.0517} \\
    \midrule

    \multirow{4}{*}{\textbf{BBH}}
          & \textsc{REPCORE} & 0.7180 & {\uline{0.0950}} & \uline{0.8240} & {\uline{0.0687}} & \uline{0.8700} & {\textbf{0.0569}} & \textbf{0.8980} & {\textbf{0.0494}} & \textbf{0.9130} & {\textbf{0.0452}} & \uline{0.8446} & {\textbf{0.0630}} \\
          & \textsc{Random Pseudo} & \uline{0.7313} & {0.1051} & \textbf{0.8263} & {0.0742} & \textbf{0.8714} & {0.0600} & \uline{0.8963} & {0.0525} & \uline{0.9129} & {0.0476} & \textbf{0.8476} & {0.0679} \\
          & \textsc{Advdist Pseudo} & \textbf{0.7336} & {0.1051} & 0.8216 & {0.0733} & 0.8613 & {0.0618} & 0.8925 & {0.0532} & 0.9073 & {0.0477} & 0.8433 & {0.0682} \\
          & \textsc{Best Baseline} & 0.7010 & {\textbf{0.0931}} & 0.7990 & {\textbf{0.0678}} & 0.8520 & {\uline{0.0583}} & 0.8840 & {\uline{0.0516}} & 0.9050 & {\uline{0.0454}} & 0.8282 & {\uline{0.0632}} \\
    \midrule

    \multirow{4}{*}{\textbf{GSM8K}}
          & \textsc{REPCORE} & \textbf{0.7630} & {\uline{0.0714}} & \textbf{0.8720} & {\textbf{0.0473}} & \textbf{0.9050} & {\textbf{0.0393}} & \textbf{0.9260} & {\textbf{0.0350}} & \textbf{0.9380} & {\textbf{0.0316}} & \textbf{0.8808} & {\textbf{0.0449}} \\
          & \textsc{Random Pseudo} & \uline{0.7593} & {0.0735} & \uline{0.8578} & {0.0556} & \uline{0.9010} & {\uline{0.0463}} & \uline{0.9230} & {\uline{0.0407}} & \uline{0.9367} & {0.0369} & \uline{0.8756} & {0.0506} \\
          & \textsc{Advdist Pseudo} & 0.7505 & {0.0738} & 0.8545 & {\uline{0.0555}} & 0.8944 & {0.0488} & 0.9185 & {0.0434} & 0.9333 & {0.0399} & 0.8702 & {0.0523} \\
          & \textsc{Best Baseline} & 0.7500 & {\textbf{0.0644}} & 0.8390 & {0.0569} & 0.8700 & {0.0488} & 0.8980 & {0.0418} & 0.9170 & {\uline{0.0336}} & 0.8548 & {\uline{0.0491}} \\
    \midrule

    \multirow{4}{*}{\textbf{MMLU-Pro}}
          & \textsc{REPCORE} & \uline{0.7320} & {\textbf{0.0910}} & 0.8290 & {\uline{0.0666}} & 0.8750 & {\uline{0.0548}} & 0.9060 & {\uline{0.0476}} & \textbf{0.9240} & {\uline{0.0426}} & \textbf{0.8532} & {\uline{0.0605}} \\
          & \textsc{Random Pseudo} & 0.7140 & {0.0958} & \textbf{0.8327} & {0.0692} & \textbf{0.8825} & {0.0560} & \uline{0.9078} & {0.0491} & \uline{0.9236} & {0.0441} & 0.8521 & {0.0628} \\
          & \textsc{Advdist Pseudo} & 0.7212 & {0.0936} & \uline{0.8314} & {0.0688} & \uline{0.8811} & {0.0556} & \textbf{0.9086} & {0.0491} & 0.9233 & {0.0441} & \uline{0.8531} & {0.0622} \\
          & \textsc{Best Baseline} & \textbf{0.7540} & {\uline{0.0917}} & 0.8120 & {\textbf{0.0651}} & 0.8660 & {\textbf{0.0528}} & 0.8990 & {\textbf{0.0457}} & 0.9200 & {\textbf{0.0405}} & 0.8502 & {\textbf{0.0592}} \\
    \midrule

    \multirow{4}{*}{\shortstack[l]{\textbf{SEED-Bench}\\\textbf{-2-Plus}}}
          & \textsc{REPCORE} & \uline{0.6420} & {\uline{0.0835}} & 0.7550 & {\textbf{0.0600}} & \textbf{0.8190} & {0.0508} & \textbf{0.8530} & {0.0446} & \textbf{0.8740} & {0.0402} & \uline{0.7886} & {\uline{0.0558}} \\
          & \textsc{Random Pseudo} & 0.6239 & {0.0881} & 0.7504 & {0.0624} & 0.8088 & {0.0514} & \uline{0.8435} & {0.0452} & \uline{0.8682} & {0.0407} & 0.7790 & {0.0576} \\
          & \textsc{Advdist Pseudo} & 0.6302 & {0.0894} & \uline{0.7651} & {\uline{0.0621}} & 0.8057 & {\textbf{0.0500}} & 0.8370 & {\uline{0.0442}} & 0.8628 & {\uline{0.0399}} & 0.7802 & {0.0571} \\
          & \textsc{Best Baseline} & \textbf{0.6760} & {\textbf{0.0810}} & \textbf{0.7760} & {0.0628} & \uline{0.8130} & {\uline{0.0500}} & 0.8360 & {\textbf{0.0428}} & 0.8630 & {\textbf{0.0378}} & \textbf{0.7928} & {\textbf{0.0549}} \\
    \midrule

    \multirow{4}{*}{\textbf{Average}}
          & \textsc{REPCORE} & 0.7124 & {\uline{0.0826}} & 0.8200 & {\textbf{0.0585}} & \uline{0.8680} & {\textbf{0.0485}} & \textbf{0.8966} & {\textbf{0.0425}} & \textbf{0.9140} & {\textbf{0.0383}} & \textbf{0.8422} & {\textbf{0.0541}} \\
          & \textsc{Random Pseudo} & 0.7110 & {0.0876} & \textbf{0.8206} & {0.0631} & \textbf{0.8692} & {\uline{0.0517}} & \uline{0.8958} & {0.0453} & \uline{0.9131} & {0.0408} & \uline{0.8419} & {0.0577} \\
          & \textsc{Advdist Pseudo} & \uline{0.7164} & {0.0883} & \uline{0.8206} & {0.0633} & 0.8634 & {0.0526} & 0.8917 & {0.0462} & 0.9091 & {0.0418} & 0.8402 & {0.0584} \\
          & \textsc{Best Baseline} & \textbf{0.7266} & {\textbf{0.0804}} & 0.8098 & {\uline{0.0620}} & 0.8530 & {0.0521} & 0.8814 & {\uline{0.0450}} & 0.9046 & {\uline{0.0386}} & 0.8351 & {\uline{0.0556}} \\

    \bottomrule
    \end{tabular}
\end{table*}

\begin{table*}[!htbp]
    \centering
    \footnotesize
    \setlength{\tabcolsep}{2.2pt}
    \renewcommand{\arraystretch}{1.1}

    \captionof{table}{\textbf{Mean performance under 20\% masking.} Spearman $\rho$ ($\uparrow$) and MAE ($\downarrow$) are averaged over 10 source-model combinations $\times$ 50 random anchor selections. For Random Pseudo, scores are further averaged over 10 stochastic masking draws; \textsc{RepCore} and \textsc{Advdist Pseudo} are deterministic under each source-model combination. \textbf{Bold} indicates the best and \uline{underline} the second best per column.}
    \label{tab:masking_mean_20pct}

    \begin{tabular}{l l cc cc cc cc cc cc}
    \toprule
    \multirow{2}{*}{\textbf{Benchmark}} & \multirow{2}{*}{\textbf{Condition}} &
    \multicolumn{2}{c}{\textbf{$K=10$}} & \multicolumn{2}{c}{\textbf{$K=20$}} & \multicolumn{2}{c}{\textbf{$K=30$}} &
    \multicolumn{2}{c}{\textbf{$K=40$}} & \multicolumn{2}{c}{\textbf{$K=50$}} & \multicolumn{2}{c}{\textbf{Avg}} \\
    \cmidrule(lr){3-4}\cmidrule(lr){5-6}\cmidrule(lr){7-8}\cmidrule(lr){9-10}\cmidrule(lr){11-12}\cmidrule(lr){13-14}
    & & $\rho$ & {MAE} & $\rho$ & {MAE} & $\rho$ & {MAE} & $\rho$ & {MAE} & $\rho$ & {MAE} & $\rho$ & {MAE} \\
    \midrule

    \multirow{4}{*}{\textbf{ARC-Challenge}}
          & \textsc{REPCORE} & 0.7070 & {\uline{0.0720}} & 0.8200 & {\textbf{0.0500}} & 0.8710 & {\textbf{0.0409}} & 0.9000 & {\textbf{0.0358}} & \uline{0.9210} & {\textbf{0.0321}} & 0.8438 & {\textbf{0.0462}} \\
          & \textsc{Random Pseudo} & 0.7288 & {0.0772} & \textbf{0.8401} & {\uline{0.0557}} & \textbf{0.8851} & {\uline{0.0457}} & \textbf{0.9103} & {\uline{0.0398}} & \textbf{0.9265} & {\uline{0.0357}} & \textbf{0.8582} & {\uline{0.0508}} \\
          & \textsc{Advdist Pseudo} & \uline{0.7407} & {0.0843} & \uline{0.8373} & {0.0632} & \uline{0.8797} & {0.0512} & \uline{0.9055} & {0.0448} & 0.9208 & {0.0399} & \uline{0.8568} & {0.0567} \\
          & \textsc{Best Baseline} & \textbf{0.7520} & {\textbf{0.0717}} & 0.8230 & {0.0573} & 0.8640 & {0.0505} & 0.8900 & {0.0432} & 0.9180 & {0.0357} & 0.8494 & {0.0517} \\
    \midrule

    \multirow{4}{*}{\textbf{BBH}}
          & \textsc{REPCORE} & \uline{0.7180} & {\uline{0.0950}} & \uline{0.8240} & {\uline{0.0687}} & \uline{0.8700} & {\textbf{0.0569}} & \textbf{0.8980} & {\textbf{0.0494}} & \uline{0.9130} & {\textbf{0.0452}} & \uline{0.8446} & {\textbf{0.0630}} \\
          & \textsc{Random Pseudo} & \textbf{0.7226} & {0.1054} & \textbf{0.8289} & {0.0744} & \textbf{0.8704} & {0.0615} & \uline{0.8979} & {0.0538} & \textbf{0.9143} & {0.0487} & \textbf{0.8468} & {0.0688} \\
          & \textsc{Advdist Pseudo} & 0.6851 & {0.1098} & 0.8231 & {0.0767} & 0.8632 & {0.0639} & 0.8870 & {0.0560} & 0.9084 & {0.0494} & 0.8334 & {0.0712} \\
          & \textsc{Best Baseline} & 0.7010 & {\textbf{0.0931}} & 0.7990 & {\textbf{0.0678}} & 0.8520 & {\uline{0.0583}} & 0.8840 & {\uline{0.0516}} & 0.9050 & {\uline{0.0454}} & 0.8282 & {\uline{0.0632}} \\
    \midrule

    \multirow{4}{*}{\textbf{GSM8K}}
          & \textsc{REPCORE} & \uline{0.7630} & {\uline{0.0714}} & \textbf{0.8720} & {\textbf{0.0473}} & \textbf{0.9050} & {\textbf{0.0393}} & \textbf{0.9260} & {\textbf{0.0350}} & \uline{0.9380} & {\textbf{0.0316}} & \textbf{0.8808} & {\textbf{0.0449}} \\
          & \textsc{Random Pseudo} & \textbf{0.7680} & {0.0724} & \uline{0.8626} & {\uline{0.0534}} & \uline{0.9029} & {\uline{0.0449}} & \uline{0.9252} & {\uline{0.0397}} & \textbf{0.9381} & {0.0363} & \uline{0.8794} & {0.0493} \\
          & \textsc{Advdist Pseudo} & 0.7547 & {0.0741} & 0.8607 & {0.0565} & 0.9015 & {0.0486} & 0.9195 & {0.0436} & 0.9337 & {0.0399} & 0.8740 & {0.0525} \\
          & \textsc{Best Baseline} & 0.7500 & {\textbf{0.0644}} & 0.8390 & {0.0569} & 0.8700 & {0.0488} & 0.8980 & {0.0418} & 0.9170 & {\uline{0.0336}} & 0.8548 & {\uline{0.0491}} \\
    \midrule

    \multirow{4}{*}{\textbf{MMLU-Pro}}
          & \textsc{REPCORE} & \uline{0.7320} & {\textbf{0.0910}} & \uline{0.8290} & {\uline{0.0666}} & 0.8750 & {\uline{0.0548}} & \textbf{0.9060} & {\uline{0.0476}} & \textbf{0.9240} & {\uline{0.0426}} & \textbf{0.8532} & {\uline{0.0605}} \\
          & \textsc{Random Pseudo} & 0.7103 & {0.0973} & 0.8264 & {0.0686} & \uline{0.8759} & {0.0567} & 0.9043 & {0.0491} & \uline{0.9223} & {0.0442} & 0.8478 & {0.0632} \\
          & \textsc{Advdist Pseudo} & 0.7154 & {0.0979} & \textbf{0.8301} & {0.0692} & \textbf{0.8760} & {0.0567} & \uline{0.9048} & {0.0492} & 0.9210 & {0.0444} & 0.8495 & {0.0635} \\
          & \textsc{Best Baseline} & \textbf{0.7540} & {\uline{0.0917}} & 0.8120 & {\textbf{0.0651}} & 0.8660 & {\textbf{0.0528}} & 0.8990 & {\textbf{0.0457}} & 0.9200 & {\textbf{0.0405}} & \uline{0.8502} & {\textbf{0.0592}} \\
    \midrule

    \multirow{4}{*}{\shortstack[l]{\textbf{SEED-Bench}\\\textbf{-2-Plus}}}
          & \textsc{REPCORE} & \uline{0.6420} & {\uline{0.0835}} & \uline{0.7550} & {\textbf{0.0600}} & \textbf{0.8190} & {0.0508} & \textbf{0.8530} & {\uline{0.0446}} & \textbf{0.8740} & {\uline{0.0402}} & \uline{0.7886} & {\uline{0.0558}} \\
          & \textsc{Random Pseudo} & 0.5965 & {0.0879} & 0.7339 & {0.0622} & 0.7961 & {0.0514} & 0.8370 & {0.0446} & 0.8608 & {0.0403} & 0.7649 & {0.0573} \\
          & \textsc{Advdist Pseudo} & 0.5688 & {0.0856} & 0.7325 & {\uline{0.0603}} & \uline{0.8163} & {\uline{0.0506}} & \uline{0.8491} & {0.0452} & \uline{0.8715} & {0.0408} & 0.7676 & {0.0565} \\
          & \textsc{Best Baseline} & \textbf{0.6760} & {\textbf{0.0810}} & \textbf{0.7760} & {0.0628} & 0.8130 & {\textbf{0.0500}} & 0.8360 & {\textbf{0.0428}} & 0.8630 & {\textbf{0.0378}} & \textbf{0.7928} & {\textbf{0.0549}} \\
    \midrule

    \multirow{4}{*}{\textbf{Average}}
          & \textsc{REPCORE} & \uline{0.7124} & {\uline{0.0826}} & \textbf{0.8200} & {\textbf{0.0585}} & \textbf{0.8680} & {\textbf{0.0485}} & \textbf{0.8966} & {\textbf{0.0425}} & \textbf{0.9140} & {\textbf{0.0383}} & \textbf{0.8422} & {\textbf{0.0541}} \\
          & \textsc{Random Pseudo} & 0.7052 & {0.0880} & \uline{0.8184} & {0.0629} & 0.8661 & {\uline{0.0520}} & \uline{0.8949} & {0.0454} & \uline{0.9124} & {0.0410} & \uline{0.8394} & {0.0579} \\
          & \textsc{Advdist Pseudo} & 0.6929 & {0.0903} & 0.8167 & {0.0652} & \uline{0.8673} & {0.0542} & 0.8932 & {0.0478} & 0.9111 & {0.0429} & 0.8363 & {0.0601} \\
          & \textsc{Best Baseline} & \textbf{0.7266} & {\textbf{0.0804}} & 0.8098 & {\uline{0.0620}} & 0.8530 & {0.0521} & 0.8814 & {\uline{0.0450}} & 0.9046 & {\uline{0.0386}} & 0.8351 & {\uline{0.0556}} \\

    \bottomrule
    \end{tabular}
\end{table*}

\begin{table*}[!htbp]
    \centering
    \footnotesize
    \setlength{\tabcolsep}{2.2pt}
    \renewcommand{\arraystretch}{1.1}

    \captionof{table}{\textbf{Mean performance under 30\% masking.} Spearman $\rho$ ($\uparrow$) and MAE ($\downarrow$) averaged over 10 source-model combinations $\times$ 50 random anchor selections. For Random Pseudo, scores are further averaged over 10 stochastic masking draws; \textsc{RepCore} and \textsc{Advdist Pseudo} are deterministic under each source-model combination. \textbf{Bold} indicates the best and \uline{underline} the second best per column.}
    \label{tab:masking_mean_30pct}

    \begin{tabular}{l l cc cc cc cc cc cc}
    \toprule
    \multirow{2}{*}{\textbf{Benchmark}} & \multirow{2}{*}{\textbf{Condition}} &
    \multicolumn{2}{c}{\textbf{$K=10$}} & \multicolumn{2}{c}{\textbf{$K=20$}} & \multicolumn{2}{c}{\textbf{$K=30$}} &
    \multicolumn{2}{c}{\textbf{$K=40$}} & \multicolumn{2}{c}{\textbf{$K=50$}} & \multicolumn{2}{c}{\textbf{Avg}} \\
    \cmidrule(lr){3-4}\cmidrule(lr){5-6}\cmidrule(lr){7-8}\cmidrule(lr){9-10}\cmidrule(lr){11-12}\cmidrule(lr){13-14}
    & & $\rho$ & {MAE} & $\rho$ & {MAE} & $\rho$ & {MAE} & $\rho$ & {MAE} & $\rho$ & {MAE} & $\rho$ & {MAE} \\
    \midrule

    \multirow{4}{*}{\textbf{ARC-Challenge}}
          & \textsc{REPCORE} & 0.7070 & {\uline{0.0720}} & 0.8200 & {\textbf{0.0500}} & 0.8710 & {\textbf{0.0409}} & 0.9000 & {\textbf{0.0358}} & 0.9210 & {\textbf{0.0321}} & 0.8438 & {\textbf{0.0462}} \\
          & \textsc{Random Pseudo} & \uline{0.7238} & {0.0768} & \textbf{0.8344} & {\uline{0.0545}} & \textbf{0.8841} & {\uline{0.0452}} & \textbf{0.9102} & {\uline{0.0396}} & \textbf{0.9257} & {\uline{0.0355}} & \textbf{0.8556} & {\uline{0.0503}} \\
          & \textsc{Advdist Pseudo} & 0.7149 & {0.0800} & \uline{0.8299} & {0.0594} & \uline{0.8806} & {0.0492} & \uline{0.9018} & {0.0433} & \uline{0.9211} & {0.0391} & \uline{0.8497} & {0.0542} \\
          & \textsc{Best Baseline} & \textbf{0.7520} & {\textbf{0.0717}} & 0.8230 & {0.0573} & 0.8640 & {0.0505} & 0.8900 & {0.0432} & 0.9180 & {0.0357} & 0.8494 & {0.0517} \\
    \midrule

    \multirow{4}{*}{\textbf{BBH}}
          & \textsc{REPCORE} & \uline{0.7180} & {\uline{0.0950}} & \uline{0.8240} & {\uline{0.0687}} & \textbf{0.8700} & {\textbf{0.0569}} & \textbf{0.8980} & {\textbf{0.0494}} & \textbf{0.9130} & {\textbf{0.0452}} & \textbf{0.8446} & {\textbf{0.0630}} \\
          & \textsc{Random Pseudo} & \textbf{0.7200} & {0.1051} & \textbf{0.8251} & {0.0749} & \uline{0.8666} & {0.0626} & \uline{0.8918} & {0.0547} & \uline{0.9092} & {0.0496} & \uline{0.8425} & {0.0694} \\
          & \textsc{Advdist Pseudo} & 0.6868 & {0.1076} & 0.8068 & {0.0774} & 0.8553 & {0.0638} & 0.8823 & {0.0565} & 0.9014 & {0.0504} & 0.8265 & {0.0711} \\
          & \textsc{Best Baseline} & 0.7010 & {\textbf{0.0931}} & 0.7990 & {\textbf{0.0678}} & 0.8520 & {\uline{0.0583}} & 0.8840 & {\uline{0.0516}} & 0.9050 & {\uline{0.0454}} & 0.8282 & {\uline{0.0632}} \\
    \midrule

    \multirow{4}{*}{\textbf{GSM8K}}
          & \textsc{REPCORE} & \uline{0.7630} & {\uline{0.0714}} & \textbf{0.8720} & {\textbf{0.0473}} & \textbf{0.9050} & {\textbf{0.0393}} & \textbf{0.9260} & {\textbf{0.0350}} & \textbf{0.9380} & {\textbf{0.0316}} & \textbf{0.8808} & {\textbf{0.0449}} \\
          & \textsc{Random Pseudo} & \textbf{0.7634} & {0.0747} & \uline{0.8603} & {\uline{0.0548}} & \uline{0.9003} & {\uline{0.0466}} & \uline{0.9215} & {\uline{0.0411}} & \uline{0.9356} & {0.0372} & \uline{0.8762} & {0.0509} \\
          & \textsc{Advdist Pseudo} & 0.7499 & {0.0766} & 0.8534 & {0.0608} & 0.8909 & {0.0517} & 0.9148 & {0.0447} & 0.9329 & {0.0396} & 0.8684 & {0.0547} \\
          & \textsc{Best Baseline} & 0.7500 & {\textbf{0.0644}} & 0.8390 & {0.0569} & 0.8700 & {0.0488} & 0.8980 & {0.0418} & 0.9170 & {\uline{0.0336}} & 0.8548 & {\uline{0.0491}} \\
    \midrule

    \multirow{4}{*}{\textbf{MMLU-Pro}}
          & \textsc{REPCORE} & \uline{0.7320} & {\textbf{0.0910}} & \textbf{0.8290} & {\uline{0.0666}} & \uline{0.8750} & {\uline{0.0548}} & \textbf{0.9060} & {\uline{0.0476}} & \textbf{0.9240} & {\uline{0.0426}} & \textbf{0.8532} & {\uline{0.0605}} \\
          & \textsc{Random Pseudo} & 0.7124 & {0.0975} & \uline{0.8250} & {0.0688} & \textbf{0.8760} & {0.0568} & \uline{0.9034} & {0.0495} & \uline{0.9210} & {0.0445} & 0.8476 & {0.0634} \\
          & \textsc{Advdist Pseudo} & 0.6828 & {0.0995} & 0.8130 & {0.0711} & 0.8646 & {0.0593} & 0.8944 & {0.0523} & 0.9108 & {0.0474} & 0.8331 & {0.0659} \\
          & \textsc{Best Baseline} & \textbf{0.7540} & {\uline{0.0917}} & 0.8120 & {\textbf{0.0651}} & 0.8660 & {\textbf{0.0528}} & 0.8990 & {\textbf{0.0457}} & 0.9200 & {\textbf{0.0405}} & \uline{0.8502} & {\textbf{0.0592}} \\
    \midrule

    \multirow{4}{*}{\shortstack[l]{\textbf{SEED-Bench}\\\textbf{-2-Plus}}}
          & \textsc{REPCORE} & \uline{0.6420} & {\uline{0.0835}} & \uline{0.7550} & {\textbf{0.0600}} & \textbf{0.8190} & {\uline{0.0508}} & \textbf{0.8530} & {\uline{0.0446}} & \textbf{0.8740} & {\uline{0.0402}} & \uline{0.7886} & {\uline{0.0558}} \\
          & \textsc{Random Pseudo} & 0.6035 & {0.0881} & 0.7308 & {\uline{0.0627}} & 0.7942 & {0.0519} & 0.8338 & {0.0453} & 0.8613 & {0.0411} & 0.7647 & {0.0578} \\
          & \textsc{Advdist Pseudo} & 0.5848 & {0.0893} & 0.7211 & {0.0648} & 0.7968 & {0.0539} & \uline{0.8419} & {0.0469} & \uline{0.8663} & {0.0425} & 0.7622 & {0.0595} \\
          & \textsc{Best Baseline} & \textbf{0.6760} & {\textbf{0.0810}} & \textbf{0.7760} & {0.0628} & \uline{0.8130} & {\textbf{0.0500}} & 0.8360 & {\textbf{0.0428}} & 0.8630 & {\textbf{0.0378}} & \textbf{0.7928} & {\textbf{0.0549}} \\
    \midrule

    \multirow{4}{*}{\textbf{Average}}
          & \textsc{REPCORE} & \uline{0.7124} & {\uline{0.0826}} & \textbf{0.8200} & {\textbf{0.0585}} & \textbf{0.8680} & {\textbf{0.0485}} & \textbf{0.8966} & {\textbf{0.0425}} & \textbf{0.9140} & {\textbf{0.0383}} & \textbf{0.8422} & {\textbf{0.0541}} \\
          & \textsc{Random Pseudo} & 0.7046 & {0.0884} & \uline{0.8151} & {0.0631} & \uline{0.8642} & {0.0526} & \uline{0.8921} & {0.0460} & \uline{0.9106} & {0.0416} & \uline{0.8373} & {0.0584} \\
          & \textsc{Advdist Pseudo} & 0.6838 & {0.0906} & 0.8048 & {0.0667} & 0.8576 & {0.0556} & 0.8870 & {0.0487} & 0.9065 & {0.0438} & 0.8280 & {0.0611} \\
          & \textsc{Best Baseline} & \textbf{0.7266} & {\textbf{0.0804}} & 0.8098 & {\uline{0.0620}} & 0.8530 & {\uline{0.0521}} & 0.8814 & {\uline{0.0450}} & 0.9046 & {\uline{0.0386}} & 0.8351 & {\uline{0.0556}} \\

    \bottomrule
    \end{tabular}
\end{table*}

\begin{table*}[!htbp]
    \centering
    \footnotesize
    \setlength{\tabcolsep}{2.2pt}
    \renewcommand{\arraystretch}{1.1}

    \captionof{table}{\textbf{Mean performance under 40\% masking.} Spearman $\rho$ ($\uparrow$) and MAE ($\downarrow$) are averaged over 10 source-model combinations $\times$ 50 random anchor selections. For Random Pseudo, scores are further averaged over 10 stochastic masking draws; \textsc{RepCore} and \textsc{Advdist Pseudo} are deterministic under each source-model combination. \textbf{Bold} indicates the best and \uline{underline} the second best per column.}
    \label{tab:masking_mean_40pct}

    \begin{tabular}{l l cc cc cc cc cc cc}
    \toprule
    \multirow{2}{*}{\textbf{Benchmark}} & \multirow{2}{*}{\textbf{Condition}} &
    \multicolumn{2}{c}{\textbf{$K=10$}} & \multicolumn{2}{c}{\textbf{$K=20$}} & \multicolumn{2}{c}{\textbf{$K=30$}} &
    \multicolumn{2}{c}{\textbf{$K=40$}} & \multicolumn{2}{c}{\textbf{$K=50$}} & \multicolumn{2}{c}{\textbf{Avg}} \\
    \cmidrule(lr){3-4}\cmidrule(lr){5-6}\cmidrule(lr){7-8}\cmidrule(lr){9-10}\cmidrule(lr){11-12}\cmidrule(lr){13-14}
    & & $\rho$ & {MAE} & $\rho$ & {MAE} & $\rho$ & {MAE} & $\rho$ & {MAE} & $\rho$ & {MAE} & $\rho$ & {MAE} \\
    \midrule

    \multirow{4}{*}{\textbf{ARC-Challenge}}
          & \textsc{REPCORE} & 0.7070 & {\uline{0.0720}} & 0.8200 & {\textbf{0.0500}} & \uline{0.8710} & {\textbf{0.0409}} & \uline{0.9000} & {\textbf{0.0358}} & \uline{0.9210} & {\textbf{0.0321}} & 0.8438 & {\textbf{0.0462}} \\
          & \textsc{Random Pseudo} & \uline{0.7243} & {0.0771} & \textbf{0.8345} & {\uline{0.0548}} & \textbf{0.8824} & {\uline{0.0457}} & \textbf{0.9085} & {\uline{0.0400}} & \textbf{0.9250} & {0.0360} & \textbf{0.8549} & {\uline{0.0507}} \\
          & \textsc{Advdist Pseudo} & 0.7004 & {0.0810} & 0.8078 & {0.0589} & 0.8486 & {0.0514} & 0.8717 & {0.0477} & 0.8843 & {0.0441} & 0.8226 & {0.0566} \\
          & \textsc{Best Baseline} & \textbf{0.7520} & {\textbf{0.0717}} & \uline{0.8230} & {0.0573} & 0.8640 & {0.0505} & 0.8900 & {0.0432} & 0.9180 & {\uline{0.0357}} & \uline{0.8494} & {0.0517} \\
    \midrule

    \multirow{4}{*}{\textbf{BBH}}
          & \textsc{REPCORE} & \textbf{0.7180} & {\uline{0.0950}} & \textbf{0.8240} & {\uline{0.0687}} & \textbf{0.8700} & {\textbf{0.0569}} & \textbf{0.8980} & {\textbf{0.0494}} & \textbf{0.9130} & {\textbf{0.0452}} & \textbf{0.8446} & {\textbf{0.0630}} \\
          & \textsc{Random Pseudo} & \uline{0.7094} & {0.1065} & \uline{0.8210} & {0.0752} & \uline{0.8674} & {0.0624} & \uline{0.8936} & {0.0546} & \uline{0.9102} & {0.0494} & \uline{0.8403} & {0.0696} \\
          & \textsc{Advdist Pseudo} & 0.6768 & {0.1149} & 0.7984 & {0.0775} & 0.8460 & {0.0651} & 0.8722 & {0.0571} & 0.8913 & {0.0517} & 0.8169 & {0.0733} \\
          & \textsc{Best Baseline} & 0.7010 & {\textbf{0.0931}} & 0.7990 & {\textbf{0.0678}} & 0.8520 & {\uline{0.0583}} & 0.8840 & {\uline{0.0516}} & 0.9050 & {\uline{0.0454}} & 0.8282 & {\uline{0.0632}} \\
    \midrule

    \multirow{4}{*}{\textbf{GSM8K}}
          & \textsc{REPCORE} & \textbf{0.7630} & {\uline{0.0714}} & \textbf{0.8720} & {\textbf{0.0473}} & \textbf{0.9050} & {\textbf{0.0393}} & \textbf{0.9260} & {\textbf{0.0350}} & \textbf{0.9380} & {\textbf{0.0316}} & \textbf{0.8808} & {\textbf{0.0449}} \\
          & \textsc{Random Pseudo} & 0.7593 & {0.0736} & \uline{0.8601} & {\uline{0.0546}} & \uline{0.9011} & {\uline{0.0454}} & \uline{0.9225} & {\uline{0.0402}} & \uline{0.9364} & {0.0365} & \uline{0.8759} & {0.0501} \\
          & \textsc{Advdist Pseudo} & \uline{0.7595} & {0.0785} & 0.8479 & {0.0607} & 0.8894 & {0.0497} & 0.9156 & {0.0431} & 0.9308 & {0.0387} & 0.8686 & {0.0541} \\
          & \textsc{Best Baseline} & 0.7500 & {\textbf{0.0644}} & 0.8390 & {0.0569} & 0.8700 & {0.0488} & 0.8980 & {0.0418} & 0.9170 & {\uline{0.0336}} & 0.8548 & {\uline{0.0491}} \\
    \midrule

    \multirow{4}{*}{\textbf{MMLU-Pro}}
          & \textsc{REPCORE} & \uline{0.7320} & {\textbf{0.0910}} & \textbf{0.8290} & {\uline{0.0666}} & \textbf{0.8750} & {\uline{0.0548}} & \textbf{0.9060} & {\uline{0.0476}} & \textbf{0.9240} & {\uline{0.0426}} & \textbf{0.8532} & {\uline{0.0605}} \\
          & \textsc{Random Pseudo} & 0.7034 & {0.0982} & \uline{0.8248} & {0.0691} & \uline{0.8741} & {0.0570} & \uline{0.9026} & {0.0498} & \uline{0.9204} & {0.0447} & 0.8451 & {0.0638} \\
          & \textsc{Advdist Pseudo} & 0.6719 & {0.1006} & 0.8185 & {0.0718} & 0.8632 & {0.0595} & 0.8939 & {0.0527} & 0.9126 & {0.0483} & 0.8320 & {0.0666} \\
          & \textsc{Best Baseline} & \textbf{0.7540} & {\uline{0.0917}} & 0.8120 & {\textbf{0.0651}} & 0.8660 & {\textbf{0.0528}} & 0.8990 & {\textbf{0.0457}} & 0.9200 & {\textbf{0.0405}} & \uline{0.8502} & {\textbf{0.0592}} \\
    \midrule

    \multirow{4}{*}{\shortstack[l]{\textbf{SEED-Bench}\\\textbf{-2-Plus}}}
          & \textsc{REPCORE} & \uline{0.6420} & {\uline{0.0835}} & \uline{0.7550} & {\textbf{0.0600}} & \textbf{0.8190} & {\uline{0.0508}} & \textbf{0.8530} & {\uline{0.0446}} & \textbf{0.8740} & {\uline{0.0402}} & \uline{0.7886} & {\uline{0.0558}} \\
          & \textsc{Random Pseudo} & 0.6025 & {0.0875} & 0.7328 & {0.0637} & 0.7972 & {0.0520} & \uline{0.8361} & {0.0457} & 0.8629 & {0.0412} & 0.7663 & {0.0580} \\
          & \textsc{Advdist Pseudo} & 0.6061 & {0.0889} & 0.7149 & {0.0648} & 0.7901 & {0.0544} & 0.8325 & {0.0496} & \uline{0.8666} & {0.0454} & 0.7620 & {0.0606} \\
          & \textsc{Best Baseline} & \textbf{0.6760} & {\textbf{0.0810}} & \textbf{0.7760} & {\uline{0.0628}} & \uline{0.8130} & {\textbf{0.0500}} & 0.8360 & {\textbf{0.0428}} & 0.8630 & {\textbf{0.0378}} & \textbf{0.7928} & {\textbf{0.0549}} \\
    \midrule

    \multirow{4}{*}{\textbf{Average}}
          & \textsc{REPCORE} & \uline{0.7124} & {\uline{0.0826}} & \textbf{0.8200} & {\textbf{0.0585}} & \textbf{0.8680} & {\textbf{0.0485}} & \textbf{0.8966} & {\textbf{0.0425}} & \textbf{0.9140} & {\textbf{0.0383}} & \textbf{0.8422} & {\textbf{0.0541}} \\
          & \textsc{Random Pseudo} & 0.6998 & {0.0886} & \uline{0.8146} & {0.0635} & \uline{0.8644} & {0.0525} & \uline{0.8927} & {0.0461} & \uline{0.9110} & {0.0416} & \uline{0.8365} & {0.0584} \\
          & \textsc{Advdist Pseudo} & 0.6829 & {0.0928} & 0.7975 & {0.0667} & 0.8475 & {0.0560} & 0.8772 & {0.0500} & 0.8971 & {0.0456} & 0.8204 & {0.0622} \\
          & \textsc{Best Baseline} & \textbf{0.7266} & {\textbf{0.0804}} & 0.8098 & {\uline{0.0620}} & 0.8530 & {\uline{0.0521}} & 0.8814 & {\uline{0.0450}} & 0.9046 & {\uline{0.0386}} & 0.8351 & {\uline{0.0556}} \\

    \bottomrule
    \end{tabular}
\end{table*}

\begin{table*}[!htbp]
    \centering
    \footnotesize
    \setlength{\tabcolsep}{2.2pt}
    \renewcommand{\arraystretch}{1.1}

    \captionof{table}{\textbf{Mean performance under 50\% masking.} Spearman $\rho$ ($\uparrow$) and MAE ($\downarrow$) are averaged over 10 source-model combinations $\times$ 50 random anchor selections. For Random Pseudo, scores are further averaged over 10 stochastic masking draws; \textsc{RepCore} and \textsc{Advdist Pseudo} are deterministic under each source-model combination. \textbf{Bold} indicates the best and \uline{underline} the second best per column.}
    \label{tab:masking_mean_50pct}

    \begin{tabular}{l l cc cc cc cc cc cc}
    \toprule
    \multirow{2}{*}{\textbf{Benchmark}} & \multirow{2}{*}{\textbf{Condition}} &
    \multicolumn{2}{c}{\textbf{$K=10$}} & \multicolumn{2}{c}{\textbf{$K=20$}} & \multicolumn{2}{c}{\textbf{$K=30$}} &
    \multicolumn{2}{c}{\textbf{$K=40$}} & \multicolumn{2}{c}{\textbf{$K=50$}} & \multicolumn{2}{c}{\textbf{Avg}} \\
    \cmidrule(lr){3-4}\cmidrule(lr){5-6}\cmidrule(lr){7-8}\cmidrule(lr){9-10}\cmidrule(lr){11-12}\cmidrule(lr){13-14}
    & & $\rho$ & {MAE} & $\rho$ & {MAE} & $\rho$ & {MAE} & $\rho$ & {MAE} & $\rho$ & {MAE} & $\rho$ & {MAE} \\
    \midrule

    \multirow{4}{*}{\textbf{ARC-Challenge}}
          & \textsc{REPCORE} & 0.7070 & {\uline{0.0720}} & 0.8200 & {\textbf{0.0500}} & \uline{0.8710} & {\textbf{0.0409}} & \uline{0.9000} & {\textbf{0.0358}} & \textbf{0.9210} & {\textbf{0.0321}} & 0.8438 & {\textbf{0.0462}} \\
          & \textsc{Random Pseudo} & \uline{0.7133} & {0.0765} & \textbf{0.8277} & {\uline{0.0550}} & \textbf{0.8766} & {\uline{0.0454}} & \textbf{0.9041} & {\uline{0.0396}} & \uline{0.9207} & {\uline{0.0355}} & \uline{0.8485} & {\uline{0.0504}} \\
          & \textsc{Advdist Pseudo} & 0.6929 & {0.0806} & 0.7934 & {0.0615} & 0.8515 & {0.0516} & 0.8880 & {0.0454} & 0.9091 & {0.0399} & 0.8270 & {0.0558} \\
          & \textsc{Best Baseline} & \textbf{0.7520} & {\textbf{0.0717}} & \uline{0.8230} & {0.0573} & 0.8640 & {0.0505} & 0.8900 & {0.0432} & 0.9180 & {0.0357} & \textbf{0.8494} & {0.0517} \\
    \midrule

    \multirow{4}{*}{\textbf{BBH}}
          & \textsc{REPCORE} & \textbf{0.7180} & {\uline{0.0950}} & \textbf{0.8240} & {\uline{0.0687}} & \textbf{0.8700} & {\textbf{0.0569}} & \textbf{0.8980} & {\textbf{0.0494}} & \textbf{0.9130} & {\textbf{0.0452}} & \textbf{0.8446} & {\textbf{0.0630}} \\
          & \textsc{Random Pseudo} & 0.6999 & {0.1073} & \uline{0.8105} & {0.0762} & \uline{0.8578} & {0.0625} & 0.8834 & {0.0548} & 0.9023 & {0.0497} & \uline{0.8308} & {0.0701} \\
          & \textsc{Advdist Pseudo} & 0.6528 & {0.1121} & 0.7507 & {0.0845} & 0.8067 & {0.0713} & 0.8423 & {0.0628} & 0.8616 & {0.0576} & 0.7828 & {0.0777} \\
          & \textsc{Best Baseline} & \uline{0.7010} & {\textbf{0.0931}} & 0.7990 & {\textbf{0.0678}} & 0.8520 & {\uline{0.0583}} & \uline{0.8840} & {\uline{0.0516}} & \uline{0.9050} & {\uline{0.0454}} & 0.8282 & {\uline{0.0632}} \\
    \midrule

    \multirow{4}{*}{\textbf{GSM8K}}
          & \textsc{REPCORE} & \textbf{0.7630} & {\uline{0.0714}} & \textbf{0.8720} & {\textbf{0.0473}} & \textbf{0.9050} & {\textbf{0.0393}} & \textbf{0.9260} & {\textbf{0.0350}} & \textbf{0.9380} & {\textbf{0.0316}} & \textbf{0.8808} & {\textbf{0.0449}} \\
          & \textsc{Random Pseudo} & \uline{0.7629} & {0.0744} & \uline{0.8604} & {\uline{0.0550}} & \uline{0.9006} & {\uline{0.0459}} & \uline{0.9215} & {\uline{0.0403}} & \uline{0.9358} & {0.0367} & \uline{0.8762} & {0.0505} \\
          & \textsc{Advdist Pseudo} & 0.7420 & {0.0755} & 0.8441 & {0.0604} & 0.8903 & {0.0499} & 0.9168 & {0.0430} & 0.9301 & {0.0383} & 0.8647 & {0.0534} \\
          & \textsc{Best Baseline} & 0.7500 & {\textbf{0.0644}} & 0.8390 & {0.0569} & 0.8700 & {0.0488} & 0.8980 & {0.0418} & 0.9170 & {\uline{0.0336}} & 0.8548 & {\uline{0.0491}} \\
    \midrule

    \multirow{4}{*}{\textbf{MMLU-Pro}}
          & \textsc{REPCORE} & \uline{0.7320} & {\textbf{0.0910}} & \textbf{0.8290} & {\uline{0.0666}} & \textbf{0.8750} & {\uline{0.0548}} & \textbf{0.9060} & {\uline{0.0476}} & \textbf{0.9240} & {\uline{0.0426}} & \textbf{0.8532} & {\uline{0.0605}} \\
          & \textsc{Random Pseudo} & 0.7059 & {0.0988} & \uline{0.8162} & {0.0700} & \uline{0.8706} & {0.0576} & \uline{0.8990} & {0.0504} & 0.9170 & {0.0456} & 0.8417 & {0.0645} \\
          & \textsc{Advdist Pseudo} & 0.6485 & {0.1050} & 0.7843 & {0.0800} & 0.8378 & {0.0662} & 0.8759 & {0.0587} & 0.8980 & {0.0527} & 0.8089 & {0.0725} \\
          & \textsc{Best Baseline} & \textbf{0.7540} & {\uline{0.0917}} & 0.8120 & {\textbf{0.0651}} & 0.8660 & {\textbf{0.0528}} & 0.8990 & {\textbf{0.0457}} & \uline{0.9200} & {\textbf{0.0405}} & \uline{0.8502} & {\textbf{0.0592}} \\
    \midrule

    \multirow{4}{*}{\shortstack[l]{\textbf{SEED-Bench}\\\textbf{-2-Plus}}}
          & \textsc{REPCORE} & \uline{0.6420} & {\uline{0.0835}} & \uline{0.7550} & {\textbf{0.0600}} & \textbf{0.8190} & {\uline{0.0508}} & \textbf{0.8530} & {\uline{0.0446}} & \textbf{0.8740} & {\uline{0.0402}} & \uline{0.7886} & {\uline{0.0558}} \\
          & \textsc{Random Pseudo} & 0.6155 & {0.0876} & 0.7343 & {0.0636} & 0.7932 & {0.0531} & 0.8320 & {0.0465} & 0.8579 & {0.0420} & 0.7666 & {0.0586} \\
          & \textsc{Advdist Pseudo} & 0.5642 & {0.0912} & 0.7153 & {0.0718} & 0.7776 & {0.0608} & 0.8144 & {0.0550} & 0.8361 & {0.0506} & 0.7415 & {0.0659} \\
          & \textsc{Best Baseline} & \textbf{0.6760} & {\textbf{0.0810}} & \textbf{0.7760} & {\uline{0.0628}} & \uline{0.8130} & {\textbf{0.0500}} & \uline{0.8360} & {\textbf{0.0428}} & \uline{0.8630} & {\textbf{0.0378}} & \textbf{0.7928} & {\textbf{0.0549}} \\
    \midrule

    \multirow{4}{*}{\textbf{Average}}
          & \textsc{REPCORE} & \uline{0.7124} & {\uline{0.0826}} & \textbf{0.8200} & {\textbf{0.0585}} & \textbf{0.8680} & {\textbf{0.0485}} & \textbf{0.8966} & {\textbf{0.0425}} & \textbf{0.9140} & {\textbf{0.0383}} & \textbf{0.8422} & {\textbf{0.0541}} \\
          & \textsc{Random Pseudo} & 0.6995 & {0.0889} & \uline{0.8098} & {0.0640} & \uline{0.8598} & {0.0529} & \uline{0.8880} & {0.0463} & \uline{0.9067} & {0.0419} & 0.8328 & {0.0588} \\
          & \textsc{Advdist Pseudo} & 0.6601 & {0.0929} & 0.7776 & {0.0716} & 0.8328 & {0.0600} & 0.8675 & {0.0530} & 0.8870 & {0.0478} & 0.8050 & {0.0651} \\
          & \textsc{Best Baseline} & \textbf{0.7266} & {\textbf{0.0804}} & 0.8098 & {\uline{0.0620}} & 0.8530 & {\uline{0.0521}} & 0.8814 & {\uline{0.0450}} & 0.9046 & {\uline{0.0386}} & \uline{0.8351} & {\uline{0.0556}} \\

    \bottomrule
    \end{tabular}
\end{table*}

\begin{table*}[!htbp]
    \centering
    \footnotesize
    \setlength{\tabcolsep}{2.2pt}
    \renewcommand{\arraystretch}{1.1}

    \captionof{table}{\textbf{$Combo_{std}$ under 10\% masking:} standard deviation across 10 source-model combinations. For \colorbox{gray!20}{Random Pseudo}, each combo score is first averaged over 10 stochastic masking draws (i.e., an additional $Draw_{std}$ averaging layer is applied before computing $Combo_{std}$), which structurally reduces the resulting variance, so its values are shown as reference only. \textsc{RepCore} (0\% masking) and \textsc{Advdist Pseudo} are deterministic under each source-model combination, so their comparison is fair. \textbf{Bold} marks the lower (better) value per column.}
    \label{tab:masking_combo_std_10pct}

    \begin{tabular}{l l cc cc cc cc cc cc}
    \toprule
    \multirow{2}{*}{\textbf{Benchmark}} & \multirow{2}{*}{\textbf{Condition}} &
    \multicolumn{2}{c}{\textbf{$K=10$}} & \multicolumn{2}{c}{\textbf{$K=20$}} & \multicolumn{2}{c}{\textbf{$K=30$}} &
    \multicolumn{2}{c}{\textbf{$K=40$}} & \multicolumn{2}{c}{\textbf{$K=50$}} & \multicolumn{2}{c}{\textbf{Avg}} \\
    \cmidrule(lr){3-4}\cmidrule(lr){5-6}\cmidrule(lr){7-8}\cmidrule(lr){9-10}\cmidrule(lr){11-12}\cmidrule(lr){13-14}
    & & $\rho$ & {MAE} & $\rho$ & {MAE} & $\rho$ & {MAE} & $\rho$ & {MAE} & $\rho$ & {MAE} & $\rho$ & {MAE} \\
    \midrule

    \rowcolor{gray!20} \cellcolor{white}\multirow{3}{*}{\textbf{ARC-Challenge}}
          & \textsc{Random Pseudo} & 0.0193 & {0.0043} & 0.0105 & {0.0038} & 0.0080 & {0.0032} & 0.0057 & {0.0028} & 0.0057 & {0.0024} & 0.0098 & {0.0033} \\
          & \textsc{REPCORE} & 0.0461 & {\textbf{0.0049}} & 0.0229 & {\textbf{0.0023}} & 0.0197 & {\textbf{0.0017}} & 0.0145 & {\textbf{0.0015}} & \textbf{0.0094} & {\textbf{0.0014}} & 0.0225 & {\textbf{0.0024}} \\
          & \textsc{Advdist Pseudo} & \textbf{0.0233} & {0.0089} & \textbf{0.0189} & {0.0059} & \textbf{0.0167} & {0.0038} & \textbf{0.0133} & {0.0047} & 0.0138 & {0.0047} & \textbf{0.0172} & {0.0056} \\
    \midrule

    \rowcolor{gray!20} \cellcolor{white}\multirow{3}{*}{\textbf{BBH}}
          & \textsc{Random Pseudo} & 0.0148 & {0.0043} & 0.0089 & {0.0022} & 0.0063 & {0.0018} & 0.0051 & {0.0015} & 0.0048 & {0.0014} & 0.0080 & {0.0022} \\
          & \textsc{REPCORE} & \textbf{0.0357} & {\textbf{0.0074}} & 0.0290 & {0.0048} & \textbf{0.0151} & {\textbf{0.0020}} & \textbf{0.0087} & {\textbf{0.0014}} & \textbf{0.0057} & {\textbf{0.0021}} & \textbf{0.0188} & {\textbf{0.0035}} \\
          & \textsc{Advdist Pseudo} & 0.0513 & {0.0088} & \textbf{0.0276} & {\textbf{0.0034}} & 0.0223 & {0.0043} & 0.0143 & {0.0030} & 0.0110 & {0.0023} & 0.0253 & {0.0044} \\
    \midrule

    \rowcolor{gray!20} \cellcolor{white}\multirow{3}{*}{\textbf{GSM8K}}
          & \textsc{Random Pseudo} & 0.0153 & {0.0061} & 0.0080 & {0.0061} & 0.0055 & {0.0043} & 0.0034 & {0.0034} & 0.0022 & {0.0028} & 0.0069 & {0.0045} \\
          & \textsc{REPCORE} & 0.0377 & {\textbf{0.0051}} & \textbf{0.0106} & {\textbf{0.0017}} & \textbf{0.0095} & {\textbf{0.0020}} & \textbf{0.0065} & {\textbf{0.0022}} & \textbf{0.0060} & {\textbf{0.0025}} & \textbf{0.0141} & {\textbf{0.0027}} \\
          & \textsc{Advdist Pseudo} & \textbf{0.0230} & {0.0122} & 0.0202 & {0.0081} & 0.0181 & {0.0047} & 0.0116 & {0.0046} & 0.0085 & {0.0047} & 0.0163 & {0.0069} \\
    \midrule

    \rowcolor{gray!20} \cellcolor{white}\multirow{3}{*}{\textbf{MMLU-Pro}}
          & \textsc{Random Pseudo} & 0.0197 & {0.0033} & 0.0090 & {0.0010} & 0.0044 & {0.0010} & 0.0045 & {0.0007} & 0.0036 & {0.0007} & 0.0082 & {0.0013} \\
          & \textsc{REPCORE} & \textbf{0.0341} & {\textbf{0.0052}} & \textbf{0.0183} & {0.0031} & \textbf{0.0121} & {\textbf{0.0021}} & \textbf{0.0061} & {\textbf{0.0010}} & 0.0072 & {0.0017} & \textbf{0.0156} & {0.0026} \\
          & \textsc{Advdist Pseudo} & 0.0456 & {0.0057} & 0.0219 & {\textbf{0.0017}} & 0.0123 & {0.0024} & 0.0101 & {0.0018} & \textbf{0.0067} & {\textbf{0.0015}} & 0.0193 & {0.0026} \\
    \midrule

    \rowcolor{gray!20} \cellcolor{white}\multirow{3}{*}{\shortstack[l]{\textbf{SEED-Bench}\\\textbf{-2-Plus}}}
          & \textsc{Random Pseudo} & 0.0320 & {0.0023} & 0.0205 & {0.0017} & 0.0159 & {0.0018} & 0.0164 & {0.0019} & 0.0125 & {0.0017} & 0.0195 & {0.0019} \\
          & \textsc{REPCORE} & \textbf{0.0387} & {0.0106} & 0.0503 & {\textbf{0.0037}} & 0.0356 & {\textbf{0.0018}} & 0.0212 & {\textbf{0.0022}} & 0.0192 & {\textbf{0.0018}} & 0.0330 & {\textbf{0.0040}} \\
          & \textsc{Advdist Pseudo} & 0.0424 & {\textbf{0.0089}} & \textbf{0.0324} & {0.0041} & \textbf{0.0183} & {0.0030} & \textbf{0.0172} & {0.0028} & \textbf{0.0154} & {0.0029} & \textbf{0.0251} & {0.0043} \\
    \midrule

    \rowcolor{gray!20} \cellcolor{white}\multirow{3}{*}{\textbf{Average}}
          & \textsc{Random Pseudo} & 0.0202 & {0.0041} & 0.0114 & {0.0030} & 0.0080 & {0.0024} & 0.0070 & {0.0021} & 0.0058 & {0.0018} & 0.0105 & {0.0027} \\
          & \textsc{REPCORE} & 0.0385 & {\textbf{0.0066}} & 0.0262 & {\textbf{0.0031}} & 0.0184 & {\textbf{0.0019}} & \textbf{0.0114} & {\textbf{0.0017}} & \textbf{0.0095} & {\textbf{0.0019}} & 0.0208 & {\textbf{0.0030}} \\
          & \textsc{Advdist Pseudo} & \textbf{0.0371} & {0.0089} & \textbf{0.0242} & {0.0046} & \textbf{0.0175} & {0.0036} & 0.0133 & {0.0034} & 0.0111 & {0.0032} & \textbf{0.0206} & {0.0048} \\

    \bottomrule
    \end{tabular}
\end{table*}

\begin{table*}[!htbp]
    \centering
    \footnotesize
    \setlength{\tabcolsep}{2.2pt}
    \renewcommand{\arraystretch}{1.1}

    \captionof{table}{\textbf{$Combo_{std}$ under 20\% masking:} standard deviation across 10 source-model combinations. For \colorbox{gray!20}{Random Pseudo}, each combo score is first averaged over 10 stochastic masking draws (i.e., an additional $Draw_{std}$ averaging layer is applied before computing $Combo_{std}$), which structurally reduces the resulting variance, so its values are shown as reference only. \textsc{RepCore} (0\% masking) and \textsc{Advdist Pseudo} are deterministic under each source-model combination, so their comparison is fair. \textbf{Bold} marks the lower (better) value per column.}
    \label{tab:masking_combo_std_20pct}

    \begin{tabular}{l l cc cc cc cc cc cc}
    \toprule
    \multirow{2}{*}{\textbf{Benchmark}} & \multirow{2}{*}{\textbf{Condition}} &
    \multicolumn{2}{c}{\textbf{$K=10$}} & \multicolumn{2}{c}{\textbf{$K=20$}} & \multicolumn{2}{c}{\textbf{$K=30$}} &
    \multicolumn{2}{c}{\textbf{$K=40$}} & \multicolumn{2}{c}{\textbf{$K=50$}} & \multicolumn{2}{c}{\textbf{Avg}} \\
    \cmidrule(lr){3-4}\cmidrule(lr){5-6}\cmidrule(lr){7-8}\cmidrule(lr){9-10}\cmidrule(lr){11-12}\cmidrule(lr){13-14}
    & & $\rho$ & {MAE} & $\rho$ & {MAE} & $\rho$ & {MAE} & $\rho$ & {MAE} & $\rho$ & {MAE} & $\rho$ & {MAE} \\
    \midrule

    \rowcolor{gray!20} \cellcolor{white}\multirow{3}{*}{\textbf{ARC-Challenge}}
          & \textsc{Random Pseudo} & 0.0193 & {0.0032} & 0.0085 & {0.0028} & 0.0051 & {0.0022} & 0.0040 & {0.0022} & 0.0027 & {0.0020} & 0.0079 & {0.0025} \\
          & \textsc{REPCORE} & 0.0461 & {\textbf{0.0049}} & 0.0229 & {\textbf{0.0023}} & 0.0197 & {\textbf{0.0017}} & 0.0145 & {\textbf{0.0015}} & 0.0094 & {\textbf{0.0014}} & 0.0225 & {\textbf{0.0024}} \\
          & \textsc{Advdist Pseudo} & \textbf{0.0388} & {0.0064} & \textbf{0.0173} & {0.0091} & \textbf{0.0112} & {0.0057} & \textbf{0.0066} & {0.0041} & \textbf{0.0064} & {0.0033} & \textbf{0.0161} & {0.0057} \\
    \midrule

    \rowcolor{gray!20} \cellcolor{white}\multirow{3}{*}{\textbf{BBH}}
          & \textsc{Random Pseudo} & 0.0170 & {0.0032} & 0.0062 & {0.0019} & 0.0080 & {0.0012} & 0.0070 & {0.0015} & 0.0054 & {0.0017} & 0.0087 & {0.0019} \\
          & \textsc{REPCORE} & \textbf{0.0357} & {0.0074} & 0.0290 & {\textbf{0.0048}} & \textbf{0.0151} & {\textbf{0.0020}} & \textbf{0.0087} & {\textbf{0.0014}} & \textbf{0.0057} & {\textbf{0.0021}} & \textbf{0.0188} & {\textbf{0.0035}} \\
          & \textsc{Advdist Pseudo} & 0.0563 & {\textbf{0.0055}} & \textbf{0.0280} & {0.0070} & 0.0235 & {0.0047} & 0.0152 & {0.0032} & 0.0080 & {0.0030} & 0.0262 & {0.0047} \\
    \midrule

    \rowcolor{gray!20} \cellcolor{white}\multirow{3}{*}{\textbf{GSM8K}}
          & \textsc{Random Pseudo} & 0.0167 & {0.0030} & 0.0068 & {0.0021} & 0.0033 & {0.0023} & 0.0025 & {0.0023} & 0.0029 & {0.0021} & 0.0064 & {0.0024} \\
          & \textsc{REPCORE} & 0.0377 & {\textbf{0.0051}} & \textbf{0.0106} & {\textbf{0.0017}} & \textbf{0.0095} & {\textbf{0.0020}} & \textbf{0.0065} & {\textbf{0.0022}} & 0.0060 & {\textbf{0.0025}} & \textbf{0.0141} & {\textbf{0.0027}} \\
          & \textsc{Advdist Pseudo} & \textbf{0.0302} & {0.0089} & 0.0193 & {0.0074} & 0.0107 & {0.0054} & 0.0084 & {0.0046} & \textbf{0.0046} & {0.0053} & 0.0146 & {0.0063} \\
    \midrule

    \rowcolor{gray!20} \cellcolor{white}\multirow{3}{*}{\textbf{MMLU-Pro}}
          & \textsc{Random Pseudo} & 0.0199 & {0.0023} & 0.0059 & {0.0009} & 0.0044 & {0.0006} & 0.0022 & {0.0005} & 0.0029 & {0.0005} & 0.0071 & {0.0010} \\
          & \textsc{REPCORE} & 0.0341 & {0.0052} & \textbf{0.0183} & {\textbf{0.0031}} & 0.0121 & {0.0021} & \textbf{0.0061} & {\textbf{0.0010}} & \textbf{0.0072} & {\textbf{0.0017}} & \textbf{0.0156} & {\textbf{0.0026}} \\
          & \textsc{Advdist Pseudo} & \textbf{0.0340} & {\textbf{0.0049}} & 0.0235 & {0.0033} & \textbf{0.0103} & {\textbf{0.0020}} & 0.0092 & {0.0018} & 0.0079 & {0.0018} & 0.0170 & {0.0028} \\
    \midrule

    \rowcolor{gray!20} \cellcolor{white}\multirow{3}{*}{\shortstack[l]{\textbf{SEED-Bench}\\\textbf{-2-Plus}}}
          & \textsc{Random Pseudo} & 0.0305 & {0.0025} & 0.0171 & {0.0008} & 0.0146 & {0.0013} & 0.0089 & {0.0010} & 0.0076 & {0.0009} & 0.0157 & {0.0013} \\
          & \textsc{REPCORE} & \textbf{0.0387} & {0.0106} & 0.0503 & {0.0037} & 0.0356 & {\textbf{0.0018}} & \textbf{0.0212} & {\textbf{0.0022}} & 0.0192 & {\textbf{0.0018}} & 0.0330 & {0.0040} \\
          & \textsc{Advdist Pseudo} & 0.0643 & {\textbf{0.0074}} & \textbf{0.0323} & {\textbf{0.0032}} & \textbf{0.0232} & {0.0026} & 0.0245 & {0.0033} & \textbf{0.0181} & {0.0029} & \textbf{0.0325} & {\textbf{0.0039}} \\
    \midrule

    \rowcolor{gray!20} \cellcolor{white}\multirow{3}{*}{\textbf{Average}}
          & \textsc{Random Pseudo} & 0.0207 & {0.0028} & 0.0089 & {0.0017} & 0.0071 & {0.0015} & 0.0049 & {0.0015} & 0.0043 & {0.0014} & 0.0092 & {0.0018} \\
          & \textsc{REPCORE} & \textbf{0.0385} & {0.0066} & 0.0262 & {\textbf{0.0031}} & 0.0184 & {\textbf{0.0019}} & \textbf{0.0114} & {\textbf{0.0017}} & 0.0095 & {\textbf{0.0019}} & \textbf{0.0208} & {\textbf{0.0030}} \\
          & \textsc{Advdist Pseudo} & 0.0447 & {\textbf{0.0066}} & \textbf{0.0241} & {0.0060} & \textbf{0.0158} & {0.0041} & 0.0128 & {0.0034} & \textbf{0.0090} & {0.0033} & 0.0213 & {0.0047} \\

    \bottomrule
    \end{tabular}
\end{table*}

\begin{table*}[!htbp]
    \centering
    \footnotesize
    \setlength{\tabcolsep}{2.2pt}
    \renewcommand{\arraystretch}{1.1}

    \captionof{table}{\textbf{$Combo_{std}$ under 30\% masking:} standard deviation across 10 source-model combinations. For \colorbox{gray!20}{Random Pseudo}, each combo score is first averaged over 10 stochastic masking draws (i.e., an additional $Draw_{std}$ averaging layer is applied before computing $Combo_{std}$), which structurally reduces the resulting variance, so its values are shown as reference only. \textsc{RepCore} (0\% masking) and \textsc{Advdist Pseudo} are deterministic under each source-model combination, so their comparison is fair. \textbf{Bold} marks the lower (better) value per column.}
    \label{tab:masking_combo_std_30pct}

    \begin{tabular}{l l cc cc cc cc cc cc}
    \toprule
    \multirow{2}{*}{\textbf{Benchmark}} & \multirow{2}{*}{\textbf{Condition}} &
    \multicolumn{2}{c}{\textbf{$K=10$}} & \multicolumn{2}{c}{\textbf{$K=20$}} & \multicolumn{2}{c}{\textbf{$K=30$}} &
    \multicolumn{2}{c}{\textbf{$K=40$}} & \multicolumn{2}{c}{\textbf{$K=50$}} & \multicolumn{2}{c}{\textbf{Avg}} \\
    \cmidrule(lr){3-4}\cmidrule(lr){5-6}\cmidrule(lr){7-8}\cmidrule(lr){9-10}\cmidrule(lr){11-12}\cmidrule(lr){13-14}
    & & $\rho$ & {MAE} & $\rho$ & {MAE} & $\rho$ & {MAE} & $\rho$ & {MAE} & $\rho$ & {MAE} & $\rho$ & {MAE} \\
    \midrule

    \rowcolor{gray!20} \cellcolor{white}\multirow{3}{*}{\textbf{ARC-Challenge}}
          & \textsc{Random Pseudo} & 0.0203 & {0.0017} & 0.0100 & {0.0022} & 0.0054 & {0.0023} & 0.0043 & {0.0025} & 0.0032 & {0.0022} & 0.0086 & {0.0022} \\
          & \textsc{REPCORE} & 0.0461 & {0.0049} & \textbf{0.0229} & {\textbf{0.0023}} & 0.0197 & {\textbf{0.0017}} & 0.0145 & {\textbf{0.0015}} & \textbf{0.0094} & {\textbf{0.0014}} & 0.0225 & {\textbf{0.0024}} \\
          & \textsc{Advdist Pseudo} & \textbf{0.0341} & {\textbf{0.0046}} & 0.0243 & {0.0052} & \textbf{0.0158} & {0.0056} & \textbf{0.0138} & {0.0059} & 0.0106 & {0.0051} & \textbf{0.0197} & {0.0053} \\
    \midrule

    \rowcolor{gray!20} \cellcolor{white}\multirow{3}{*}{\textbf{BBH}}
          & \textsc{Random Pseudo} & 0.0105 & {0.0028} & 0.0134 & {0.0016} & 0.0111 & {0.0012} & 0.0099 & {0.0016} & 0.0078 & {0.0017} & 0.0105 & {0.0018} \\
          & \textsc{REPCORE} & \textbf{0.0357} & {\textbf{0.0074}} & \textbf{0.0290} & {\textbf{0.0048}} & \textbf{0.0151} & {\textbf{0.0020}} & \textbf{0.0087} & {\textbf{0.0014}} & \textbf{0.0057} & {\textbf{0.0021}} & \textbf{0.0188} & {\textbf{0.0035}} \\
          & \textsc{Advdist Pseudo} & 0.0733 & {0.0109} & 0.0468 & {0.0050} & 0.0344 & {0.0035} & 0.0227 & {0.0036} & 0.0194 & {0.0033} & 0.0393 & {0.0053} \\
    \midrule

    \rowcolor{gray!20} \cellcolor{white}\multirow{3}{*}{\textbf{GSM8K}}
          & \textsc{Random Pseudo} & 0.0127 & {0.0026} & 0.0078 & {0.0020} & 0.0034 & {0.0019} & 0.0022 & {0.0021} & 0.0029 & {0.0018} & 0.0058 & {0.0021} \\
          & \textsc{REPCORE} & 0.0377 & {\textbf{0.0051}} & \textbf{0.0106} & {\textbf{0.0017}} & \textbf{0.0095} & {\textbf{0.0020}} & \textbf{0.0065} & {\textbf{0.0022}} & \textbf{0.0060} & {\textbf{0.0025}} & \textbf{0.0141} & {\textbf{0.0027}} \\
          & \textsc{Advdist Pseudo} & \textbf{0.0336} & {0.0082} & 0.0173 & {0.0087} & 0.0144 & {0.0073} & 0.0116 & {0.0066} & 0.0078 & {0.0059} & 0.0169 & {0.0073} \\
    \midrule

    \rowcolor{gray!20} \cellcolor{white}\multirow{3}{*}{\textbf{MMLU-Pro}}
          & \textsc{Random Pseudo} & 0.0125 & {0.0032} & 0.0090 & {0.0009} & 0.0049 & {0.0008} & 0.0033 & {0.0009} & 0.0027 & {0.0006} & 0.0065 & {0.0013} \\
          & \textsc{REPCORE} & \textbf{0.0341} & {0.0052} & \textbf{0.0183} & {\textbf{0.0031}} & \textbf{0.0121} & {\textbf{0.0021}} & \textbf{0.0061} & {\textbf{0.0010}} & \textbf{0.0072} & {\textbf{0.0017}} & \textbf{0.0156} & {\textbf{0.0026}} \\
          & \textsc{Advdist Pseudo} & 0.0573 & {\textbf{0.0045}} & 0.0252 & {0.0056} & 0.0162 & {0.0023} & 0.0105 & {0.0023} & 0.0107 & {0.0021} & 0.0240 & {0.0034} \\
    \midrule

    \rowcolor{gray!20} \cellcolor{white}\multirow{3}{*}{\shortstack[l]{\textbf{SEED-Bench}\\\textbf{-2-Plus}}}
          & \textsc{Random Pseudo} & 0.0203 & {0.0026} & 0.0185 & {0.0011} & 0.0161 & {0.0012} & 0.0101 & {0.0010} & 0.0076 & {0.0008} & 0.0145 & {0.0013} \\
          & \textsc{REPCORE} & \textbf{0.0387} & {0.0106} & 0.0503 & {\textbf{0.0037}} & 0.0356 & {\textbf{0.0018}} & \textbf{0.0212} & {\textbf{0.0022}} & \textbf{0.0192} & {\textbf{0.0018}} & \textbf{0.0330} & {\textbf{0.0040}} \\
          & \textsc{Advdist Pseudo} & 0.0825 & {\textbf{0.0042}} & \textbf{0.0388} & {0.0049} & \textbf{0.0268} & {0.0056} & 0.0287 & {0.0044} & 0.0219 & {0.0028} & 0.0397 & {0.0044} \\
    \midrule

    \rowcolor{gray!20} \cellcolor{white}\multirow{3}{*}{\textbf{Average}}
          & \textsc{Random Pseudo} & 0.0153 & {0.0026} & 0.0117 & {0.0016} & 0.0082 & {0.0015} & 0.0060 & {0.0016} & 0.0048 & {0.0014} & 0.0092 & {0.0017} \\
          & \textsc{REPCORE} & \textbf{0.0385} & {0.0066} & \textbf{0.0262} & {\textbf{0.0031}} & \textbf{0.0184} & {\textbf{0.0019}} & \textbf{0.0114} & {\textbf{0.0017}} & \textbf{0.0095} & {\textbf{0.0019}} & \textbf{0.0208} & {\textbf{0.0030}} \\
          & \textsc{Advdist Pseudo} & 0.0562 & {\textbf{0.0065}} & 0.0305 & {0.0059} & 0.0215 & {0.0049} & 0.0175 & {0.0046} & 0.0141 & {0.0038} & 0.0279 & {0.0051} \\

    \bottomrule
    \end{tabular}
\end{table*}

\begin{table*}[!htbp]
    \centering
    \footnotesize
    \setlength{\tabcolsep}{2.2pt}
    \renewcommand{\arraystretch}{1.1}

    \captionof{table}{\textbf{$Combo_{std}$ under 40\% masking:} standard deviation across 10 source-model combinations. For \colorbox{gray!20}{Random Pseudo}, each combo score is first averaged over 10 stochastic masking draws (i.e., an additional $Draw_{std}$ averaging layer is applied before computing $Combo_{std}$), which structurally reduces the resulting variance, so its values are shown as reference only. \textsc{RepCore} (0\% masking) and \textsc{Advdist Pseudo} are deterministic under each source-model combination, so their comparison is fair. \textbf{Bold} marks the lower (better) value per column.}
    \label{tab:masking_combo_std_40pct}

    \begin{tabular}{l l cc cc cc cc cc cc}
    \toprule
    \multirow{2}{*}{\textbf{Benchmark}} & \multirow{2}{*}{\textbf{Condition}} &
    \multicolumn{2}{c}{\textbf{$K=10$}} & \multicolumn{2}{c}{\textbf{$K=20$}} & \multicolumn{2}{c}{\textbf{$K=30$}} &
    \multicolumn{2}{c}{\textbf{$K=40$}} & \multicolumn{2}{c}{\textbf{$K=50$}} & \multicolumn{2}{c}{\textbf{Avg}} \\
    \cmidrule(lr){3-4}\cmidrule(lr){5-6}\cmidrule(lr){7-8}\cmidrule(lr){9-10}\cmidrule(lr){11-12}\cmidrule(lr){13-14}
    & & $\rho$ & {MAE} & $\rho$ & {MAE} & $\rho$ & {MAE} & $\rho$ & {MAE} & $\rho$ & {MAE} & $\rho$ & {MAE} \\
    \midrule

    \rowcolor{gray!20} \cellcolor{white}\multirow{3}{*}{\textbf{ARC-Challenge}}
          & \textsc{Random Pseudo} & 0.0108 & {0.0024} & 0.0053 & {0.0025} & 0.0056 & {0.0019} & 0.0048 & {0.0017} & 0.0032 & {0.0014} & 0.0059 & {0.0020} \\
          & \textsc{REPCORE} & \textbf{0.0461} & {\textbf{0.0049}} & \textbf{0.0229} & {\textbf{0.0023}} & \textbf{0.0197} & {\textbf{0.0017}} & \textbf{0.0145} & {\textbf{0.0015}} & \textbf{0.0094} & {\textbf{0.0014}} & \textbf{0.0225} & {\textbf{0.0024}} \\
          & \textsc{Advdist Pseudo} & 0.0474 & {0.0074} & 0.0355 & {0.0061} & 0.0746 & {0.0120} & 0.0901 & {0.0162} & 0.1011 & {0.0187} & 0.0697 & {0.0121} \\
    \midrule

    \rowcolor{gray!20} \cellcolor{white}\multirow{3}{*}{\textbf{BBH}}
          & \textsc{Random Pseudo} & 0.0257 & {0.0035} & 0.0107 & {0.0017} & 0.0075 & {0.0015} & 0.0089 & {0.0015} & 0.0064 & {0.0013} & 0.0118 & {0.0019} \\
          & \textsc{REPCORE} & \textbf{0.0357} & {\textbf{0.0074}} & \textbf{0.0290} & {\textbf{0.0048}} & \textbf{0.0151} & {\textbf{0.0020}} & \textbf{0.0087} & {\textbf{0.0014}} & \textbf{0.0057} & {\textbf{0.0021}} & \textbf{0.0188} & {\textbf{0.0035}} \\
          & \textsc{Advdist Pseudo} & 0.0508 & {0.0118} & 0.0375 & {0.0049} & 0.0189 & {0.0039} & 0.0113 & {0.0030} & 0.0128 & {0.0035} & 0.0263 & {0.0054} \\
    \midrule

    \rowcolor{gray!20} \cellcolor{white}\multirow{3}{*}{\textbf{GSM8K}}
          & \textsc{Random Pseudo} & 0.0093 & {0.0040} & 0.0055 & {0.0024} & 0.0036 & {0.0019} & 0.0023 & {0.0017} & 0.0018 & {0.0014} & 0.0045 & {0.0023} \\
          & \textsc{REPCORE} & 0.0377 & {\textbf{0.0051}} & \textbf{0.0106} & {\textbf{0.0017}} & \textbf{0.0095} & {\textbf{0.0020}} & \textbf{0.0065} & {\textbf{0.0022}} & \textbf{0.0060} & {\textbf{0.0025}} & \textbf{0.0141} & {\textbf{0.0027}} \\
          & \textsc{Advdist Pseudo} & \textbf{0.0229} & {0.0062} & 0.0167 & {0.0086} & 0.0148 & {0.0069} & 0.0093 & {0.0073} & 0.0079 & {0.0060} & 0.0143 & {0.0070} \\
    \midrule

    \rowcolor{gray!20} \cellcolor{white}\multirow{3}{*}{\textbf{MMLU-Pro}}
          & \textsc{Random Pseudo} & 0.0221 & {0.0020} & 0.0096 & {0.0007} & 0.0063 & {0.0007} & 0.0039 & {0.0007} & 0.0029 & {0.0006} & 0.0090 & {0.0009} \\
          & \textsc{REPCORE} & \textbf{0.0341} & {\textbf{0.0052}} & \textbf{0.0183} & {0.0031} & \textbf{0.0121} & {0.0021} & \textbf{0.0061} & {\textbf{0.0010}} & \textbf{0.0072} & {\textbf{0.0017}} & \textbf{0.0156} & {\textbf{0.0026}} \\
          & \textsc{Advdist Pseudo} & 0.0739 & {0.0077} & 0.0209 & {\textbf{0.0024}} & 0.0192 & {0.0021} & 0.0184 & {0.0030} & 0.0152 & {0.0020} & 0.0295 & {0.0034} \\
    \midrule

    \rowcolor{gray!20} \cellcolor{white}\multirow{3}{*}{\shortstack[l]{\textbf{SEED-Bench}\\\textbf{-2-Plus}}}
          & \textsc{Random Pseudo} & 0.0283 & {0.0029} & 0.0143 & {0.0016} & 0.0065 & {0.0013} & 0.0088 & {0.0013} & 0.0061 & {0.0012} & 0.0128 & {0.0017} \\
          & \textsc{REPCORE} & \textbf{0.0387} & {0.0106} & 0.0503 & {\textbf{0.0037}} & 0.0356 & {\textbf{0.0018}} & \textbf{0.0212} & {\textbf{0.0022}} & 0.0192 & {\textbf{0.0018}} & \textbf{0.0330} & {\textbf{0.0040}} \\
          & \textsc{Advdist Pseudo} & 0.0779 & {\textbf{0.0073}} & \textbf{0.0362} & {0.0039} & \textbf{0.0309} & {0.0033} & 0.0255 & {0.0032} & \textbf{0.0151} & {0.0041} & 0.0371 & {0.0044} \\
    \midrule

    \rowcolor{gray!20} \cellcolor{white}\multirow{3}{*}{\textbf{Average}}
          & \textsc{Random Pseudo} & 0.0192 & {0.0030} & 0.0091 & {0.0018} & 0.0059 & {0.0015} & 0.0057 & {0.0014} & 0.0041 & {0.0012} & 0.0088 & {0.0018} \\
          & \textsc{REPCORE} & \textbf{0.0385} & {\textbf{0.0066}} & \textbf{0.0262} & {\textbf{0.0031}} & \textbf{0.0184} & {\textbf{0.0019}} & \textbf{0.0114} & {\textbf{0.0017}} & \textbf{0.0095} & {\textbf{0.0019}} & \textbf{0.0208} & {\textbf{0.0030}} \\
          & \textsc{Advdist Pseudo} & 0.0546 & {0.0081} & 0.0294 & {0.0052} & 0.0317 & {0.0056} & 0.0309 & {0.0065} & 0.0304 & {0.0069} & 0.0354 & {0.0065} \\

    \bottomrule
    \end{tabular}
\end{table*}

\begin{table*}[!htbp]
    \centering
    \footnotesize
    \setlength{\tabcolsep}{2.2pt}
    \renewcommand{\arraystretch}{1.1}

    \captionof{table}{\textbf{$Combo_{std}$ under 50\% masking:} standard deviation across 10 source-model combinations. For \colorbox{gray!20}{Random Pseudo}, each combo score is first averaged over 10 stochastic masking draws (i.e., an additional $Draw_{std}$ averaging layer is applied before computing $Combo_{std}$), which structurally reduces the resulting variance, so its values are shown as reference only. \textsc{RepCore} (0\% masking) and \textsc{Advdist Pseudo} are deterministic under each source-model combination, so their comparison is fair. \textbf{Bold} marks the lower (better) value per column.}
    \label{tab:masking_combo_std_50pct}

    \begin{tabular}{l l cc cc cc cc cc cc}
    \toprule
    \multirow{2}{*}{\textbf{Benchmark}} & \multirow{2}{*}{\textbf{Condition}} &
    \multicolumn{2}{c}{\textbf{$K=10$}} & \multicolumn{2}{c}{\textbf{$K=20$}} & \multicolumn{2}{c}{\textbf{$K=30$}} &
    \multicolumn{2}{c}{\textbf{$K=40$}} & \multicolumn{2}{c}{\textbf{$K=50$}} & \multicolumn{2}{c}{\textbf{Avg}} \\
    \cmidrule(lr){3-4}\cmidrule(lr){5-6}\cmidrule(lr){7-8}\cmidrule(lr){9-10}\cmidrule(lr){11-12}\cmidrule(lr){13-14}
    & & $\rho$ & {MAE} & $\rho$ & {MAE} & $\rho$ & {MAE} & $\rho$ & {MAE} & $\rho$ & {MAE} & $\rho$ & {MAE} \\
    \midrule

    \rowcolor{gray!20} \cellcolor{white}\multirow{3}{*}{\textbf{ARC-Challenge}}
          & \textsc{Random Pseudo} & 0.0156 & {0.0032} & 0.0131 & {0.0012} & 0.0098 & {0.0011} & 0.0087 & {0.0009} & 0.0081 & {0.0009} & 0.0111 & {0.0015} \\
          & \textsc{REPCORE} & \textbf{0.0461} & {\textbf{0.0049}} & \textbf{0.0229} & {\textbf{0.0023}} & \textbf{0.0197} & {\textbf{0.0017}} & \textbf{0.0145} & {\textbf{0.0015}} & \textbf{0.0094} & {\textbf{0.0014}} & \textbf{0.0225} & {\textbf{0.0024}} \\
          & \textsc{Advdist Pseudo} & 0.0681 & {0.0050} & 0.0400 & {0.0091} & 0.0300 & {0.0062} & 0.0207 & {0.0035} & 0.0155 & {0.0026} & 0.0349 & {0.0053} \\
    \midrule

    \rowcolor{gray!20} \cellcolor{white}\multirow{3}{*}{\textbf{BBH}}
          & \textsc{Random Pseudo} & 0.0307 & {0.0038} & 0.0096 & {0.0015} & 0.0091 & {0.0010} & 0.0058 & {0.0012} & 0.0062 & {0.0015} & 0.0123 & {0.0018} \\
          & \textsc{REPCORE} & \textbf{0.0357} & {\textbf{0.0074}} & \textbf{0.0290} & {\textbf{0.0048}} & \textbf{0.0151} & {\textbf{0.0020}} & \textbf{0.0087} & {\textbf{0.0014}} & \textbf{0.0057} & {\textbf{0.0021}} & \textbf{0.0188} & {\textbf{0.0035}} \\
          & \textsc{Advdist Pseudo} & 0.0633 & {0.0083} & 0.0316 & {0.0060} & 0.0353 & {0.0049} & 0.0220 & {0.0038} & 0.0235 & {0.0040} & 0.0351 & {0.0054} \\
    \midrule

    \rowcolor{gray!20} \cellcolor{white}\multirow{3}{*}{\textbf{GSM8K}}
          & \textsc{Random Pseudo} & 0.0116 & {0.0028} & 0.0075 & {0.0027} & 0.0053 & {0.0022} & 0.0038 & {0.0018} & 0.0025 & {0.0019} & 0.0061 & {0.0023} \\
          & \textsc{REPCORE} & 0.0377 & {\textbf{0.0051}} & \textbf{0.0106} & {\textbf{0.0017}} & \textbf{0.0095} & {\textbf{0.0020}} & \textbf{0.0065} & {\textbf{0.0022}} & 0.0060 & {\textbf{0.0025}} & 0.0141 & {\textbf{0.0027}} \\
          & \textsc{Advdist Pseudo} & \textbf{0.0278} & {0.0125} & 0.0125 & {0.0059} & 0.0151 & {0.0041} & 0.0088 & {0.0049} & \textbf{0.0054} & {0.0030} & \textbf{0.0139} & {0.0061} \\
    \midrule

    \rowcolor{gray!20} \cellcolor{white}\multirow{3}{*}{\textbf{MMLU-Pro}}
          & \textsc{Random Pseudo} & 0.0171 & {0.0020} & 0.0087 & {0.0010} & 0.0079 & {0.0008} & 0.0064 & {0.0009} & 0.0062 & {0.0010} & 0.0093 & {0.0011} \\
          & \textsc{REPCORE} & \textbf{0.0341} & {\textbf{0.0052}} & \textbf{0.0183} & {\textbf{0.0031}} & \textbf{0.0121} & {\textbf{0.0021}} & \textbf{0.0061} & {\textbf{0.0010}} & \textbf{0.0072} & {\textbf{0.0017}} & \textbf{0.0156} & {\textbf{0.0026}} \\
          & \textsc{Advdist Pseudo} & 0.0704 & {0.0115} & 0.0330 & {0.0065} & 0.0228 & {0.0059} & 0.0277 & {0.0046} & 0.0261 & {0.0039} & 0.0360 & {0.0065} \\
    \midrule

    \rowcolor{gray!20} \cellcolor{white}\multirow{3}{*}{\shortstack[l]{\textbf{SEED-Bench}\\\textbf{-2-Plus}}}
          & \textsc{Random Pseudo} & 0.0253 & {0.0031} & 0.0126 & {0.0013} & 0.0169 & {0.0012} & 0.0145 & {0.0011} & 0.0106 & {0.0011} & 0.0160 & {0.0016} \\
          & \textsc{REPCORE} & \textbf{0.0387} & {0.0106} & 0.0503 & {\textbf{0.0037}} & \textbf{0.0356} & {\textbf{0.0018}} & \textbf{0.0212} & {\textbf{0.0022}} & \textbf{0.0192} & {\textbf{0.0018}} & \textbf{0.0330} & {\textbf{0.0040}} \\
          & \textsc{Advdist Pseudo} & 0.0577 & {\textbf{0.0092}} & \textbf{0.0337} & {0.0066} & 0.0390 & {0.0087} & 0.0279 & {0.0074} & 0.0215 & {0.0074} & 0.0360 & {0.0079} \\
    \midrule

    \rowcolor{gray!20} \cellcolor{white}\multirow{3}{*}{\textbf{Average}}
          & \textsc{Random Pseudo} & 0.0201 & {0.0030} & 0.0103 & {0.0015} & 0.0098 & {0.0013} & 0.0078 & {0.0012} & 0.0067 & {0.0013} & 0.0109 & {0.0016} \\
          & \textsc{REPCORE} & \textbf{0.0385} & {\textbf{0.0066}} & \textbf{0.0262} & {\textbf{0.0031}} & \textbf{0.0184} & {\textbf{0.0019}} & \textbf{0.0114} & {\textbf{0.0017}} & \textbf{0.0095} & {\textbf{0.0019}} & \textbf{0.0208} & {\textbf{0.0030}} \\
          & \textsc{Advdist Pseudo} & 0.0575 & {0.0093} & 0.0302 & {0.0068} & 0.0284 & {0.0060} & 0.0214 & {0.0048} & 0.0184 & {0.0042} & 0.0312 & {0.0062} \\

    \bottomrule
    \end{tabular}
\end{table*}

\begin{table*}[!htbp]
    \centering
    \footnotesize
    \setlength{\tabcolsep}{4pt}
    \renewcommand{\arraystretch}{1.1}

    \captionof{table}{\textbf{$Draw_{std}$ (Random Pseudo-label Masking):} standard deviation of Spearman $\rho$ and MAE across 10 random masking draws, measuring sensitivity to which source models are masked. Only random masking is stochastic; REPCORE (0\% masking) and adversarial-distance masking are deterministic under each source-model combination.}
    \label{tab:masking_draw_std_combined}
    
    \begin{tabular}{l l cc cc cc cc cc}
    \toprule
    \multirow{2}{*}{\textbf{Benchmark}} & \multirow{2}{*}{\textbf{Ratio}} &
    \multicolumn{2}{c}{\textbf{$K=10$}} & \multicolumn{2}{c}{\textbf{$K=20$}} & \multicolumn{2}{c}{\textbf{$K=30$}} &  
    \multicolumn{2}{c}{\textbf{$K=40$}} & \multicolumn{2}{c}{\textbf{$K=50$}} \\
    \cmidrule(lr){3-4}\cmidrule(lr){5-6}\cmidrule(lr){7-8}\cmidrule(lr){9-10}\cmidrule(lr){11-12}
    & & $\rho$ & {MAE} & $\rho$ & {MAE} & $\rho$ & {MAE} & $\rho$ & {MAE} & $\rho$ & {MAE} \\
    \midrule
    
    \multirow{5}{*}{\textbf{ARC-Challenge}}
      & 10\% & .0324 & {.0049} & .0207 & {.0032} & .0134 & {.0033} & .0094 & {.0030} & .0080 & {.0026} \\
      & 20\% & .0404 & {.0062} & .0191 & {.0050} & .0118 & {.0038} & .0087 & {.0033} & .0060 & {.0030} \\
      & 30\% & .0374 & {.0072} & .0226 & {.0040} & .0139 & {.0038} & .0095 & {.0037} & .0076 & {.0033} \\
      & 40\% & .0426 & {.0062} & .0210 & {.0053} & .0139 & {.0049} & .0105 & {.0045} & .0092 & {.0042} \\
      & 50\% & .0478 & {.0069} & .0295 & {.0044} & .0214 & {.0038} & .0160 & {.0036} & .0132 & {.0033} \\
    \midrule
    
    \multirow{5}{*}{\textbf{BBH}}
      & 10\% & .0464 & {.0064} & .0257 & {.0040} & .0164 & {.0027} & .0123 & {.0022} & .0099 & {.0021} \\
      & 20\% & .0563 & {.0069} & .0275 & {.0040} & .0198 & {.0034} & .0132 & {.0031} & .0117 & {.0028} \\
      & 30\% & .0509 & {.0077} & .0268 & {.0048} & .0183 & {.0042} & .0149 & {.0040} & .0122 & {.0039} \\
      & 40\% & .0593 & {.0088} & .0299 & {.0057} & .0180 & {.0043} & .0154 & {.0044} & .0121 & {.0039} \\
      & 50\% & .0640 & {.0104} & .0358 & {.0062} & .0228 & {.0047} & .0170 & {.0044} & .0149 & {.0045} \\
    \midrule
    
    \multirow{5}{*}{\textbf{GSM8K}}
      & 10\% & .0267 & {.0078} & .0156 & {.0061} & .0095 & {.0048} & .0069 & {.0041} & .0058 & {.0038} \\
      & 20\% & .0324 & {.0071} & .0156 & {.0051} & .0095 & {.0045} & .0068 & {.0041} & .0063 & {.0040} \\
      & 30\% & .0315 & {.0087} & .0135 & {.0058} & .0100 & {.0053} & .0085 & {.0047} & .0070 & {.0046} \\
      & 40\% & .0377 & {.0077} & .0160 & {.0061} & .0091 & {.0052} & .0077 & {.0045} & .0066 & {.0042} \\
      & 50\% & .0371 & {.0089} & .0156 & {.0062} & .0108 & {.0052} & .0092 & {.0047} & .0069 & {.0046} \\
    \midrule
    
    \multirow{5}{*}{\textbf{MMLU-Pro}}
      & 10\% & .0331 & {.0043} & .0177 & {.0026} & .0110 & {.0017} & .0071 & {.0014} & .0055 & {.0015} \\
      & 20\% & .0587 & {.0063} & .0209 & {.0028} & .0150 & {.0021} & .0100 & {.0018} & .0077 & {.0016} \\
      & 30\% & .0486 & {.0074} & .0244 & {.0033} & .0150 & {.0023} & .0099 & {.0020} & .0068 & {.0018} \\
      & 40\% & .0468 & {.0060} & .0221 & {.0029} & .0147 & {.0023} & .0100 & {.0022} & .0086 & {.0021} \\
      & 50\% & .0516 & {.0064} & .0263 & {.0034} & .0157 & {.0027} & .0134 & {.0026} & .0105 & {.0027} \\
    \midrule
    
    \multirow{5}{*}{\shortstack[l]{\textbf{SEED-Bench}\\\textbf{-2-Plus}}}
      & 10\% & .0550 & {.0067} & .0342 & {.0033} & .0213 & {.0023} & .0199 & {.0022} & .0178 & {.0023} \\
      & 20\% & .0833 & {.0073} & .0423 & {.0031} & .0317 & {.0026} & .0220 & {.0025} & .0186 & {.0023} \\
      & 30\% & .0733 & {.0071} & .0430 & {.0040} & .0333 & {.0031} & .0255 & {.0027} & .0188 & {.0026} \\
      & 40\% & .0672 & {.0068} & .0428 & {.0038} & .0254 & {.0033} & .0213 & {.0031} & .0190 & {.0030} \\
      & 50\% & .0645 & {.0075} & .0398 & {.0042} & .0303 & {.0036} & .0232 & {.0035} & .0203 & {.0034} \\
    
    \bottomrule
    \end{tabular}
\end{table*}

\clearpage
\section{Model Pools for Each Benchmark}
\label{sec:model_pools}

This section provides the full model pools used for each benchmark in our experiments. 
The corresponding model lists are reported in Tables~\ref{tab:model_pool_arc}--\ref{tab:model_pool_seed}.

\newcolumntype{Y}{>{\raggedright\arraybackslash}X}
\newcommand{\msep}{,\allowbreak\ }

\newcommand{\ModelPoolTable}[4]{%
\begin{table}[!htbp]
  \centering
  \small
  \setlength{\abovecaptionskip}{6pt}
  \setlength{\belowcaptionskip}{6pt}
  \caption{#3}
  \label{#4}
  \setlength{\tabcolsep}{6pt}
  \begin{tabularx}{0.85\linewidth}{@{}lY@{}}
    \toprule
    \textbf{Benchmark} & \textbf{Model Names} \\
    \midrule
    #1 & #2 \\
    \bottomrule
  \end{tabularx}
\end{table}%
}

\newcommand{\ARCChallengeModels}{%
qwen2.5-72b-instruct\msep phi-4\msep qwen2.5-32b-instruct\msep llama-3.1-nemotron-70b-instruct-hf\msep llama-3.3-70b-instruct\msep qwen2.5-32b\msep deepseek-r1-distill-qwen-14b\msep qwen2-72b-instruct\msep qwen2.5-coder-32b-instruct\msep qwen2.5-14b-instruct-1m\msep dracarys-72b-instruct\msep qwen2.5-14b-instruct\msep qwen2.5-72b\msep aceinstruct-72b\msep smaug-llama-3-70b-instruct-32k\msep qwq-32b-preview\msep acemath-72b-instruct\msep qwen2.5-math-72b-instruct\msep yi-1.5-34b-chat\msep phi-3-medium-4k-instruct\msep yi-1.5-34b-chat-16k\msep dolphin-2.9.2-qwen2-72b\msep qwen2-math-72b-instruct\msep qwen2.5-14b\msep phi-3-medium-128k-instruct\msep qwen2.5-7b-instruct\msep qwen2.5-coder-14b-instruct\msep llama-3.1-tulu-3-70b-dpo\msep solar-pro-preview-instruct\msep hermes-3-llama-3.1-70b\msep llama-3.1-tulu-3-70b\msep aya-expanse-32b\msep internlm2\_5-20b-chat\msep aceinstruct-7b\msep llama-3.1-tulu-3-70b-sft\msep internlm2\_5-7b-chat\msep phi-3.5-mini-instruct\msep dolphin-2.9.2-phi-3-medium-abliterated\msep mistral-small-instruct-2409\msep deephermes-3-mistral-24b-preview\msep dolphin-2.9.1-yi-1.5-34b\msep falcon3-7b-instruct\msep yi-1.5-9b-chat\msep llama-3.1-8b-instruct\msep yi-1.5-9b\msep dolphin3.0-r1-mistral-24b\msep yi-1.5-9b-chat-16k\msep dolphin-2.9.3-yi-1.5-34b-32k\msep qwen2.5-coder-7b-instruct\msep falcon3-10b-instruct\msep qwen2.5-3b-instruct\msep openchat-3.6-8b-20240522\msep qwen2-7b-instruct\msep dolphin-2.9.1-yi-1.5-9b\msep ministral-8b-instruct-2410\msep mistral-nemo-instruct-2407\msep acemath-7b-instruct\msep granite-3.1-8b-instruct\msep dolphin-2.9.3-mistral-nemo-12b\msep c4ai-command-r7b-12-2024\msep daredevil-8b-abliterated\msep llama-3-8b-sfr-iterative-dpo-r\msep meta-llama-3.1-8b-instruct-abliterated\msep aya-expanse-8b\msep neuraldaredevil-8b-abliterated\msep hermes-3-llama-3.1-8b\msep granite-3.0-8b-instruct\msep granite-3.2-8b-instruct\msep qwen2.5-7b\msep llama-3.1-tulu-3-8b-dpo\msep aya-23-35b\msep llama-3.2-3b-instruct\msep llama-3.1-tulu-3-8b\msep yi-1.5-6b-chat\msep olmo-2-1124-7b-instruct\msep llama-3.1-tulu-3-8b-sft\msep mistral-7b-instruct-v0.3\msep dolphin-2.9.3-mistral-7b-32k\msep llama-3.1-8b-magpie-ultra\msep llama-3-refueled\msep dolphin3.0-llama3.1-8b\msep yi-1.5-34b\msep qwen2.5-math-7b-instruct\msep falcon-11b\msep yi-1.5-6b\msep falcon3-3b-instruct\msep yi-1.5-34b-32k\msep granite-3.2-2b-instruct\msep granite-3.1-2b-instruct\msep hermes-3-llama-3.2-3b\msep yi-1.5-9b-32k\msep codestral-22b-v0.1\msep aya-23-8b\msep granite-3.0-2b-instruct\msep qwen2-math-7b\msep openmath2-llama3.1-8b%
}

\newcommand{\BBHModels}{%
phi-4\msep qwen2.5-72b-instruct\msep qwq-32b-preview\msep llama-3.3-70b-instruct\msep deepseek-r1-distill-llama-70b\msep llama-3.1-70b-instruct\msep qwen2.5-14b-instruct\msep qwen2.5-32b-instruct\msep llama-3.1-nemotron-70b-instruct-hf\msep acemath-72b-instruct\msep qwen2.5-coder-32b-instruct\msep deepseek-r1-distill-qwen-14b\msep deepseek-r1-distill-qwen-32b\msep qwen2.5-72b\msep smaug-llama-3-70b-instruct-32k\msep qwen2-72b-instruct\msep aceinstruct-72b\msep dracarys-72b-instruct\msep phi-3-medium-4k-instruct\msep qwen2.5-14b-instruct-1m\msep internlm2\_5-20b-chat\msep qwen2.5-math-72b-instruct\msep dolphin-2.9.2-qwen2-72b\msep mistral-small-instruct-2409\msep phi-3-medium-128k-instruct\msep qwen2.5-coder-14b-instruct\msep dolphin3.0-r1-mistral-24b\msep qwen2-math-72b-instruct\msep internlm2\_5-7b-chat\msep qwen2.5-7b-instruct\msep deephermes-3-mistral-24b-preview\msep gemma-2-27b-it\msep llama-3.1-tulu-3-70b-dpo\msep llama-3.1-tulu-3-70b\msep hermes-3-llama-3.1-70b\msep dolphin-2.9.1-yi-1.5-34b\msep yi-1.5-34b-chat\msep phi-3-small-128k-instruct\msep aya-expanse-32b\msep deepseek-r1-distill-qwen-7b\msep solar-pro-preview-instruct\msep llama-3.1-tulu-3-70b-sft\msep dolphin-2.9.2-phi-3-medium-abliterated\msep dolphin-2.9.3-yi-1.5-34b-32k\msep aceinstruct-7b\msep deepseek-r1-distill-llama-8b\msep llama-3.1-8b-instruct\msep phi-3-small-8k-instruct\msep mistral-nemo-instruct-2407\msep gemma-2-9b-it\msep qwen2.5-coder-7b-instruct\msep acemath-7b-instruct\msep codestral-22b-v0.1\msep granite-3.2-8b-instruct\msep granite-3.1-8b-instruct\msep ministral-8b-instruct-2410\msep falcon3-10b-instruct\msep yi-1.5-34b-chat-16k\msep phi-3.5-mini-instruct\msep llama-3-refueled\msep qwen2.5-7b\msep granite-3.0-8b-instruct\msep dolphin-2.9.3-mistral-nemo-12b\msep qwen2.5-14b\msep yi-1.5-34b-32k\msep qwen2-7b-instruct\msep aya-23-35b\msep yi-1.5-9b-chat\msep openchat-3.6-8b-20240522\msep llama-3-8b-sfr-iterative-dpo-r\msep qwen2.5-3b-instruct\msep hermes-3-llama-3.1-8b\msep k2-chat\msep daredevil-8b-abliterated\msep neuraldaredevil-8b-abliterated\msep falcon3-7b-instruct\msep llama-3.2-3b-instruct\msep dolphin-2.9.1-yi-1.5-9b\msep llama-3.1-tulu-3-8b-dpo\msep llama-3.1-tulu-3-8b\msep qwen2.5-math-7b-instruct\msep meta-llama-3.1-8b-instruct-abliterated\msep llama-3.1-tulu-3-8b-sft\msep openmath2-llama3.1-8b\msep llama-3.1-8b-magpie-ultra\msep yi-1.5-9b-chat-16k\msep dolphin-2.9.3-mistral-7b-32k\msep dolphin3.0-llama3.1-8b\msep yi-1.5-9b\msep aya-expanse-8b\msep yi-1.5-6b-chat\msep granite-3.0-8b-base\msep granite-3.0-2b-instruct\msep granite-3.2-2b-instruct\msep olmo-2-1124-7b-instruct\msep granite-3.1-8b-base\msep mistral-7b-instruct-v0.3\msep granite-3.1-2b-instruct\msep falcon3-3b-instruct\msep hermes-3-llama-3.2-3b\msep aya-23-8b%
}

\newcommand{\GSMModels}{%
acemath-72b-instruct\msep qwen2-math-72b-instruct\msep llama-3.3-70b-instruct\msep qwen3-30b-a3b\msep qwen2.5-72b-instruct\msep phi-4\msep qwq-32b-preview\msep qwen2.5-math-72b-instruct\msep qwen2.5-32b-instruct\msep llama-3.1-tulu-3-70b\msep llama-3.1-70b-instruct\msep qwen2.5-coder-32b-instruct\msep deepseek-r1-distill-qwen-14b\msep llama-3.1-tulu-3-70b-dpo\msep aceinstruct-72b\msep qwen2.5-14b-instruct\msep qwen2-72b-instruct\msep dracarys-72b-instruct\msep qwen2.5-14b-instruct-1m\msep hermes-3-llama-3.1-70b\msep smaug-llama-3-70b-instruct-32k\msep qwen2.5-72b\msep dolphin-2.9.2-qwen2-72b\msep qwen2.5-7b-instruct\msep deephermes-3-mistral-24b-preview\msep llama-3.1-tulu-3-70b-sft\msep mistral-small-instruct-2409\msep openmath2-llama3.1-8b\msep qwen2.5-32b\msep aceinstruct-7b\msep phi-3-medium-4k-instruct\msep llama-3.1-nemotron-70b-instruct-hf\msep yi-1.5-34b-chat\msep falcon3-7b-instruct\msep qwen2.5-14b\msep dolphin-2.9.2-phi-3-medium-abliterated\msep phi-3-medium-128k-instruct\msep deepseek-r1-distill-qwen-7b\msep gemma-2-9b-it\msep qwen2.5-coder-7b-instruct\msep llama-3.1-tulu-3-8b-dpo\msep ministral-8b-instruct-2410\msep aya-expanse-32b\msep dolphin-2.9.1-yi-1.5-34b\msep qwen2.5-7b\msep internlm2\_5-20b-chat\msep phi-3.5-mini-instruct\msep llama-3.1-tulu-3-8b\msep acemath-7b-instruct\msep yi-1.5-9b-chat\msep llama-3.1-8b-instruct\msep deepseek-r1-distill-qwen-32b\msep granite-3.2-8b-instruct\msep mistral-nemo-instruct-2407\msep qwen2.5-math-7b\msep daredevil-8b-abliterated\msep qwen2.5-3b-instruct\msep dolphin-2.9.3-yi-1.5-34b-32k\msep dolphin-2.9.3-mistral-nemo-12b\msep neuraldaredevil-8b-abliterated\msep granite-3.1-8b-instruct\msep yi-1.5-9b-chat-16k\msep internlm2\_5-7b-chat\msep hermes-3-llama-3.1-8b\msep olmo-2-1124-7b-instruct\msep qwen2-7b-instruct\msep llama-3.2-3b-instruct\msep openchat-3.6-8b-20240522\msep qwen2-math-7b\msep dolphin-2.9.1-yi-1.5-9b\msep deepseek-r1-distill-llama-70b\msep aya-expanse-8b\msep llama-3.1-tulu-3-8b-sft\msep dolphin-2.9.3-mistral-7b-32k\msep meta-llama-3.1-8b-instruct-abliterated\msep granite-3.0-8b-instruct\msep codestral-22b-v0.1\msep yi-1.5-6b-chat\msep falcon3-10b-instruct\msep granite-3.0-8b-base\msep granite-3.2-2b-instruct\msep granite-3.1-2b-instruct\msep llama-3-refueled\msep granite-3.0-2b-instruct\msep hermes-3-llama-3.2-3b\msep granite-3.1-8b-base\msep aya-23-35b\msep falcon3-3b-instruct\msep mistral-7b-instruct-v0.3\msep deepseek-r1-distill-llama-8b%
}

\newcommand{\MMLUProModels}{%
qwq-32b-preview\msep qwen2.5-72b-instruct\msep llama-3.3-70b-instruct\msep qwen2.5-32b-instruct\msep deepseek-r1-distill-qwen-14b\msep llama-3.1-nemotron-70b-instruct-hf\msep llama-3.1-70b-instruct\msep aceinstruct-72b\msep llama-3.1-tulu-3-70b-dpo\msep dracarys-72b-instruct\msep llama-3.1-tulu-3-70b\msep qwen2.5-14b-instruct\msep qwen2.5-14b-instruct-1m\msep qwen2.5-72b\msep qwen2.5-coder-32b-instruct\msep acemath-72b-instruct\msep qwen2.5-math-72b-instruct\msep qwen2.5-32b\msep solar-pro-preview-instruct\msep dolphin-2.9.2-qwen2-72b\msep smaug-llama-3-70b-instruct-32k\msep qwen2-math-72b-instruct\msep hermes-3-llama-3.1-70b\msep phi-3-medium-4k-instruct\msep qwen2.5-7b-instruct\msep deephermes-3-mistral-24b-preview\msep qwen2.5-coder-14b-instruct\msep phi-3-medium-128k-instruct\msep phi-3-small-128k-instruct\msep gemma-2-27b-it\msep llama-3.1-tulu-3-70b-sft\msep mistral-small-instruct-2409\msep aceinstruct-7b\msep dolphin-2.9.2-phi-3-medium-abliterated\msep phi-3-small-8k-instruct\msep yi-1.5-34b-chat\msep qwen2.5-14b\msep yi-1.5-34b-chat-16k\msep qwen2.5-7b\msep falcon3-7b-instruct\msep acemath-7b-instruct\msep gemma-2-9b-it\msep phi-3.5-mini-instruct\msep dolphin-2.9.1-yi-1.5-34b\msep aya-expanse-32b\msep qwen2.5-coder-7b-instruct\msep yi-1.5-9b-chat\msep qwen2-7b-instruct\msep mistral-nemo-instruct-2407\msep llama-3.1-tulu-3-8b-dpo\msep qwen2.5-3b-instruct\msep falcon3-10b-instruct\msep dolphin-2.9.3-yi-1.5-34b-32k\msep yi-1.5-9b-chat-16k\msep daredevil-8b-abliterated\msep llama-3.1-tulu-3-8b\msep ministral-8b-instruct-2410\msep dolphin-2.9.1-yi-1.5-9b\msep neuraldaredevil-8b-abliterated\msep llama-3.1-70b\msep yi-1.5-34b-32k\msep qwen2.5-math-7b-instruct\msep granite-3.2-8b-instruct\msep meta-llama-3.1-8b-instruct-abliterated\msep dolphin-2.9.3-mistral-nemo-12b\msep k2-chat\msep falcon3-3b-instruct\msep hermes-3-llama-3.1-8b\msep llama-3.1-tulu-3-8b-sft\msep llama-3-8b-sfr-iterative-dpo-r\msep yi-1.5-9b\msep yi-1.5-6b-chat\msep qwen2.5-math-7b\msep codestral-22b-v0.1\msep openchat-3.6-8b-20240522\msep dolphin3.0-llama3.1-8b\msep granite-3.0-8b-instruct\msep olmo-2-1124-7b-instruct\msep llama-3.1-8b-magpie-ultra\msep aya-23-35b\msep aya-expanse-8b\msep llama-3-refueled\msep mistral-7b-instruct-v0.3\msep qwen2-math-7b\msep granite-3.2-2b-instruct\msep granite-3.1-2b-instruct\msep yi-1.5-9b-32k\msep yi-1.5-6b\msep dolphin-2.9.3-mistral-7b-32k\msep granite-3.0-2b-instruct\msep hermes-3-llama-3.2-3b%
}

\newcommand{\SEEDBenchTwoPlusModels}{%
qwen2.5-vl-32b-instruct\msep internvl2\_5-78b-mpo\msep qwen2.5-vl-72b-instruct\msep internvl3-78b\msep internvl3-38b\msep internvl2\_5-38b-mpo\msep qwen2-vl-72b-instruct\msep ovis2-16b\msep ovis2-34b\msep mimo-vl-7b-rl\msep internvl2\_5-78b\msep vARCo-vision-2.0-14b\msep internvl3-14b\msep internvl2\_5-26b-mpo\msep wethink-qwen2.5vl-7b\msep internvl2\_5-38b\msep internvl2-llama3-76b\msep ovis2-8b\msep internvl3-9b\msep qwen2.5-vl-7b-instruct\msep internvl2\_5-26b\msep internvl2\_5-8b-mpo\msep ovis-u1-3b\msep ovis2-4b\msep internvl2-40b\msep internvl3-8b\msep internvl2\_5-8b\msep sail-vl-1d6-8b\msep ovis1.6-gemma2-9b\msep gemma-3-27b-it\msep sail-vl-1d5-8b\msep internvl2-8b-mpo\msep internvl2\_5-4b-mpo\msep qwen2-vl-7b-instruct\msep internvl2-26b\msep gemma-3-12b-it\msep ovis2-2b\msep internvl2\_5-4b\msep internvl-chat-v1-5\msep ristretto-3b\msep sail-vl-1d5-2b\msep points-qwen-2-5-7b-chat\msep internvl2-8b\msep qwen2.5-vl-3b-instruct\msep vARCo-vision-2.0-1.7b\msep ovis1.5-llama3-8b\msep llava-onevision-qwen2-7b-si-hf\msep llava-onevision-qwen2-7b-ov-hf\msep points-yi-1-5-9b-chat\msep phi-3-vision-128k-instruct\msep internvl2-4b\msep llama-3.2-11b-vision-instruct\msep xinyuan-vl-2b\msep qwen2-vl-2b-instruct\msep internvl3-2b\msep internvl2\_5-2b-mpo\msep qtunevl1\_5-2b\msep phi-3.5-vision-instruct\msep granite-vision-3.3-2b\msep gemma-3-4b-it\msep minicpm-llama3-v-2\_5\msep granite-vision-3.2-2b\msep internvl2\_5-2b\msep ovis2-1b\msep vintern-3b-beta\msep smolvlm2-2.2b-instruct\msep granite-vision-3.1-2b-preview\msep h2ovl-mississippi-2b\msep smolvlm-instruct\msep llama3-llava-next-8b-hf\msep mini-internvl-chat-4b-v1-5\msep internvl2\_5-1b-mpo\msep mini-internvl-chat-2b-v1-5\msep llava-interleave-qwen-7b-hf\msep llava-interleave-qwen-7b-dpo-hf\msep molmo-7b-d-0924\msep internvl2-2b\msep qtunevl1\_5-3b\msep molmo-7b-o-0924\msep pixtral-12b\msep llava-onevision-qwen2-0.5b-ov-hf\msep internvl2\_5-1b\msep internvl2-1b\msep phi-4-multimodal-instruct\msep llava-1.5-13b-hf\msep smolvlm-synthetic\msep vintern-1b-v2\msep sharegpt4v-7b-hf\msep ovis1.6-llama3.2-3b\msep llava-onevision-qwen2-0.5b-si-hf\msep llava-1.5-7b-hf\msep minicpm-v\msep minicpm-v-2%
}

\ModelPoolTable{ARC Challenge}{\ARCChallengeModels}{Models used for ARC Challenge benchmark.}{tab:model_pool_arc}

\ModelPoolTable{BBH}{\BBHModels}{Models used for BBH benchmark.}{tab:model_pool_bbh}

\ModelPoolTable{GSM8K}{\GSMModels}{Models used for GSM8K benchmark.}{tab:model_pool_gsm8k}

\ModelPoolTable{MMLU-Pro}{\MMLUProModels}{Models used for MMLU-Pro benchmark.}{tab:model_pool_mmlu_pro}

\ModelPoolTable{SEED-Bench-2-Plus}{\SEEDBenchTwoPlusModels}{Models used for SEED-Bench-2-plus benchmark.}{tab:model_pool_seed}


\end{document}